\documentclass[lettersize,journal]{IEEEtran}
\usepackage{amsmath,amsfonts}
\usepackage{algorithmic}
\usepackage{algorithm}
\usepackage{array}
\usepackage[caption=false,font=normalsize,labelfont=sf,textfont=sf]{subfig}
\usepackage{textcomp}
\usepackage{stfloats}
\usepackage{url}
\usepackage{verbatim}

\usepackage{tikz}
\usetikzlibrary{shapes,arrows}

\usepackage{pifont}

\usepackage{url}

\usepackage{booktabs}       % professional-quality tables
\usepackage{nicefrac}       % compact symbols for 1/2, etc.
\usepackage{microtype}      % microtypography
\usepackage{xcolor}         % colors
\usepackage{graphicx}

\usepackage[normalem]{ulem} 
\usepackage{multirow}
\usepackage{multicol}
\usepackage{wrapfig}

\usepackage[scr=boondox,cal=esstix]{mathalpha}
\usepackage{float}                  % 图片浮动位置
\usepackage{overpic}                % 叭啦叭啦，与图片排版相关\
\usepackage{listings}
\usepackage[numbers]{natbib}

\usepackage{graphicx}
\usepackage{hyperref}
\begin{document}

\title{SimAD: A Simple Dissimilarity-based Approach for Time Series Anomaly Detection}

% 定义默认颜色为蓝色的命令
% \newcommand{\rv}[1]{\textcolor{blue}{#1}}
  
\newcommand{\rv}[1]{#1}

% \author{Anonymous Authors}
\author{Zhijie Zhong, Zhiwen Yu\IEEEauthorrefmark{1},~\IEEEmembership{Senior Member~IEEE}, Xing Xi, Yue Xu, Wenming Cao,~\IEEEmembership{Member~IEEE}, Yiyuan Yang, Kaixiang Yang,~\IEEEmembership{Member~IEEE}, Jane You
\IEEEcompsocitemizethanks{
\IEEEcompsocthanksitem Zhijie Zhong is with Pengcheng Laboratory, Shenzhen, Guangdong, 518066, China, and also with the School of Future Technology, South China University of Technology, Guangzhou, Guangdong 510650, China.
\IEEEcompsocthanksitem  Zhiwen~Yu is with the School of Computer Science and Engineering, South China University of Technology, Guangzhou, Guangdong 510650, China, and also with the Pengcheng Laboratory, Shenzhen, Guangdong 518066, China. Email: zhwyu@scut.edu.cn. Telephone number: 86-20-62893506. Fax number: 86-20-39380288.
\IEEEcompsocthanksitem Xing Xi, Yue Xu and Kaixiang Yang are with the School of Computer Science and Engineering, South China University of Technology, Guangzhou, Guangdong 510650, China.
\IEEEcompsocthanksitem Wenming Cao is with the Department of Information and Computing Science, Chongqing Jiaotong University.
\IEEEcompsocthanksitem Yiyuan Yang, from the Department of Computer Science at the University of Oxford, Oxford, UK, worked down as an intern at Alibaba Group.
\IEEEcompsocthanksitem Jane You is with the Department of Industrial and Systems Engineering in The Hong Kong Polytechnic University, China. (Email: jane.you@polyu.edu.hk)
\IEEEcompsocthanksitem \IEEEauthorrefmark{1}Corresponding author: Zhiwen Yu.
}
}

        % <-this % stops a space
% \thanks{This paper was produced by the IEEE Publication Technology Group. They are in Piscataway, NJ.}stops a space

\markboth{Journal of \LaTeX\ Class Files,~Vol.~14, No.~8, August~2021}%
{Shell \MakeLowercase{\textit{et al.}}: A Sample Article Using IEEEtran.cls for IEEE Journals}


\maketitle

\begin{abstract}
Despite the prevalence of reconstruction-based deep learning methods, time series anomaly detection remains a tremendous challenge.
Existing approaches often struggle with limited temporal contexts, insufficient representation of normal patterns, and flawed evaluation metrics, all of which hinder their effectiveness in detecting anomalous behavior.
To address these issues, we introduce a $\textbf{\uline{Sim}}$ple dissimilarity-based approach for time series $\textbf{\uline{A}}$nomaly $\textbf{\uline{D}}$etection, referred to as $\textbf{\uline{SimAD}}$. 
Specifically, SimAD first incorporates a patching-based feature extractor capable of processing extended temporal windows and employs the EmbedPatch encoder to fully integrate normal behavioral patterns. 
Second, we design an innovative ContrastFusion module in SimAD, which strengthens the robustness of anomaly detection by highlighting the distributional differences between normal and abnormal data.
Third, we introduce two robust enhanced evaluation metrics, Unbiased Affiliation (UAff) and Normalized Affiliation (NAff), designed to overcome the limitations of existing metrics by providing better distinctiveness and semantic clarity.
The reliability of these two metrics has been demonstrated by both theoretical and experimental analyses. Experiments conducted on seven diverse time series datasets clearly demonstrate SimAD's superior performance compared to state-of-the-art methods, achieving relative improvements of $\textbf{19.85\%}$ on F1, $\textbf{4.44\%}$ on Aff-F1, $\textbf{77.79\%}$ on NAff-F1, and $\textbf{9.69\%}$ on AUC on six multivariate datasets. 
Code and pre-trained models are available at 
% \url{https://anonymous.4open.science/r/SimAD-1116}.
\url{https://github.com/EmorZz1G/SimAD}.
\end{abstract}


\section{Introduction}

% 时序异常检测是时序中一项重要的任务，旨在发现时间序列数据中的异常模式并进行准确定位\cite{intro1,intro2}。过去的研究主要聚焦于使用基于重构的无监督方案解决这一问题\cite{intro3}，它们假设，通常模型在正常数据上完美地训练，因此在测试阶段会对异常数据输出更高的异常分数。
Time series anomaly detection (TSAD) is a critical component of time series analysis, focused on accurately detecting abnormal patterns in time series data and identifying their specific locations \cite{ada_mem_bls, tnnls_app1}. TSAD methods utilize time series data to identify anomalies in web traffic, which play a vital role in ensuring the stability, security, and efficient functioning of web services \cite{tnnls_gnn}. Unsupervised methods have garnered considerable attention in academic research, particularly for addressing this challenge through reconstruction-based approaches \cite{bls2,intro3,tnnls_app2,yang2024multi,zhao2016gsead}. These methods assume that models are perfectly trained on normal data and assign higher anomaly scores to anomalous data during the testing phase.
However, these methods have shown insufficient performance in practical applications. 
A thorough review of existing research \cite{amsl, sim_net, uni_ad, bls}, coupled with comprehensive experiments (detailed in our analyses), has enabled us to identify several critical challenges in the field of time series anomaly detection:
\begin{itemize}
    \item \textbf{Challenge 1}: Many methods rely on the reconstruction assumption, which is inadequate for enhancing the detection performance and may not always hold true \cite{anom_trans,amsl}. Our experiments, including both sensitivity and ablation analyses, validate this limitation.
    \item \textbf{Challenge 2}: Failure to adequately utilize extended time windows: The complexity of the attention mechanism has constrained previous methods, capping the window length at 200 or fewer \cite{anom_trans,tcn_ed}. This limitation, in turn, prevents the capture of more informative data.
    \item \textbf{Challenge 3}: Limited expressive power hinders the representation of normal features. On one hand, certain methods fail to effectively model either normal or abnormal data, or both. On the other hand, most models are constrained by a limited number of parameters, which restricts their expressive capacity.
\end{itemize}

To address the above challenges, we propose a \textbf{Sim}ple dissimilarity-based approach for time series \textbf{A}nomaly \textbf{D}etection method, referred to as \textbf{SimAD}, in both univariate and multivariate settings. Specifically,  
to tackle \textbf{Challenge 2}, we design a feature extractor that can process longer time windows by splitting the sequence into multiple patches. This strategy enables SimAD to learn extended temporal receptive fields while using fewer parameters. To address \textbf{Challenges 1} and \textbf{3}, we design the EmbedPatch encoder and incorporate an enhanced attention mechanism for layer-wise modeling of dissimilarities between normal and abnormal data features. Thus, the EmbedPatch encoder can learn discriminative representations of normal data more effectively. Essentially, the proposed EmbedPatch encoder is a prototype, using $V$ vectors to store prototypical features of normal data.  
The key differences between the EmbedPatch encoder and traditional prototypes can be highlighted in two aspects: 1) Prototypes typically use the last layer to preserve features, while the EmbedPatch encoder, acting as the value matrices for the attention mechanism, not only preserves features at different layers but also serves as a querying repository for each level; 2) In traditional prototypes, query results are obtained by calculating the similarity between prototypes and features. In contrast, the EmbedPatch encoder generates query results by interacting the attention scores with the value matrices.
With the EmbedPatch encoder, each layer of attention mechanism incorporates a series of embedded patches to learn features of that specific layer, giving it a stronger capability to learn richer and more distinctive representations, which are beneficial to performance improvement of anomaly detection. 
Finally, we introduce the ContrastFusion module to amplify the divergence of normal and abnormal data's distributions for further accentuating their dissimilarity.

Furthermore, we have analyzed the shortcomings of existing evaluation metrics, including inflated metrics, low discriminative power, and insufficient semantic relevance. These issues have contributed to false perceptions of progress in TSAD \cite{metric1, metric2, metric3}. To enable a fair comparison of performance, we propose two improved evaluation metrics: Unbiased Affiliation (\textbf{UAff}) and Normalized Affiliation (\textbf{NAff}). These metrics are designed to address the limitations of existing metrics by providing better distinctiveness and semantics. We have conducted analyses from both theoretical and experimental perspectives to establish the reliability of the newly proposed metrics. 
As demonstrated via extensive experiments, the proposed approach outperforms the state-of-the-art (SOTA) methods in time series anomaly detection.

% 我们的论文的贡献总结如下：
% （1）一种全新的简单到有效的长周期时序异常检测算法SimAD，能够支持更长的时间窗口大小。
% （2）提出了改进的异常检测评估指标UAff与NAff，并对其进行了理论与实验分析说明其可靠性。
% （3）我们的算法在真实世界数据集上的效果远超目前的state-of-the-arts水平，且代码与模型都是公开的。

In summary, our paper makes the following contributions: 
\begin{enumerate}
    \item We propose \textbf{SimAD}, a simple yet effective algorithm designed for time series anomaly detection that can handle extended time windows. SimAD is a straightforward framework based on dissimilarity measures, and its effectiveness has been validated through comprehensive evaluations from multiple perspectives.
    \item 
    We introduce two improved TSAD evaluation metrics, \textbf{UAff} and \textbf{NAff}, which address challenges such as inaccurate assessments and lack of semantic clarity. The reliability of the proposed metrics is demonstrated through theoretical and experimental analyses. Furthermore, a comprehensive evaluation of the issues, limitations, and strengths of existing metrics is conducted.
    \item Our algorithm outperforms significantly SOTA methods on real-world datasets, excelling in point-level evaluations such as F1 and AUC, as well as sequence-level evaluations like Aff, UAff and NAff (See Fig. \ref{sfig:comparison}). Moreover, we have released both our code and pre-trained models.
\end{enumerate}

\begin{figure}[ht]
    \centering
    \includegraphics[width=0.9\linewidth]{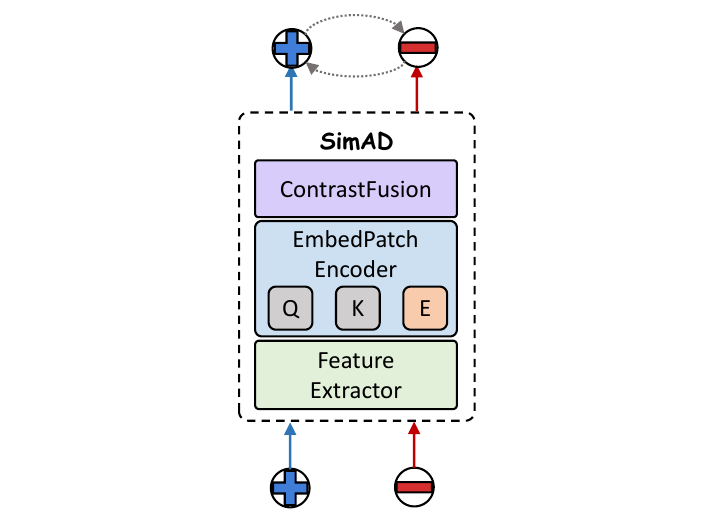}
    \caption{The Overview of SimAD.}
    \label{fig:fwa}
\end{figure}

\section{Related Work}
% 由于时序异常检测的标签稀缺，无监督方法受到广泛关注。通常可以将这些方法简单分类为：基于经典机器学习的算法：\cite{deep_svdd,deep_if}利用经典的机器学习算法改造为深度网络结构，能够好处理复杂数据；基于重构的：\cite{usad,tcn_ed,ae1}都使用正常数据训练模型，采用重构误差作为异常分数，并给予测试中的异常数据更高的异常分数；基于预测的：\cite{timesnet}学习历史数据并预测未来数据，将预测误差作为检测方法：生成对抗学习的：\cite{gan1,gan2,gan3}利用生成代理任务学习正常数据分布，并使用判别器网络检测异常。
\rv{Unsupervised learning methods \cite{bls2,ZHAO2016126} have garnered considerable attention due to the scarcity of labeled data for anomaly detection in time series sequences \cite{ada_mem_bls,usad,anom_trans,dcdetector,amsl,trans_ad}.} 
% 在过去的研究中，异常检测算法已经成为数据分析的重要算法之一，在遥感、图像等领域广泛应用。
\rv{In past studies, anomaly detection algorithms have become one of the key data analysis tools and are widely used in fields like remote sensing and imaging \cite{7862818,7317572,6784134,zhao2024novel}.}
These methods can be categorized as follows (detailed discussion in Appendix \ref{sec:app_releted}):
(1) Algorithms based on classical machine learning \cite{deep_svdd, deep_if,du2016spectral} transform traditional machine learning approaches into deep networks, enhancing their ability to handle complex data.
(2) Reconstruction-based approaches \cite{usad, tcn_ed, ae1} involve training models using normal data and leveraging reconstruction error as an anomaly score, attributing higher scores to anomalous data during testing.
(3) Prediction-based techniques, as demonstrated in \cite{timesnet}, learn from historical data to predict future observations, considering prediction errors as the foundation for anomaly detection.
(4) Generative adversarial learning based methods \cite{gan1, gan2, gan3} utilize generative models to learn the distribution of normal data and a discriminator network to detect anomalies.

Recently, some innovative algorithms have emerged in the field of anomaly detection, including: (1) Transformer-based approaches \cite{anom_trans, trans1, trans_ad} leverage the power of Transformer, which has shown exceptional success in natural language processing tasks, and are increasingly being applied to anomaly detection.
(2) Contrastive learning-based methods \cite{couta, ncad, dcdetector, cont1, cont2} utilize contrastive learning to obtain robust representations, specifically tailored for anomaly detection.
(3) Diffusion-based methods \cite{d3r, diffusion1} model the propagation of anomalies in complex networks and time series sequences by using the diffusion process.
(4) Large Language Models (LLMs) \cite{gpt2adpt} 
exploit cutting-edge models, like GPT-2, adapted specifically for anomaly detection tasks, capitalizing on the sophisticated architectures and knowledge representation capabilities.

These algorithms can be considered extensions or variations of the above categories, which incorporate advanced network structures and knowledge representation methods to enhance anomaly detection performance. 
However, existing methods have not effectively addressed the above three challenges. 
Most methods are constrained by a limited number of learnable parameters, which hinders them from capturing long-term dependencies in data.
Moreover, some reconstruction-based methods solely focus on the task of reconstruction. 
% The relationship between our dissimilarity-based approach and Contrastive learning-based methods 的区别体现在框架的结构、动机与实现 discussed in detail in the Appendix \ref{sec:insights}.
The differences between our dissimilarity-based approach and Contrastive learning-based methods are in the framework (EmbedPatch \& ContrastFusion), motivation (dissimilarity of perspectives \cite{tnnls_diss}), and implementation, as discussed in Appendix \ref{sec:insights}.

\section{Proposed Approach}
\begin{figure*}
    \centering
    \includegraphics[width=\linewidth]{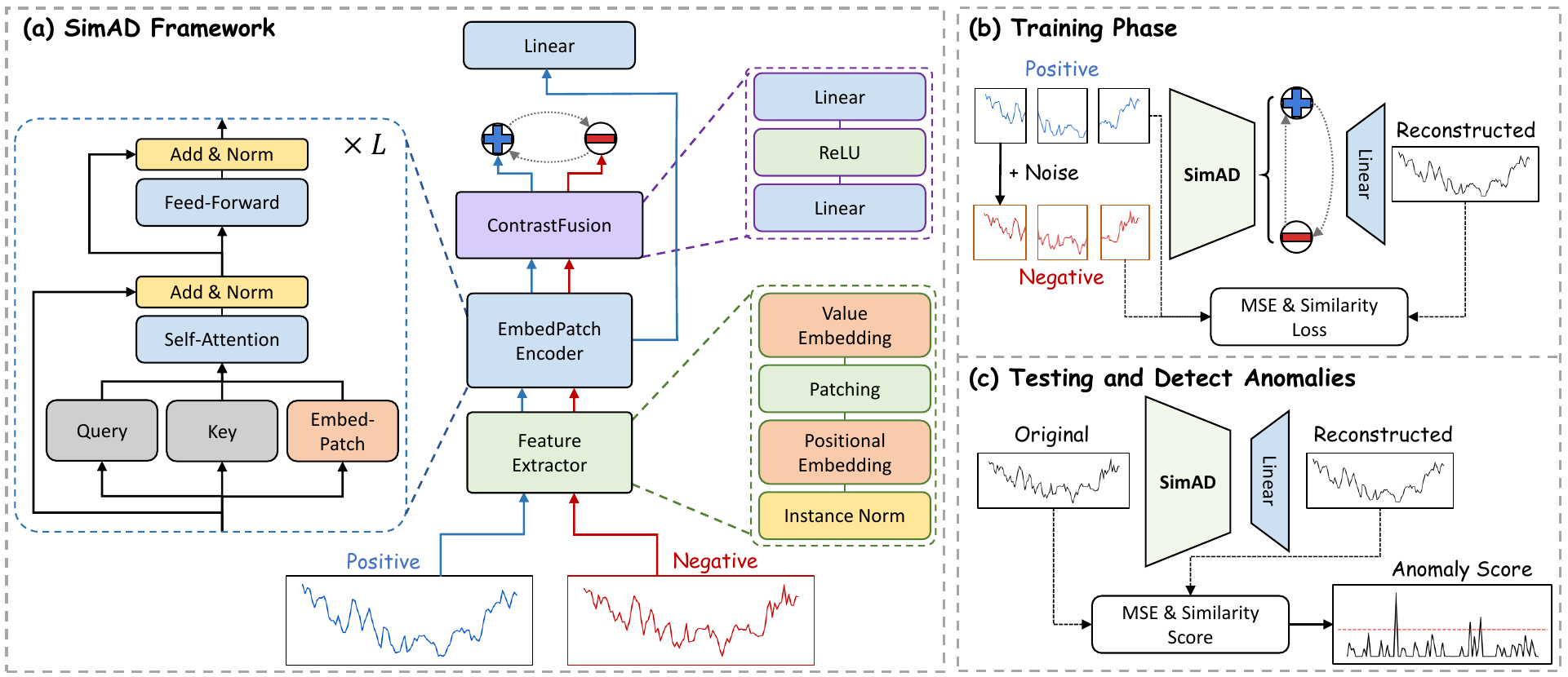}
    \caption{The framework and workflow of SimAD.}
    \label{fig:fw}
\end{figure*}

\subsection{Framework of SimAD}

\subsubsection{Overview}
In Fig. \ref{fig:fw}(a), SimAD comprises a Feature Extractor, EmbedPatch Encoder, and ContrastFusion module. In the context of temporal anomaly detection, the original time series data is denoted as \(\mathbf{X} \in \mathbb{R}^{T\times C}\), where \(T\) denotes the length of time series and \(C\)  the number of channels. The goal of this task is to predict the label \(\mathcal{y} \in \mathbb{R}^{T}\) for $\textbf{X}$, where  \(y_i = 1\) indicates an anomaly at the \(i\)-th time point and $y_i = 0$ otherwise. The input data \(\mathbf{X}\) is first processed by the Feature Extractor to generate patch tokens \(\mathbf{N}\). 
Next, at each layer of the EmbedPatch Encoder, the attention mechanism combines the patch tokens from the previous layer with the current tokens to capture their interdependencies. We then utilize a linear layer to reconstruct the original features and compute anomaly scores. Finally, we design the ContrastFusion module 
to strengthen the distinction between normal and abnormal data distributions, thereby enhancing overall detection performance.

\subsubsection{Feature Extractor}
To extract unified sequence-level temporal features, we devised a patch-based feature extractor, which allows SimAD to process longer time windows and capture richer semantic information. Specifically, \(\mathbf{X}\) is first fed to the feature extractor.
To address distribution shifts \rv{\cite{instance_norm,llm4ts}}, we introduce Instance Normalization (IN) \rv{\cite{instance_norm}}. 
The improved Positional Embedding (\(\operatorname{PE}\)) is then used to incorporate temporal positional encoding, enabling SimAD to learn the relationships between \rv{\(C\)} channels actively. 
% 改进后的PE与原始的PE\cite{attention_is}相比，只对时序的位置进行正弦位置编码，但不需要对通道编号编码，这是因为时序的通道与语言模型的词嵌入含义不同导致。
\rv{The improved positional encoding (PE) applies sinusoidal positional encoding solely to the temporal positions, in contrast to the original PE \cite{attention_is}, without the necessity of encoding the channel indices. This distinction arises from the differing implications of the channels in time series and the word embeddings in the language model.}
% The \(\operatorname{Patching}\) operation involves dividing \(\mathbf{X}^{\prime}\) into multiple (\(M\)) patches of length \(P\) (\(\mathbf{N}^{\prime} \in \mathbb{R}^{ M \times P \times C}\)) and rearranging them as \(\mathbf{N}^{\prime} \in \mathbb{R}^{ M \times (P \cdot C)}\). 
% 这意味着，首先对于\(C\)个通道的时序通过两个操作，第一步是由\(\mathbf{X}\in \mathbb{R}^{T\times C}\)切分为\(\mathbf{N}^{\prime} \in \mathbb{R}^{ M \times P \times C}\)，其中\(T=M\cdot P\)，第二步将\(\mathbf{N}^\prime\)的维度重新排列，即\(\mathbf{N}^{\prime} \in \mathbb{R}^{ M \times P \times C} \rightarrow \mathbf{N}^{\prime} \in \mathbb{R}^{ M \times (P \cdot C)}\)。

\rv{Then, for \(C\) channels, the temporal sequence is processed through two operations via \(\operatorname{Patching}(\cdot)\). 
First, the original input \(\mathbf{X} \in \mathbb{R}^{T \times C}\) is segmented into \(M\) patches of length \(P\), resulting in an intermediate representation \(\mathbf{N}^{\prime} \in \mathbb{R}^{M \times P \times C}\), where  \(T = M \times P\). Next, \(\mathbf{N}^{\prime}\) is reshaped to obtain \(\mathbf{N}^{\prime} \in \mathbb{R}^{M \times (P \cdot C)}\).}
% 请注意，在本文中，我们使用带上标\prime的符号表示中间过程的变量。
\rv{In this paper, we use the superscript ``\(\prime\)" to denote intermediate process variables, such as \(\mathbf{N}^\prime\).}
% Finally, Value Embedding (\(\operatorname{VE}\)) refers to a linear transformation that maps all the patches into a unified space
% , \(\mathbf{N} \in \mathbb{R}^{ M \times D}\)，具体为\(\mathbf{N}=\operatorname{VE}(\mathbf{N}^\prime)=\operatorname{LayerNorm}\left(\operatorname{Linear}\left(\operatorname{LayerNorm}\left(\mathbf{N}^\prime\right)\right)\right)\). Here, \(\operatorname{LayerNorm}\)表示层规划化\cite{attention_is}，它被证明更适合序列数据。
\rv{Finally, Value Embedding (\(\operatorname{VE}\)) is a simple linear transformation that maps all patches into a unified \(D\)-dimensional space, specifically \(\mathbf{N} \in \mathbb{R}^{M \times D}\). It is defined as \(\mathbf{N}=\operatorname{VE}(\mathbf{N}^\prime)=\operatorname{LayerNorm}\left(\operatorname{Linear}\left(\operatorname{LayerNorm}\left(\mathbf{N}^\prime\right)\right)\right)\). Here, \(D\) is a predefined and represents the final dimension of the model. The term \(\operatorname{LayerNorm}\) refers to layer normalization \cite{attention_is}, which has been shown to be more suitable for sequential data. The term \(\operatorname{Linear}\) refers to linear layer.}
The processing performed by the Feature Extractor can be described as below:
\begin{equation}
\begin{aligned}
    \mathbf{X}^{\prime} &= \operatorname{PE}(\operatorname{IN}(\mathbf{X})),\\
    \mathbf{N} &= \operatorname{VE}(\operatorname{Patching}(\mathbf{X}^{\prime})),
\end{aligned}
\end{equation}
\rv{where \(\mathbf{N}\) refers to the \textbf{naive representation} and the original input of EmbedPatch Encoder.}
% 得益于我们设计的Feature Extractor，SimAD能够处理更长的时间窗口的数据，我们的实验证明更长的窗口是必需的。
With the designed feature extractor that maintains the framework's simplicity, SimAD can handle data with longer time windows, whose necessity will be demonstrated by in Section \ref{sec:sen_ana}. 
%In fact, to maintain the overall simplicity of SimAD, we did not overly design the feature extractor but left some room for flexibility.

\subsubsection{EmbedPatch Encoder}
% 为了增强SimAD记忆正常样本的能力，提升模型的泛化能力。我们简单地改进了原始的注意力模型，使用嵌入查询的EmbedPatch $\mathbf{E}^{(i)} \in \mathbb{R}^{V \times D}$融入骨架中，其中$V$是patch embedding的数量。由于该数量的设置需要一定先验，这导致往往需要较多的嵌入数量能够满足不同情况。因此一个线性投影跟在EmbedPatch之后，得到 $\mathbf{E}^{(i),\prime} \in \mathbb{R}^{N \times D}$，其中$N \ll V$。这一简单的组件能够让SimAD高效地，无需先验的学习正常样本的相关信息。为了实现这个，我们采用改进的多头注意力层。具体而言，对于每一个头$k={1,2,\ldots,U}$，查询矩阵为$\mathbf{Q}^{(i)}_k =  \mathbf{N}^{(i)} W^Q_k$，key矩阵为$\mathbf{K}^{(i)}_k =  \mathbf{N}^{(i)} W^K_k$，value矩阵为$\mathbf{K}^{(i)}_k =  \mathbf{W}^V_k \mathbf{E}^{(i),\prime}$。其中$\mathbf{W}^Q_k,\mathbf{W}^K_k \in \mathbb{R}^{d \times d}$，$\mathbf{W}^V_k \in \mathbb{R}^{N \times V}$，其中$d=\lfloor\frac{D}{U}\rfloor$。之后，我们定义每一层中不同patch之间的注意力计算如下：
To enhance SimAD's capability of modeling the dissimilarity between normal and abnormal samples, we made an improvement to the original attention mechanism. Inspired by \cite{amsl,uni_ad,bls},
this improvement involves incorporating the EmbedPatch \(\mathbf{E}^{(i)} \in \mathbb{R}^{V \times D}\) with embedded queries into the EmbedPatch Encoder backbone of SimAD, where \(V\) denotes the number of patch embeddings (EmbedPatch) and \((i)\) for \(i\)-th layer of encoder. Determining a proper \(V\) often requires prior and a larger number of embeddings to accommodate different scenarios \cite{time_llm}. \rv{Therefore, a linear projection is stacked on EmbedPatch to obtain \(\mathbf{E}^{(i),\prime} \in \mathbb{R}^{M^* \times D}\) for reducing the dimensionality, where \(M^* \ll V\). }
\rv{Here, \(M^*\) represents the number of embeddings. It equals \(M\) from the Feature Extractor (\(M^*=M\)), but they serve different purposes in the context. Embedding (EmbedPatch) is learnable, whereas \textbf{naive representation} refers to the raw, high-dimensional features extracted directly from the original time series. Intuitively, each of the \(M\) patches finds a corresponding embedding from the \(V\) patch embeddings, resulting in \(M^*\) new embeddings for the original patches.}
This simple component allows SimAD to efficiently learn key information from normal samples without being overly dependent on prior knowledge.

To achieve this, we employ an improved multi-head attention. For each head \rv{\(k \in [1, 2, \ldots, U]\)}, we define the query matrix as \(\mathbf{Q}^{(i)}_k = \mathbf{N}^{(i)} \mathbf{W}^Q_k\), the key matrix as \(\mathbf{K}^{(i)}_k = \mathbf{N}^{(i)} \mathbf{W}^K_k\), and the value matrix as \rv{\(\mathbf{V}^{(i)}_k=\mathbf{E}^{(i),\prime}_k = \mathbf{W}^V_k \mathbf{E}^{(i)}_k\)}, where \rv{\(\mathbf{W}^Q_k, \mathbf{W}^K_k \in \mathbb{R}^{D \times d}\)}, \rv{\(\mathbf{W}^V_k \in \mathbb{R}^{M^* \times V}\)},\rv{\(\mathbf{E}^{(i)}=[\mathbf{E}^{(i)}_1,\mathbf{E}^{(i)}_2,\ldots,\mathbf{E}^{(i)}_U],\mathbf{E}^{(i)}_k \in \mathbb{R}^{V\times d}\)} , and \(d\) is defined as \(\left\lfloor\frac{D}{U}\right\rfloor\), which represents the dimension of the head. 
% 对于EmbedPatch Encoder的第一层而言，它的输入\(\mathbf{N}^{(1)}\)即为Feature Extractor的输出\(\mathbf{N}\)。
\rv{For the first layer of the EmbedPatch Encoder, its input \(\mathbf{N}^{(1)}\) is the output \(\mathbf{N}\) from the Feature Extractor.}
Next, we define the attention calculation between different patches in each layer as below:
\newenvironment{footequation}{\footnotesize \begin{equation}}{\end{equation}}
\begin{footequation}
    \mathbf{Z}^{(i)}_k = 
\operatorname{Attention}\left(\mathbf{Q}^{(i)}_k,\mathbf{K}^{(i)}_k,\mathbf{V}^{(i)}_k\right) = 
    \operatorname{Softmax}\left(\frac{(\mathbf{Q}^{(i)}_k \mathbf{K}^{(i)\top}_k}{\sqrt{d}}\right) \mathbf{V}^{(i)}_k.
\end{footequation}
\rv{Note that the $\mathbf{Q}^{(i)}_k$ and $\mathbf{K}^{(i)}_k$ here are generated by different parameters, so the attention scores are asymmetric.}
% 请注意，这里的\(\mathbf{Q}^{(i)}_k\)和\(\mathbf{K}^{(i)}_k\)是由不同的参数生成的，因此注意力分数是不对称的。
% 在每一层的末尾，用一个线性层聚集不同头的特征$\mathbf{Z}^{(i)}_k$，得到$\mathbf{Z}^{(i)} \ in \mathbb{R}^{ N \times D}$。
% 之后我们延用原始的$\operatorname{Add \& Norm}$操作产生下一层的输入$\mathbf{N}^{(i+1)}$。
On the top of each layer, we utilize a linear layer to aggregate features \(\mathbf{Z}^{(i)}_k\) from different heads, resulting in \(\mathbf{Z}^{(i)} \in \mathbb{R}^{ M \times D}\). 
\rv{Specifically, we first concatenate the features from \(U\) heads along the last dimension. For each \(\mathbf{Z}^{(i)}_k \in \mathbb{R}^{M \times d}\), the concatenation yields \(\mathbf{Z}^{(i),\prime} (\in \mathbb{R}^{M \times D}) = \left[ \mathbf{Z}^{(i)}_1, \ldots, \mathbf{Z}^{(i)}_U \right]\). Subsequently, this concatenated feature is aggregated via a linear layer \(\operatorname{Linear}\), resulting in \(\mathbf{Z}^{(i)} = \operatorname{Linear}(\mathbf{Z}^{(i),\prime})\).}
% 首先，我们将\(U\)个头的特征沿着最后一个维度拼接，对于每一个\(\mathbf{Z}^{(i)}_k \in \mathbb{R}^{M^* \times d}\)，拼接后得到\(\mathbf{Z}^{(i),\prime} (\in \mathbb{R}^{M^* \times D})=\left[ \mathbf{Z}^{(i)}_1,\ldots,\mathbf{Z}^{(i)}_U \right]\)。然后该特征经过一个线性层\(\operatorname{Linear}\)聚合，即\(\mathbf{Z}^{(i)} = \operatorname{Linear}(\mathbf{Z}^{(i),\prime})\)。

Subsequently, we proceed with the original ``Add \& Norm" operation to generate the input, \(\mathbf{N}^{(i+1)}\), for the next layer:
\begin{equation}
    \begin{aligned}
        \mathbf{N}^{(i) \prime} &= \operatorname{LayerNorm}(\ \mathbf{N}^{(i)} + \mathbf{Z}^{(i)} \ ),\\
        \mathbf{N}^{(i+1)} &= \operatorname{LayerNorm}(\ \mathbf{N}^{(i) \prime} + \rv{\operatorname{FFN}}(\mathbf{N}^{(i) \prime} )\ ),
    \end{aligned}
\end{equation}
\rv{where \(\operatorname{FFN}(\cdot)\) denotes a 2-layer feed-forward network (FFN), as in \cite{attention_is}, the superscript \(\mathbf{N}^\prime\) to denote intermediate variables, and the subscript \((i)\) to indicate features at different layers. }

Finally, a \rv{linear layer} is used to restore \(\hat{\mathbf{X}}\) from the last layer. 
\rv{As shown in Fig. \ref{fig:fw}(a), the final output of the EmbedPatch Encoder is processed by a linear layer, which can be expressed as \(\hat{\mathbf{X}} = \operatorname{Linear}(\mathbf{N}^{(-1)})\). Here, \(\mathbf{N}^{(-1)}\) represents the features from the last layer of the encoder.}

% 为了让模型具备长上下文的时序异常检测能力，我们需要优化SimAD的Feature Extractor与EmbedPatch Encoder的参数。
\rv{To enable the model to detect anomalies in long-context time series, we need to optimize the parameters of SimAD's Feature Extractor and EmbedPatch Encoder.}
During the training phase, the Mean Square Error (MSE) \cite{anom_trans,usad} is used to measure the difference between \(\mathbf{X}\) and \(\hat{\mathbf{X}}\), and a patch-based similarity loss is employed to ensure continuity between patches.
% (See Appendix \ref{sec:cos_sim_loss} for details). 
% 其中MSE项是确保模型在细节部分重构时序，但是这种单纯的点到点重构无法引导模型在长时间窗口中检测异常。similarity项的目的是为了弥补这一不足,它首先将\(\mathbf{X}\)通过之前的\(\operatorname{Patching}\)转换成\(\mathbf{N}\in \mathbb{R}^{M \times (P \cdot C)}\)。随后，沿着最后一个维度计算基于patch的相似性。最终的训练损失可以表示为Eq. (\ref{eq:L_rec})。
\rv{The MSE term ensures that the model reconstructs the time series accurately locally. However, this point-to-point reconstruction alone cannot guide the model to detect anomalies over long time windows. The similarity term addresses this limitation. It first converts \(\mathbf{X}\) into \(\mathbf{N} \in \mathbb{R}^{M \times (P \cdot C)}\) using the previous \(\operatorname{Patching}\) operation. Then, it calculates patch-based similarity along the last dimension. The final training loss is given by Eq. (\ref{eq:L_rec}).}

During the testing phase, SimAD utilizes \(\mathcal{L}_{rec}\) to calculate anomaly scores and detect anomalies. 
\begin{equation}
    \mathcal{L}_{rec} = \operatorname{MSE}(\hat{\mathbf{X}}, \mathbf{X}) + (1 - \operatorname{Similarity}( \hat{\mathbf{X}}, \mathbf{X} )),
    \label{eq:L_rec}
\end{equation}
where cosine similarity is adopted as a similarity measure. 

\subsubsection{ContrastFusion Module}
Although the EmbedPatch Encoder enables SimAD to memorize normal samples, it does not substantially enhance the discrimination between normal and abnormal data. To mitigate this limitation, the ContrastFusion module is designed to employ contrastive learning to enlarge the feature distance between normal and abnormal data' features.

% 为了构建对比学习中的负样本，且不引入过多的先验，我们采用最简单高斯噪声生成负样本。
% 其中$\mathbf{J}$从标准高斯分布中采样，$\alpha$是用于控制噪声的水平。

% 受到之前去噪自编码器的启发，我们最终的目标方程中增加去噪损失。

To generate negative samples for contrastive learning without introducing prior, we adopt the simplest method by using Gaussian noise \cite{amsl,ada_mem_bls}. Specifically, we use the equation \(\rv{\mathbf{X}^{-}} = \mathbf{X} + \alpha\cdot\mathbf{J}\), where \(\mathbf{J}\) is sampled from a Gaussian distribution, and \(\alpha\) controls the level of noise. 
% 这是一种常见的时序数据增强方法。
\rv{This is a common time series data augmentation method \cite{amsl,ada_mem_bls}.}
% 此后，我们用带下标n的变量表示negative samples或特征。
From then on, we use variables with subscript ``\(-\)" to represent negative samples or features.
Inspired by denoising autoencoders \cite{dae}, we incorporate a denoising loss into the final objective function. 
% 同样的，为了在去噪过程中，引导模型关注上下文的时序关联，我们额外增加了相似度项。
\rv{Similarly, to guide the model to focus on contextual temporal associations during denoising, we added an extra similarity term.}
\rv{\begin{equation}
     \mathcal{L}_{denoise} =  \operatorname{MSE}(\hat{\mathbf{X}}^-, \mathbf{X}^+) + (1 - \operatorname{Similarity}( \hat{\mathbf{X}}^-, \mathbf{X}^+ )\ ).
     \label{eq:L_denoise}
\end{equation}
}
\rv{$\mathbf{X}^+$ indicates the branch opposite to the negative sample $\mathbf{X}^-$, emphasizing their opposition and being differentiated by colors/symbols in Fig. \ref{fig:fw}(a) for convenience and consistency. Keep in mind that $\mathbf{X}^+=\mathbf{X}$ in this case.
The two branches of ContrastFusion, which are optimized without regard to reconstruction, share structural similarities with previous image self-supervised techniques such as SimCLR \cite{simclr} and SimSiam \cite{SimSiam}. Therefore, our methodology can also be considered a self-supervised (and/or unsupervised \cite{bls2}) learning strategy.}
% 为了方便理解，并与图\ref{fig:fw}(a)保持一致，这里我们使用\(\mathbf{X}^+\)表示与负样本\(\mathbf{X}^-\)相对的分支，说明它们的对立性；并在图\ref{fig:fw}(a)中使用不同的颜色和符号表示。请注意实际上，这里\(\mathbf{X}^+=\mathbf{X}\)。ContrastFusion的两个分支视作独立于重构的优化，在结构上与过往的图像自监督方法SimCLR\cite{simclr}，SimSiam\cite{SimSiam}相似。 因此我们的方法也可以视作一种自监督学习方法（同样属于无监督\cite{bls2}）。
% 我们将正样本$\mathbf{X}$与负样本$\mathbf{X}_n$同时依次送入Feature Extractor和EmbedPatch Encoder得到隐藏特征表征$\mathbf{N}$和$\mathbf{N}_n$。为了有效且快速学习内在的不变特征，我们采用过往对比学习方法类似的技巧，设计一个投影头$\mathscr{P}(\cdot)$。

To learn invariant features, we feed positive samples \rv{\(\mathbf{X}^+\)} and negative samples \rv{\(\mathbf{X}^-\)} into the Feature Extractor and EmbedPatch Encoder, resulting in the acquisition of representations \rv{\(\mathbf{N}^+\)} and \rv{\(\mathbf{N}^-\)}. 
\rv{Previous self-supervised works \cite{simclr,SimSiam,sim_net,ada_mem_bls} have shown that constructing different views through data augmentation can guide models to learn invariant features.}
% 过去的自监督工作\cite{simclr,SimSiam,sim_net,ada_mem_bls}均表明，通过数据增强构造不同视图可以引导模型学习不变特征。
For this purpose, we utilize a projection head \(\mathscr{P}(\cdot)\), akin to prior contrastive learning methods, to facilitate the acquisition of meaningful and discriminative representations.
\rv{
\begin{equation}
    \mathbf{H}^+ = \mathscr{P}(\mathbf{N}^+) = \operatorname{Linear}(\operatorname{ReLU}(\operatorname{Linear}(\mathbf{N}^+))),
\end{equation}
}
where \(\operatorname{ReLU}\) is the activation function. \rv{We derive low-dimensional representations \(\mathbf{H}^+\) and \(\mathbf{H}^-\) for \(\mathbf{N}^+\) and \(\mathbf{N}^-\) respectively. }To design an asymmetric feature contrastive loss, we utilize gradient stopping.

\newenvironment{smallequation}{\scriptsize\begin{equation}}{\end{equation}}
\begin{smallequation}
\begin{aligned}
   \mathcal{L}_{{cont}} &= \operatorname{MSE}(\mathbf{H}^+, \operatorname{StopGrad}(\mathbf{H}^-)) + (1 - \operatorname{Similarity}(\mathbf{H}^+, \operatorname{StopGrad}(\mathbf{H}^-))) \\
    & + \operatorname{MSE}(\mathbf{H}^-, \operatorname{StopGrad}(\mathbf{H}^+)) + (1 - \operatorname{Similarity}(\mathbf{H}^-, \operatorname{StopGrad}(\mathbf{H}^+))).
\end{aligned}
\label{eq:L_cont}
\end{smallequation}

\subsubsection{Joint optimization for Training}
As described above, SimAD aims to optimize the reconstruction of positive samples, denoise noisy samples, and
amplify the feature differences between positive and negative samples (see in Fig. \ref{fig:fw}(b)). During the initial stages of training, the model focuses primarily on reconstruction and denoising. 
Since the contrastive loss may introduce additional training complexity, the weight of contrastive loss incrementally rises 
in the initial \(N_{\text{warm-up}}\) iterations until it reaches the maximum value of \(\beta_{\text{max}}\):
% \begin{equation}
    \(\beta_i= 
        \min\{\frac{i+1}{N_{warm-up}},\beta_{max}\}\).
% \end{equation}
% 最终的目标函数为

The overall objective function combines the reconstruction loss, denoising loss, and contrastive loss, as below:
\begin{equation}
    \mathcal{L} = \mathcal{L}_{rec}+ \mathcal{L}_{denoise}- \beta \mathcal{L}_{cont}.
\end{equation}
% \section{Pseudo-code of SimAD}\label{sec:code}
% 算法\ref{alg:sim_ad}表示了SimAD的工作流程。
Algorithm \ref{alg:sim_ad} shows the training workflow of SimAD, highlighting the simplicity of our method in design and implementation. 
\begin{algorithm}[]
\caption{The Pseudo-Code for Training SimAD.}
\begin{lstlisting}
# J: i.i.d Gaussian noise
# FE: Feature Extractor
# Enc: EmbedPatch Encoder
# CF: ContrastFusion
# SimAD: Combination of FE, Enc, and CF
# R: Rearrange data
for x in data_loader:
    x1, x2 = x, x + J
    x_out1, sim1 = SimAD(x1)  # SimAD Processing
    x_out2, sim2 = SimAD(x2) 
    # Reconstructs and projects features
    x_patch = R(x)
    rec_loss = loss_func(x_out1,x_patch)
    denoise_loss = loss_func(x_out2,x_patch)
    sim_loss = loss_func(sim1,sim2.detach()) + loss_func(sim2,sim1.detach())
    loss = rec_loss + denoise_loss - sim_loss * warmup
    loss.backward()
    update(SimAD)

# loss function
def loss_func(s1,s2):
    return l2_loss(s1,s2) + (1-cos_loss(s1,s2)).mean()
\end{lstlisting}
\label{alg:sim_ad}
\end{algorithm}
\subsection{Baselines}
\subsubsection{Inference: Anomaly score}
To calculate the anomaly score for query samples, we use the following equation:
\begin{equation}
\mathcal{L}_{score} = \operatorname{MSE}(\hat{\mathbf{X}}, \mathbf{X}) + (1 - \operatorname{Similarity}(\hat{\mathbf{X}}, \mathbf{X})).   
\end{equation}
\rv{Note that the first term (MSE) has a length of \(T\) time points, while the second term (similarity) has a length of \(M\) patches. To match the length, each patch in the similarity term is replicated \(P\) times to reach \(T\) time points in the implementation.}

As shown in Fig. \ref{fig:fw}(c), this process involves inputting time series data into SimAD and performing feature extraction and representation to obtain the final feature $\mathbf{Z}^{(L)}$. This feature is then fed into a linear layer to reconstruct the data as closely as possible to the original: $\hat{X} = \operatorname{Linear}(\mathbf{Z}^{(L)})$. We then compute the MSE and cosine similarity between the reconstructed and original data. 
In inference stage, the ContrastFusion module is not used, 
eliminating the need to generate negative samples or compute similarities between positive and negative samples.

\subsection{Improving Affiliation Metrics}
Recent studies \cite{metric1, metric2, metric3} have highlighted the limitations of conventional metrics used in time series anomaly detection. The affiliation metric \cite{aff} is a parameter-free method that has shown promising performance. However, based on experiments and theoretical analysis (\rv{see Appendix \ref{sec:ana_uaff_naff}}), the Affiliation precision (Aff-Pre) tends to approach 0.5, while the Affiliation recall (Aff-Rec) tends to approach 0.99 when confronting with random anomaly scores.
This leads to an Affiliation F1 score (Aff-F1) of approximately 0.7, suggesting that the metric's discriminatory capability is insufficient.
To address this limitation, we introduce \textbf{Unbiased Affiliation (UAff)} and \textbf{Normalized Affiliation (NAff)} metrics. Both of them exhibit superior discriminatory ability and provide a more accurate reflection of an algorithm's performance.

Experimental findings (Table \ref{tab:data_set}) indicate that the Affiliation precision (Aff-Pre) varies across different datasets but consistently tends to approach 0.5, which can be interpreted as a dataset-specific bias, denoted by \(\textbf{Aff-Pre}_{bias}\). 
% 同时，高精度的模型并没有受到鼓励，低精度模型的惩罚不足。
Meanwhile, it is noteworthy that there has been insufficient encouragement for high-precision models, while the penalties for low-precision models are inadequate.
In this context, we design Unbiased Affiliation precision (UAff-Pre) and Unbiased Affiliation F1 (UAff-F1), both of which can alleviate the dataset-specific bias and offer a more equitable assessment of the algorithm's precision performance.
\begin{equation}
    \textbf{UAff-Pre} =
        \frac{\textbf{Aff-Pre} - \textbf{Aff-Pre}_{bias}}{1 - \textbf{Aff-Pre}_{bias}}
\end{equation}
\begin{equation}\label{eq:aff_f1}
    \textbf{UAff-F1} = \frac{2 \cdot |\textbf{UAff-Pre}| \cdot \textbf{Aff-Rec}}{ |\textbf{UAff-Pre}| + \textbf{Aff-Rec} } \cdot (-1)^{|\textbf{UAff-Pre}<0|}.
\end{equation}
where \textbf{UAff-F1} is negative when \(\textbf{UAff-Pre}<0 \).
Although \(\textbf{UAff-Pre}\) denotes an improvement over its original version and surpasses \(\textbf{UAff-F1}\), it does introduce additional parameters, thereby deviating from its parameter-free nature. Since the majority of real-world datasets exhibit \(\textbf{Aff-Pre}_{bias} \leq 0.55\), we further propose a modified version of NAff metrics. The computation follows a similar process, with the exception that we set \(\textbf{Aff-Pre}_{bias}=\beta\) for constant while maintaining other aspects unchanged. In other words, \(\textbf{NAff-Pre}\) can be calculated as \( \frac{\textbf{Aff-Pre} - \beta}{1 - \beta} \), and \(\textbf{NAff-F1}\) is calculated similarly to \(\textbf{UAff-F1}\). \rv{The mathematical and experimental analyses of the metrics are in the Sec. \ref{sec:ana_uaff_naff}}.

\section{Experiments}
We evaluate the effectiveness of our proposed SimAD by conducting extensive comparison experiments against state-of-the-art competing methods on six real-world multivariable datasets: MSL, SMAP, PSM, SWaT, WADI, and Swan. Additionally, we utilize a univariate dataset UCR \cite{UCR}. The details of the above six multivariable datasets are in Table \ref{tab:data_set}. % in Appendix \ref{sec:data}. 
%Besides, the implementation and pseudo-code of the algorithm can be found in Appendix \ref{sec:code}.

\begin{table}[htb]
	\centering
    % \begin{subtable}[t]{0.585\textwidth}
        \caption{Details of benchmark datasets. }
    \resizebox{1.\linewidth}{!}{
	\begin{tabular}{c|c|c|c|c|c|c}
		\toprule
		Dataset  & \#Training & \#Test (Labeled) & Dimension & Anomaly ratio (\%) & Bias & Ideal Bias \\
		\midrule
    MSL   & 58317 & 73729 & 55    & 10.5  & 51.34  & 50.55  \\
    SMAP  & 135183 & 427617 & 25    & 12.8  & 51.48  & 50.82  \\
    SWaT & 99000 & 89984 & 26 & 12.2 & 52.94  & 50.74  \\
    % WADI & 1048571 & 172801 & 123 & 5.99& 54.81  & 50.18  \\
    WADI & 241921 & 34561 & 123 & 5.74 & 54.81 & 50.16\\
    PSM & 132481 & 87841 & 25 & 27.8  & 53.17  & 53.86  \\
    Swan & 60000 & 60000 & 38 & 32.6 & 54.83  & 55.31  \\
		\bottomrule
	\end{tabular}
}
	\label{tab:data_set}
    % \end{subtable}
\end{table}

We employ the F1 without point adjustment, Aff-F1, and the improved UAff-F1 and NAff-F1 for performance evaluation. 
We exclude the point-adjusted F1 score due to its acknowledged potential for false improvements. It is noted that although the classical F1 score is not optimal for time series data, it can still be employed for evaluation purposes \cite{metric1}, as detailed in Appendix \ref{sec:flaw}. Furthermore, we present an analysis of VUS \cite{VUS} scores (Table \ref{tab:compa_vus} in Appendix) for all the datasets. 
\rv{In TSAD, the F1 score evaluates model performance by balancing precision (Prec) and recall (Rec). It is defined as:
\begin{equation}
    \text{Prec} = \frac{\text{TP}}{\text{TP} + \text{FP}}, \ \text{Rec} = \frac{\text{TP}}{\text{TP} + \text{FN}}, \ \text{F1} = 2 \times \frac{\text{Prec} \times \text{Rec}}{\text{Prec} + \text{Rec}}.
\end{equation}
Here, TP, FP, and FN represent true positives, false positives, and false negatives. The F1 score ranges from 0 to 1, with higher values indicating better anomaly detection performance. 
Due to space limitations and the need for conciseness, the calculation of Aff-F1 and VUS refer to the original papers \cite{aff,VUS}.}
% 受限于篇幅与内容精简考虑，Aff-F1的计算参考原始论文\cite{aff}。

\rv{We extensively discuss UAff and NAff in Sec.~\ref{sec:ana_uaff_naff}, comparing them with other metrics, and summarize the limitations and advantages of existing metrics in detail.}

\subsection{Baselines}\label{sec:baselines}
Our model is compared against 20 baselines, which include machine learning-based models like LOF, Isolation Forest (IForest) and PCA; deep learning-enhanced models like Deep SVDD \cite{deep_svdd} and Deep Isolation Forest (Deep IF) \cite{deep_if}; reconstruction-based models like USAD \cite{usad}, TCN-ED \cite{tcn_ed}, AdaMemBLS \cite{ada_mem_bls} and NPSR \cite{NPSR}; prediction-based methods like TimesNet \cite{timesnet}, and M2N2 \cite{M2N2}; Transformer-based models such as Anomaly Transformer (AnomTrans) \cite{anom_trans} and TranAD \cite{trans_ad}; contrastive learning-based models like COUTA \cite{couta}, NCAD \cite{ncad}, and DCdetector \cite{dcdetector}; diffusion-based model D3R \cite{d3r}; and LLM-based GPT2-Adapter \cite{gpt2adpt}.
\subsection{Quantified Comparisons}
\begin{table*}[htbp]
  \centering
  \setlength\tabcolsep{2pt}
  \caption{Comparison results. All results are in \%, the best in \textbf{Bold}, and the second in \uline{underlined}.}
  \resizebox{1.\linewidth}{!}{
\begin{tabular}{c|ccccccc|ccccccc|ccccccc}
\toprule
\textbf{Datasets} & \multicolumn{7}{c|}{\textbf{MSL}}                     & \multicolumn{7}{c|}{\textbf{SMAP}}                    & \multicolumn{7}{c}{\textbf{SWaT}} \\
\midrule
\textbf{Methods} & \textbf{F1} & \textbf{AUC} & \textbf{Aff-Pre} & \textbf{Aff-Rec} & \textbf{Aff-F1} & \textbf{UAff-F1} & \textbf{NAff-F1} & \textbf{F1} & \textbf{AUC} & \textbf{Aff-Pre} & \textbf{Aff-Rec} & \textbf{Aff-F1} & \textbf{UAff-F1} & \textbf{NAff-F1} & \textbf{F1} & \textbf{AUC} & \textbf{Aff-Pre} & \textbf{Aff-Rec} & \textbf{Aff-F1} & \textbf{UAff-F1} & \textbf{NAff-F1} \\
\midrule
Random & 18.60  & 50.01  & 51.34  & \uline{99.99 } & 67.84  & 0.00  & 5.20  & 22.06  & 50.18  & 51.48  & \textbf{100.0 } & 67.97  & 0.00  & 5.74  & 21.06  & 50.10  & 52.94  & 99.99  & 69.23  & 0.00  & 11.09  \\
LOF   & 19.36  & 55.75  & 53.57  & 88.90  & 66.86  & 8.74  & 13.22  & 23.64  & 62.39  & 46.00  & 81.26  & 58.74  & -19.84  & -14.58  & 58.22  & 84.59  & 99.87  & 2.70  & 5.25  & 5.25  & 5.25  \\
IForest & 10.90  & 59.09  & 55.09  & 97.63  & 70.44  & 14.31  & 18.45  & 23.78  & 61.15  & 39.45  & 70.23  & 50.52  & -36.64  & -32.45  & 38.52  & 83.80  & \textbf{100.0 } & 2.71  & 5.27  & 5.27  & 5.27  \\
PCA   & 10.33  & 53.25  & 55.50  & 97.44  & 70.72  & 15.73  & 19.77  & 22.96  & 58.70  & 39.67  & 70.31  & 50.72  & -36.15  & -31.93  & 26.05  & 81.85  & 99.96  & 2.70  & 5.25  & 5.25  & 5.25  \\
Deep SVDD & \uline{24.30 } & 61.80  & 52.67  & 99.93  & 68.98  & 10.53  & 10.13  & 21.28  & 61.14  & 44.98  & 88.45  & 59.64  & -17.60  & -18.02  & 22.57  & 86.81  & 53.58  & \uline{99.99 } & 69.77  & 13.75  & 13.37  \\
USAD  & 22.50  & 57.12  & 51.90  & 97.29  & 67.69  & 7.73  & 7.31  & 16.57  & \uline{62.79 } & 39.56  & 69.48  & 50.41  & -31.78  & -32.12  & 21.70  & 88.66  & 53.00  & 100.0  & 69.28  & 11.71  & 11.31  \\
TCN-ED & 19.66  & 51.14  & 51.44  & 99.99  & 67.93  & 6.04  & 5.61  & 22.90  & 58.91  & 51.39  & 99.98  & 67.89  & 5.86  & 5.42  & 21.65  & 89.23  & 52.98  & 100.0  & 69.26  & 11.64  & 11.24  \\
COUTA & 20.88  & 55.59  & 51.24  & 99.65  & 67.68  & 5.29  & 4.85  & 22.69  & 58.74  & 51.48  & \textbf{100.0 } & 67.97  & 6.18  & 5.74  & 47.65  & 75.54  & \uline{79.98 } & 33.20  & 46.92  & 42.76  & 42.73  \\
TranAD & 22.81  & 50.03  & 52.82  & 99.34  & 68.97  & 11.07  & 10.66  & 22.74  & 59.74  & 39.41  & 70.30  & 50.51  & -32.21  & -32.56  & 25.50  & 88.90  & 55.65  & 99.66  & 71.42  & 20.64  & 20.30  \\
NCAD  & 22.05  & 60.20  & \uline{56.38 } & 83.30  & 67.25  & \textbf{22.46 } & 22.14  & 23.09  & 53.45  & 51.88  & \uline{99.99 } & 68.32  & 7.67  & 7.25  & 68.72  & 82.92  & 65.01  & 85.58  & 73.89  & 44.64  & 44.46  \\
Deep IF & 19.11  & 55.94  & 51.45  & \textbf{100.0 } & 67.94  & 6.08  & 5.64  & \uline{29.14 } & 60.09  & 53.75  & 98.67  & 69.59  & 14.31  & 13.93  & 21.65  & 89.52  & 52.98  & \textbf{100.0 } & 69.26  & 11.64  & 11.24  \\
AnomTrans & 18.39  & 52.61  & 50.21  & 99.83  & 66.81  & -4.53  & 0.83  & 16.06  & 52.18  & \uline{55.84 } & 99.11  & \uline{71.44 } & \uline{16.49 } & \uline{20.90 } & 23.44  & 80.80  & 50.24  & 99.49  & 66.77  & -10.83  & 0.96  \\
TimesNet & 21.24  & 57.18  & 51.12  & 99.95  & 67.64  & 4.81  & 4.36  & 24.37  & 53.73  & 49.50  & 99.86  & 66.19  & -1.52  & -1.99  & 21.66  & 73.24  & 57.84  & 93.77  & 71.55  & 27.16  & 26.86  \\
DCdetector & 11.62  & 50.31  & 51.96  & 97.77  & 67.85  & 2.52  & 7.52  & 26.56  & 58.50  & 55.55  & 99.78  & 71.37  & 15.48  & 19.97  & 23.24  & 52.78  & 52.63  & 98.30  & 68.56  & -1.28  & 10.00  \\
D3R   & 23.98  & \textbf{63.00 } & 53.61  & 99.97  & 69.79  & 8.92  & 7.65  & 22.71  & 54.56  & 51.43  & \textbf{100.0 } & 67.92  & -0.21  & 0.07  & 45.89  & 79.95  & 61.47  & 78.52  & 68.96  & 29.47  & 36.02  \\
GPT2-Adapter & 13.72  & 52.03  & 53.31  & 97.01  & 68.81  & 7.80  & 6.75  & 24.12  & 55.48  & 53.28  & 99.84  & 69.48  & 7.18  & 3.67  & 22.30  & 52.30  & 52.51  & 98.13  & 68.41  & -1.79  & 0.07  \\
NPSR  & 23.72  & 61.16  & 52.05  & 99.81  & 68.42  & 2.89  & 7.88  & 22.68  & 61.26  & 51.46  & \textbf{100.0 } & 67.95  & -0.06  & 5.68  & \uline{76.88 } & \uline{90.18 } & 71.21  & 81.16  & \uline{75.86 } & \uline{52.54 } & \uline{55.73 } \\
\rv{M2N2}  & 21.76  & 59.15  & 51.38  & \textbf{100.00 } & 67.88  & 0.17  & 5.36  & 22.68  & 54.05  & 51.46  & \textbf{100.00 } & 67.95  & 0.52  & 5.69  & 38.50  & 67.79  & 59.15  & 96.83  & 73.44  & 27.54  & 30.77  \\
\rv{AdaMemBLS} & 19.11  & 51.78  & \textbf{57.07 } & 95.37  & \uline{71.41 } & \uline{20.99 } & \textbf{24.64 } & 26.93  & 53.66  & 53.30  & 99.80  & 69.49  & 7.24  & 12.38  & 74.10  & 81.61  & 53.57  & \textbf{100.00 } & 69.76  & 2.65  & 13.32  \\
\midrule
Ours  & \textbf{30.02 } & \uline{62.70 } & 56.30  & 99.76  & \textbf{71.98 } & 18.50  & \uline{22.37 } & \textbf{29.39 } & \textbf{65.46 } & \textbf{56.84 } & 99.82  & \textbf{72.44 } & \textbf{19.91 } & \textbf{24.07 } & \textbf{82.03 } & \textbf{90.31 } & 78.46  & 80.88  & \textbf{79.65 } & \textbf{64.93 } & \textbf{66.82 } \\
\midrule
\textbf{Datasets} & \multicolumn{7}{c|}{\textbf{WADI}}                    & \multicolumn{7}{c|}{\textbf{PSM}}                     & \multicolumn{7}{c}{\textbf{NIPS-TS-Swan}} \\
\midrule
\textbf{Methods} & \textbf{F1} & \textbf{AUC} & \textbf{Aff-Pre} & \textbf{Aff-Rec} & \textbf{Aff-F1} & \textbf{UAff-F1} & \textbf{NAff-F1} & \textbf{F1} & \textbf{AUC} & \textbf{Aff-Pre} & \textbf{Aff-Rec} & \textbf{Aff-F1} & \textbf{UAff-F1} & \textbf{NAff-F1} & \textbf{F1} & \textbf{AUC} & \textbf{Aff-Pre} & \textbf{Aff-Rec} & \textbf{Aff-F1} & \textbf{UAff-F1} & \textbf{NAff-F1} \\
\midrule
Random & 10.79  & 50.03  & 54.81  & 99.95  & 70.79  & 0.00  & 17.54  & 40.93  & 50.16  & 53.17  & 99.95  & 69.41  & 0.00  & 11.91  & 46.25  & 50.13  & 54.83  & 96.82  & 70.01  & 0.00  & 17.56  \\
LOF   & 10.58  & 65.44  & 51.76  & 25.72  & 34.36  & -10.67  & 6.20  & 23.91  & \textbf{81.28 } & 77.47  & 52.13  & 62.32  & 52.01  & 53.50  & 21.37  & 77.89  & 82.01  & 11.14  & 19.62  & 18.81  & 18.98  \\
IForest & 30.88  & 74.84  & 57.68  & 82.53  & 67.90  & 11.82  & 25.91  & 38.52  & 75.12  & 72.97  & 46.46  & 56.77  & 44.27  & 46.20  & 46.18  & \uline{85.12 } & \textbf{94.11 } & 16.31  & 27.81  & 27.47  & 27.54  \\
PCA   & 34.30  & 50.53  & 48.10  & 52.07  & 50.01  & -23.09  & -7.08  & 43.20  & 72.06  & 77.86  & 79.65  & \textbf{78.74 } & \textbf{63.44 } & \textbf{65.57 } & 49.18  & 72.88  & \uline{92.96 } & 16.03  & 27.35  & 26.95  & 27.02  \\
Deep SVDD & 7.34  & 59.47  & 50.06  & 55.79  & 52.77  & 0.72  & 0.24  & 44.46  & \uline{78.09 } & 53.92  & \uline{99.98 } & 70.06  & 14.93  & 14.56  & \uline{55.99 } & 75.32  & 56.84  & 89.93  & 69.66  & 24.07  & 23.76  \\
USAD  & 10.86  & 53.61  & 54.92  & \textbf{100.0 } & 70.90  & 18.27  & 17.91  & 47.90  & 64.53  & 55.30  & 98.71  & 70.88  & 19.48  & 19.14  & 52.63  & 66.76  & 55.26  & 68.62  & 61.22  & 18.57  & 18.25  \\
TCN-ED & 10.86  & 53.01  & 54.92  & \textbf{100.0 } & 70.90  & 18.27  & 17.91  & 43.46  & 63.97  & 53.16  & \textbf{100.0 } & 69.42  & 12.29  & 11.89  & 49.22  & 71.37  & 55.10  & 99.92  & 71.03  & 18.87  & 18.52  \\
COUTA & 26.92  & 50.52  & \textbf{99.18 } & 23.08  & 37.44  & \uline{37.38 } & 37.38  & 48.21  & 69.36  & 57.98  & 99.30  & 73.21  & 27.80  & 27.50  & 47.19  & 71.38  & 45.07  & 86.50  & 66.54  & 15.24  & 14.87  \\
TranAD & 11.78  & 52.60  & 55.91  & 91.96  & 69.54  & 21.27  & 20.94  & 43.20  & 65.22  & 60.88  & 92.09  & 73.30  & 35.44  & 35.19  & 51.93  & 70.00  & 57.86  & 90.18  & 70.49  & 27.07  & 26.78  \\
NCAD  & 10.87  & 62.84  & 54.99  & \textbf{100.0 } & 70.96  & 18.49  & 18.14  & 46.61  & 61.76  & \uline{77.79 } & 53.08  & 63.10  & \uline{54.35 } & \uline{54.30 } & 37.57  & 52.24  & 54.32  & 94.06  & 68.87  & 16.20  & 15.83  \\
Deep IF & 10.86  & 51.43  & 54.92  & \textbf{100.0 } & 70.90  & 18.27  & 17.91  & 43.47  & 69.06  & 53.15  & \textbf{100.0 } & 69.41  & 12.24  & 11.84  & 49.19  & 73.30  & 55.10  & 99.98  & 71.05  & 18.86  & 18.51  \\
AnomTrans & 11.24  & 52.98  & 51.43  & 95.15  & 66.77  & -13.84  & 5.56  & 39.81  & 52.18  & 52.78  & 96.07  & 68.13  & -1.65  & 10.51  & 44.86  & 54.62  & 55.21  & 92.36  & 69.11  & 1.68  & 18.73  \\
TimesNet & 17.65  & 65.08  & 65.25  & 39.30  & 49.05  & 34.45  & 34.34  & 43.60  & 58.74  & \textbf{77.84 } & 67.23  & 72.15  & \uline{60.97 } & \uline{60.91 } & 43.16  & 53.62  & 60.75  & 81.29  & 69.53  & \uline{34.23 } & \uline{33.99 } \\
DCdetector & 11.33  & 50.12  & 59.72  & 94.64  & 73.23  & 19.51  & 32.26  & 22.72  & 50.38  & 53.05  & 94.98  & 68.08  & -0.49  & 11.47  & 48.83  & 50.35  & 55.08  & 99.82  & 70.99  & 1.11  & 18.45  \\
D3R   & 12.86  & 51.39  & 56.47  & \uline{99.98 } & 72.17  & 7.09  & 2.46  & 44.24  & 60.64  & 53.84  & \textbf{100.0 } & 69.99  & 2.82  & 3.93  & 49.45  & 62.54  & 57.70  & 96.06  & \uline{72.10 } & 11.93  & 9.98  \\
GPT2-Adapter & 10.68  & 51.21  & 55.62  & 98.29  & 71.04  & 3.55  & -0.27  & 35.24  & 51.23  & 54.33  & 95.16  & 69.17  & 4.84  & 2.18  & 45.23  & 58.77  & 61.17  & 69.55  & 65.09  & 23.36  & 18.28  \\
NPSR  & \uline{50.67 } & \uline{80.19 } & 65.15  & 90.45  & \uline{75.74 } & 36.53  & \uline{45.39 } & \uline{51.02 } & 70.67  & 54.71  & 99.37  & 70.56  & 6.29  & 17.14  & 49.30  & 62.61  & 55.13  & \uline{99.99 } & 71.07  & 1.32  & 18.61  \\
\rv{M2N2}  & 13.55  & 55.68  & 54.92  & \textbf{100.00 } & 70.90  & 13.73  & 17.93  & 47.71  & 54.91  & 55.43  & 94.36  & 69.84  & 15.45  & 19.48  & 38.40  & 70.12  & 60.94  & 93.68  & \textbf{73.84 } & 32.60  & 35.47  \\
\rv{AdaMemBLS} & 29.09  & 53.85  & 54.95  & \textbf{100.00 } & 70.93  & 0.63  & 18.01  & 46.69  & 63.26  & 58.22  & 92.65  & 71.51  & 19.32  & 27.92  & 49.35  & 51.27  & 55.14  & \textbf{100.00 } & 71.09  & 1.38  & 18.66  \\
\midrule
Ours  & \textbf{63.98 } & \textbf{87.97 } & \uline{82.36 } & 92.81  & \textbf{87.27 } & \textbf{73.59 } & \textbf{76.26 } & \textbf{52.07 } & 71.99  & 62.46  & 91.76  & \uline{74.33 } & 32.63  & 39.20  & \textbf{71.23 } & \textbf{85.17 } & 76.17  & 53.75  & 63.03  & \textbf{50.29 } & \textbf{53.04 } \\
\bottomrule
\end{tabular}%

    }
  \label{tab:comparison}%
\end{table*}%

Table \ref{tab:comparison} shows comparison results between our proposed SimAD and baselines on six real-world datasets. It is obvious that SimAD performs superior to other models, particularly on datasets SWaT, WADI, and Swan. Specifically, compared to the best baseline on each dataset, SimAD exhibits an improvements of \textbf{5.15\%} (76.88\%\(\to\)82.03\%), \textbf{13.31\% }(50.67\%\(\to\)63.98\%), and \textbf{15.24\%} (55.99\%\(\to\)71.23\%) in F1 score. In contrast, the NAff-F1 score exhibits absolute improvements of \textbf{11.09\%} (55.73\%\(\to\)66.82\%), \textbf{30.87\%} (45.39\%\(\to\)76.26\%), and\textbf{ 19.05\%} (33.99\%\(\to\)53.04\%) respectively.
In addition, SimAD has shown slight superiority over other models on datasets MSL and SMAP. On these datasets, SimAD achieved an absolute improvement of 1.26\% and 1.00\% in Aff-F1, respectively, along with higher scores in other metrics as well.

\begin{figure}[ht]
        \centering
        % \vspace{-\baselineskip}
\includegraphics[width=1.0\linewidth]{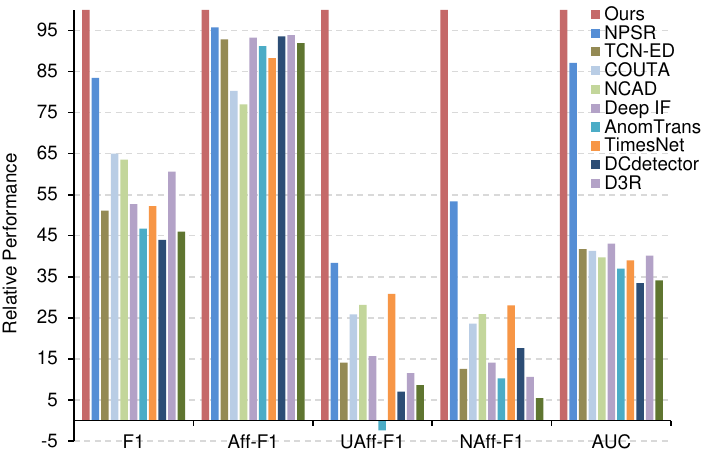}
        \caption{Comparison analysis of relative performance.}
        \label{sfig:comparison}
        % \vspace{-\baselineskip}
\end{figure}

 \begin{table}%{rh}{0.5\textwidth}
  \centering
  % \vspace{-\baselineskip}
  \setlength\tabcolsep{2pt}
  \caption{The average performance and ranking of different algorithms. \textbf{RK.} denotes the ranking.}
    \resizebox{1\linewidth}{!}{
    % Table generated by Excel2LaTeX from sheet 'WWW'
\begin{tabular}{c|cc|cc|cc|cc|cc|cc|c}
\toprule
\textbf{Avg.} & \textbf{F1} & \textbf{RK.} & \textbf{Aff-F1} & \textbf{RK.} & \textbf{UAff-F1} & \textbf{RK.} & \textbf{NAff-F1} & \textbf{RK.} & \textbf{AUC} & \textbf{RK.} & \textbf{V\_PR} & \textbf{RK.} & \textbf{Avg. RK.} \\
\midrule
Random & 26.62  & 16    & 69.21  & 9     & 0.00  & 19    & 11.51  & 15    & 50.10  & 20    & 27.71  & 18    & 16.17  \\
LOF   & 26.18  & 17    & 41.19  & 20    & 9.05  & 13    & 13.76  & 10    & 71.23  & 3     & 39.75  & 9     & 12.00  \\
IForest & 31.46  & 7     & 46.45  & 19    & 11.09  & 11    & 15.15  & 9     & 73.19  & 2     & 43.39  & 7     & 9.17  \\
PCA   & 31.00  & 8     & 47.13  & 18    & 8.69  & 14    & 13.10  & 13    & 64.88  & 8     & 37.84  & 13    & 12.33  \\
Deep SVDD & 29.32  & 11    & 65.15  & 15    & 7.73  & 15    & 7.34  & 18    & 70.44  & 5     & 50.13  & 3     & 11.17  \\
USAD  & 28.69  & 13    & 65.06  & 16    & 7.33  & 17    & 6.97  & 19    & 65.58  & 7     & 51.95  & 2     & 12.33  \\
TCN-ED & 27.96  & 15    & 69.41  & 8     & 12.16  & 10    & 11.76  & 14    & 64.61  & 9     & 46.07  & 6     & 10.33  \\
COUTA & 35.59  & 4     & 59.96  & 17    & 22.44  & 4     & 22.18  & 5     & 63.52  & 11    & 36.75  & 14    & 9.17  \\
TranAD & 29.66  & 10    & 67.37  & 13    & 13.88  & 8     & 13.55  & 11    & 64.41  & 10    & 47.66  & 5     & 9.50  \\
NCAD  & 34.82  & 5     & 68.73  & 10    & 27.30  & 2     & 27.02  & 2     & 62.24  & 12    & 38.21  & 12    & 7.17  \\
Deep IF & 28.90  & 12    & 69.69  & 7     & 13.57  & 9     & 13.18  & 12    & 66.56  & 6     & 39.69  & 10    & 9.33  \\
AnomTrans & 25.63  & 18    & 68.17  & 12    & -2.11  & 20    & 9.58  & 17    & 57.56  & 17    & 26.49  & 19    & 17.17  \\
TimesNet & 28.61  & 14    & 66.02  & 14    & 26.68  & 3     & 26.41  & 3     & 60.27  & 15    & 36.47  & 15    & 10.67  \\
DCdetector & 24.05  & 20    & 70.01  & 6     & 6.14  & 18    & 16.61  & 8     & 52.07  & 19    & 28.33  & 17    & 14.67  \\
D3R   & 33.19  & 6     & 70.16  & 5     & 10.00  & 12    & 10.02  & 16    & 62.01  & 13    & 39.59  & 11    & 10.50  \\
GPT2-Adapter & 25.22  & 19    & 68.67  & 11    & 7.49  & 16    & 5.11  & 20    & 53.51  & 18    & 28.33  & 16    & 16.67  \\
NPSR  & 45.71  & 2     & 71.60  & 2     & 16.58  & 7     & 25.07  & 4     & 71.01  & 4     & 49.29  & 4     & 3.83  \\
\rv{M2N2}  & 30.43  & 9     & 70.64  & 4     & 17.58  & 6     & 19.12  & 7     & 60.28  & 14    & 21.55  & 20    & 10.00  \\
\rv{AdaMemBLS} & 40.88  & 3     & 70.70  & 3     & 18.58  & 5     & 19.16  & 6     & 59.24  & 16    & 42.17  & 8     & 6.83  \\
\midrule
Ours  & 54.79  & 1     & 74.78  & 1     & 43.31  & 1     & 46.96  & 1     & 77.27  & 1     & 52.96  & 1     & 1.00  \\
\bottomrule
\end{tabular}%

}
  \label{tab:comparison_avg}%
  % \vspace{-2\baselineskip}
  \end{table}
Table \ref{tab:comparison_avg} summarizes the evaluation metric scores of models across all datasets, clearly indicating that SimAD achieved an improvements of \textbf{9.07\%} on F1, \textbf{3.18\%} on Aff-F1, \textbf{16.63\%} on UAff-F1, and \textbf{20.55\%} on NAff-F1, and \textbf{6.83\%} AUC, compared to the SOTA baseline.
In contrast, although models such as TimesNet and D3R exhibit good performance on specific evaluation metrics (UAff-F1, Aff-F1), their performances on other metrics (Aff-F1, NAff-F1) are inferior to those of SimAD, and even lower than those of random algorithms. 
Fig. \ref{sfig:comparison} shows the advantages of our model, compared to other baseline models across multiple evaluation metrics. 
This further confirms the superior generalization capability of SimAD, whereas other algorithms may display larger performance fluctuations due to an excessive focus on specific dataset characteristics.
(See a more detailed comparison in Appendix \ref{sec:quan_ana}.)
% 在单变量数据集UCR的评估中，SimAD同样展现了优于其他模型的性能。表\ref{tab:ucr}展示了SimAD和基线模型的具体指标分数，包括Avg.+(只使用大于0的指标计算)和Avg.。SimAD的F1指标比最佳基线分别获得了Avg.+的13.33(0.0166->0.1499)和Avg.的13.38(0.0161->0.1499)的绝对提升,UAff-F1指标比最佳基线分别获得了Avg.+的25.75(0.0941->0.3516)和Avg.的17.52(0.0201->0.1953)的相对提升。

Additionally, we evaluate the effectiveness of our model in handling univariate dataset UCR in Table \ref{tab:ucr}, which includes Avg.+ (calculated using only positive values) and Avg. scores. From this table, SimAD demonstrates superior performance. Specifically, SimAD achieves an improvement of \textbf{13.33\%} (1.66\% \(\to\) 14.99\%) in \textbf{Avg.+ F1} and \textbf{13.38\%} (1.61\%\(\to\)14.99\%) in \textbf{Avg. F1}, compared to the best baseline. Furthermore, SimAD achieves an improvement of \textbf{25.75\%} (9.41\%\(\to\)35.16\%) in \textbf{Avg.+ UAff-F1} and \textbf{17.52\%} (2.01\%\(\to\)19.53\%) in \textbf{Avg. UAff-F1}, compared to the best baseline.

\begin{table}%{r}{0.4\textwidth}
  \setlength\tabcolsep{2pt}
  \centering
  % \vspace{-5pt}
  \vspace{-\baselineskip}
      \caption{Comparison of different models on dataset UCR.}
  \resizebox{1\linewidth}{!}{
    \begin{tabular}{c|c|cccccc}
    \toprule
    \multicolumn{2}{c|}{\textbf{Datasets}} & \multicolumn{6}{c}{\textbf{UCR}} \\
    \midrule
    \textbf{Metrics} & \textbf{Methods} & \textbf{F1} & \textbf{Aff-Pre} & \textbf{Aff-Rec} & \textbf{Aff-F1} & \textbf{UAff-F1} & \textbf{NAff-F1} \\
    \midrule
    \multirow{3}[2]{*}{\textbf{Avg.+}} & AnomTrans & 1.17  & 50.66  & 98.94  & 66.88  & \uline{9.41 } & \uline{10.88 } \\
       & DCdetector & \uline{1.66 } & 50.90  & 99.96  & \uline{67.41 } & 6.71  & 5.52  \\
       & Ours & \textbf{14.99 } & 57.83  & 99.91  & \textbf{72.52 } & \textbf{35.16 } & \textbf{34.38 } \\
    \midrule
    \multirow{3}[2]{*}{\textbf{Avg.}} & AnomTrans & 1.15  & 50.39  & 97.73  & 66.55  & 1.30  & 1.23  \\
       & DCdetector & \uline{1.61 } & 50.63  & 83.30  & \uline{67.06 } & \uline{2.01 } & \uline{2.64 } \\
       & Ours & \textbf{14.99 } & 57.83  & 97.47  & \textbf{72.52 } & \textbf{19.53 } & \textbf{19.46 } \\
    \bottomrule
    \end{tabular}%
    }
  \label{tab:ucr}%
  % \vspace{-\baselineskip}
  \end{table}

\begin{figure*}[t]
    \centering
    \includegraphics[width=\linewidth]{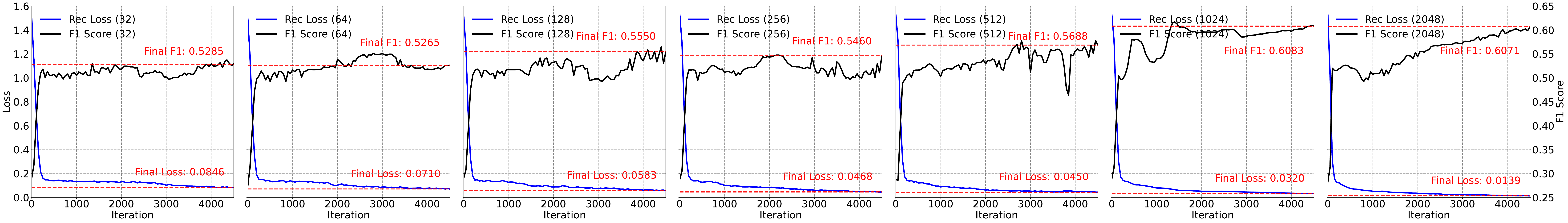}
    \caption{The impact of window size on F1-score and the reconstruction loss when the window size varies from 32 to 2048.}
    \label{fig:para_win}
    % \vspace{-10pt}
\end{figure*}

% \begin{figure*}[ht]
%     \centering
%     \begin{minipage}{.4\textwidth}
%         \centering
%         \includegraphics[width=0.9\linewidth]{jpy_plot/para_qlen_2vs_WADI.pdf}
%         \caption{Different numbers of embeddings.}
%         \label{fig:para_emb}
%     \end{minipage}%
%     \hspace{0.05\textwidth}
%     \begin{minipage}{.5\textwidth}
%         \centering
%         \includegraphics[width=\linewidth]{excel_plot/para.pdf}
%         \caption{Other hyper-parameters. The error bar represents the standard deviation.}
%         \label{fig:para_others}
%     \end{minipage}
%     % \caption{The sensitivity of different hyper-parameters.}
% \end{figure*}

\begin{figure*}[ht]
    \centering
    \includegraphics[width=0.95\linewidth]{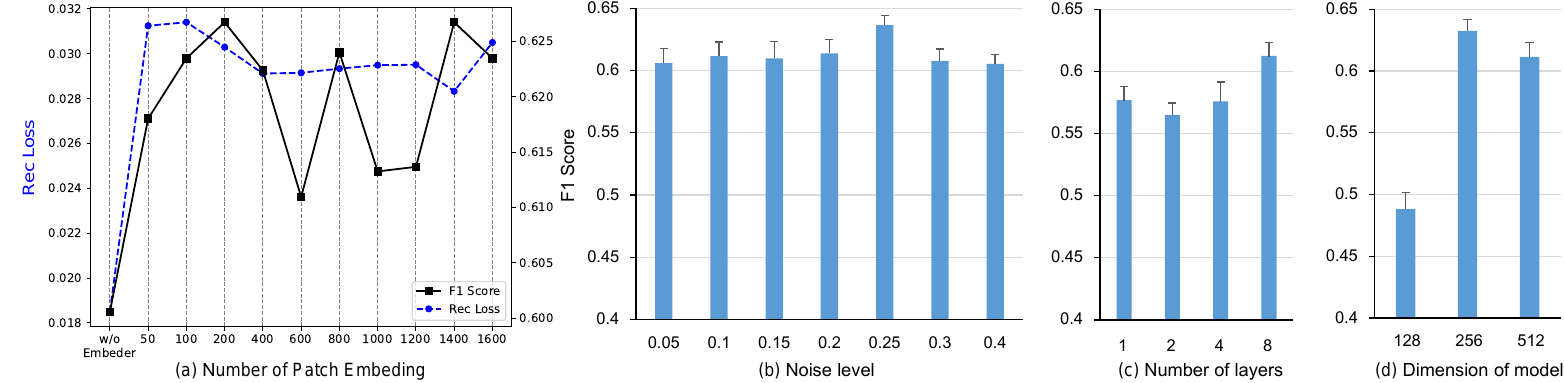}
        \caption{Different numbers of embeddings and other hyper-parameters. The error bar represents the standard deviation.}
        % \label{fig:para_others}
        % \label{fig:para_emb}
        \label{fig:para_all}
    % \caption{The sensitivity of different hyper-parameters.}
\end{figure*}

% \begin{figure}
%     \centering
%     \resizebox{0.8\linewidth}{!}{
%     \subfloat[Comparison analysis. Note that the vertical axis represents relative performance, while the data labels indicate the authenticity scores of the models.
%     ]{\includegraphics[width=.45\linewidth]{excel_plot/comparision.pdf} \label{sfig:comparison}}
%     \hspace{2mm}
%     \subfloat[Ablation analysis. The data labeling serves as an upper bound on the performance improvement achievable by our enhanced model.]{\includegraphics[width=.45\linewidth]{excel_plot/ablation.pdf} \label{sfig:ablation}}
%     }
%     % \caption{Comparison and ablation analysis.}
% \end{figure}
\subsection{Sensitivity Analysis}\label{sec:sen_ana}
% 我们对SimAD的超参数进行了敏感性分析。显著影响SimAD性能的超参数主要是窗口的大小和patch embedding的数量（其他见附录）。图\ref{fig:para_win}证明了不同窗口大小对模型在WADI数据集上性能的影响。随着窗口大小的增加，模型的loss总体上呈现下降趋势，并且模型的准确性不断提高。更大窗口包含更多的时序信息和上下文信息，使模型对数据的周期性、趋势等有更全面的了解，这提高了最终的检测表现。值得注意的是，当窗口大小从1024增加到2048时，模型损失进一步下降，但最终F1分数提高有限，说明更大的窗口也会加大优化难度。
% 图\ref{fig:para_emb}展示了Patch Embedding数量对模型性能的影响。当Patch Embedding为0(相当于原始的注意力机制)，模型性能远低于使用了嵌入查询的EmbedPatch。这可以证明我们模型性能的优越以及EmbedPatch增强记忆正常样本的有效性。随着Patch Embedding数量的增加，模型的Rec Loss和F1 Score均产生波动变化。根据图\ref{fig:para_emb}可以推断出当Patch Embedding数量在100~200之间时，模型的性能最佳。
We performed sensitivity analysis on hyperparameters of SimAD, which include the window size and the number of patch embeddings. Fig. \ref{fig:para_win} illustrates the influence of different window sizes on the model's performance on the WADI dataset. As the window size increases, the model's loss exhibits a decreasing trend, and its accuracy consistently improves. A larger window size incorporates more temporal and contextual information, enabling the model to capture patterns and trends and improve the detection performance. It is worth noting that when the window size increases from 1024 to 2048, the model's loss further decreases while the improvement in the F1 score is limited. This suggests that larger window sizes also increase the optimization difficulty.

Fig. \ref{fig:para_all}(a) depicts impacts of the number of patch embedding on performance on the WADI dataset. When the number of patch embedding is set to 0 (i.e., the original attention mechanism), the model's performance is significantly lower. This demonstrates the effectiveness of EmbedPatch in enhancing the memory of normal samples. As the number of patch embedding increases, both the model's Rec Loss and F1 Score exhibit fluctuating changes. It can be inferred that the model achieves the optimal performance when the number of patch embedding is between 100 and 200.
From Fig. \ref{fig:para_all}(b-d), it is evident that increasing the number of layers and setting an appropriate level of noise and dimension of features can result in improved model performance on the WADI dataset. However, a notable decline in performance is observed when the dimension is set to 128. This indicates that models with smaller dimensions face challenges in learning complex features.

\subsection{Ablation Analysis}
\begin{figure}[ht]%{r}{0.4\textwidth}
        \centering
        % \vspace{-5pt}
        \includegraphics[width=0.95\linewidth]{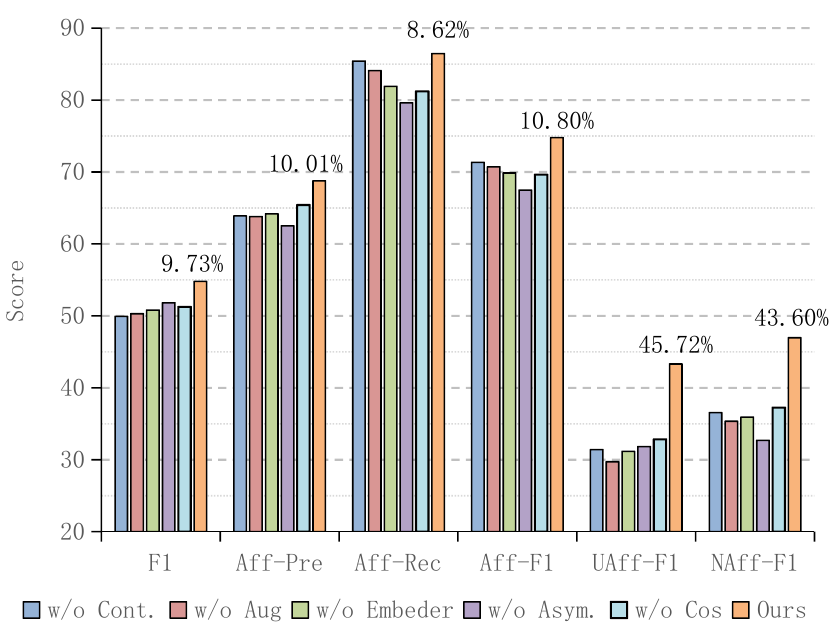}
        \caption{Ablation study. The data labeling denotes an upper bound on the performance improvement of our model.}
        \label{sfig:ablation}
        % \vspace{-5pt}
\end{figure}

\begin{table*}[htbp]
  \centering
  \setlength\tabcolsep{2pt}
  \caption{Full ablation results.}
  \resizebox{0.95\linewidth}{!}{
% Table generated by Excel2LaTeX from sheet 'LATEX'
\begin{tabular}{l|cccccc|cccccc|cccccc}
\toprule
\textbf{Datasets} & \multicolumn{6}{c|}{\textbf{MSL}}             & \multicolumn{6}{c|}{\textbf{SMAP}}            & \multicolumn{6}{c}{\textbf{SWaT}} \\
\midrule
\textbf{Variations} & \textbf{F1} & \textbf{Aff-Pre} & \textbf{Aff-Rec} & \textbf{Aff-F1} & \textbf{UAff-F1} & \textbf{NAff-F1} & \textbf{F1} & \textbf{Aff-Pre} & \textbf{Aff-Rec} & \textbf{Aff-F1} & \textbf{UAff-F1} & \textbf{NAff-F1} & \textbf{F1} & \textbf{Aff-Pre} & \textbf{Aff-Rec} & \textbf{Aff-F1} & \textbf{UAff-F1} & \textbf{NAff-F1} \\
\midrule
w/o Cont. & 23.90  & 54.13  & 99.08  & 70.01  & 10.84  & 15.23  & 22.56  & 51.94  & \uline{99.86 } & 68.34  & 1.89  & 7.47  & 80.31  & 77.66  & 76.52  & 77.09  & 62.29  & 64.21  \\
w/o Aug & \uline{29.07 } & \uline{54.64 } & 98.88  & \uline{70.39 } & \uline{12.72 } & \uline{16.98 } & \uline{23.78 } & \uline{52.66 } & 98.11  & 68.53  & \uline{4.75 } & \uline{10.09 } & 79.14  & \textbf{80.60 } & 64.14  & 71.43  & 61.34  & 62.63  \\
w/o Embeder & 27.20  & 54.12  & 98.85  & 69.94  & 10.80  & 15.20  & 17.89  & 50.04  & 99.64  & 66.62  & -5.76  & 0.15  & 80.31  & 77.66  & 76.52  & 77.09  & 62.29  & 64.21  \\
w/o Asym. & 28.23  & 54.13  & 98.94  & 69.98  & 10.86  & 15.25  & 22.39  & 51.46  & 99.77  & 67.90  & -0.05  & 5.69  & \uline{82.01 } & 77.35  & \uline{83.37 } & \textbf{80.25 } & \uline{63.95 } & \uline{66.06 } \\
w/o Cos & 28.50  & 53.95  & \uline{99.10 } & 69.87  & 10.20  & 14.64  & 23.32  & 52.24  & \textbf{99.91 } & \uline{68.61 } & 3.11  & 8.59  & 77.77  & 75.02  & \textbf{83.53 } & 79.05  & 60.09  & 62.59  \\
Ours  & \textbf{30.02 } & \textbf{56.30 } & \textbf{99.76 } & \textbf{71.98 } & \textbf{18.50 } & \textbf{22.37 } & \textbf{29.39 } & \textbf{56.84 } & 99.82  & \textbf{72.44 } & \textbf{19.91 } & \textbf{24.07 } & \textbf{82.03 } & \uline{78.46 } & 80.88  & \uline{79.65 } & \textbf{64.93 } & \textbf{66.82 } \\
\midrule
\textbf{Datasets} & \multicolumn{6}{c|}{\textbf{WADI}}            & \multicolumn{6}{c|}{\textbf{PSM}}             & \multicolumn{6}{c}{\textbf{NIPS-TS-Swan}} \\
\midrule
\textbf{Variations} & \textbf{F1} & \textbf{Aff-Pre} & \textbf{Aff-Rec} & \textbf{Aff-F1} & \textbf{UAff-F1} & \textbf{NAff-F1} & \textbf{F1} & \textbf{Aff-Pre} & \textbf{Aff-Rec} & \textbf{Aff-F1} & \textbf{UAff-F1} & \textbf{NAff-F1} & \textbf{F1} & \textbf{Aff-Pre} & \textbf{Aff-Rec} & \textbf{Aff-F1} & \textbf{UAff-F1} & \textbf{NAff-F1} \\
\midrule
w/o Cont. & 62.68  & 70.55  & \uline{84.04 } & 76.71  & 49.26  & 55.20  & 45.78  & 58.83  & \uline{96.05 } & 72.96  & 21.47  & 29.83  & 64.34  & 70.24  & 57.00  & 62.93  & 42.69  & 47.34  \\
w/o Aug & \uline{63.19 } & \uline{75.92 } & 81.53  & \uline{78.63 } & \uline{59.40 } & \uline{63.38 } & 41.79  & 56.43  & \textbf{96.15 } & 71.12  & 12.98  & 22.67  & 64.83  & 62.55  & \textbf{65.81 } & \uline{64.14 } & 27.13  & 36.34  \\
w/o Embeder & 61.36  & 70.73  & 59.61  & 64.70  & 44.29  & 48.91  & 49.78  & \uline{61.47 } & 94.80  & \textbf{74.58 } & \uline{29.86 } & \uline{36.93 } & 68.32  & 71.04  & \uline{62.03 } & \textbf{66.23 } & 45.47  & \uline{50.14 } \\
w/o Asym. & 56.89  & 63.33  & 63.98  & 63.65  & 56.26  & 37.64  & \textbf{52.26 } & 59.93  & 91.27  & 72.35  & 24.93  & 32.62  & 69.14  & 68.85  & 40.28  & 50.82  & 35.06  & 38.95  \\
w/o Cos & 63.13  & 72.66  & 74.79  & 73.71  & 51.69  & 56.43  & 44.00  & 59.83  & 86.97  & 70.89  & 24.46  & 32.07  & \uline{70.79 } & \textbf{78.76 } & 42.99  & 55.62  & \uline{47.47 } & 49.21  \\
Ours  & \textbf{63.98 } & \textbf{82.36 } & \textbf{92.81 } & \textbf{87.27 } & \textbf{73.59 } & \textbf{76.26 } & \uline{52.07 } & \textbf{62.46 } & 91.76  & \uline{74.33 } & \textbf{32.63 } & \textbf{39.20 } & \textbf{71.23 } & \uline{76.17 } & 53.75  & 63.03  & \textbf{50.29 } & \textbf{53.04 } \\
\bottomrule
\end{tabular}%
        }
  \label{tab:abl_full}%
\end{table*}%
To validate the effectiveness of our model, we conducted ablation studies on ContrastFusion, data augmentation, EmbedPatch, and asymmetric optimization in Fig. \ref{sfig:ablation}. Here we provide the descriptions about the model's variants:

1. \textbf{w/o Contrastive} (Cont.): indicates the removal of contrastive learning in the ContrastFusion. By removing this, $\mathcal{L}_{cont}$ (Eq. \eqref{eq:L_cont}) no longer works, and the similarity relationships between positive and negative samples will not be learned.

2. \textbf{w/o Aug}: signifies negative sample generation without denoising learning, where $\mathcal{L}_{denoise}$ (Eq. \eqref{eq:L_denoise}) does not work. It is noteworthy that despite generating negative samples, it is still possible to conduct contrastive learning between positive and negative samples, i.e., $\mathcal{L}_{cont}$ (Eq. \eqref{eq:L_cont}) remains active.

3. \textbf{w/o Embeder}: when the Embeder, \textit{i.e.}, 
$\mathbf{E}$, is removed, the architecture of SimAD regresses to a Transformer. Originally, the value matrix in SimAD was $\mathbf{V} = \mathbf{W}^V \mathbf{E}$. After regressing to a Transformer, the generation of $\mathbf{V}$ aligns with matrices $\mathbf{Q}$ and $\mathbf{K}$, \rv{\textit{i.e.}, $\mathbf{V} = \mathbf{N} \mathbf{W}^V$.} For simplicity, $\mathbf{W}^V$ represents learnable parameters, and \rv{$\mathbf{N}$} is the pre-input. When EmbedPatch is removed, SimAD's backbone reverts to a Transformer, where self-attention is uniform across Q, K, and V.
% This straightforward change can significantly improve SimAD's detection capabilities.

4. \textbf{w/o Asymmetric} (Asym.): our asymmetric optimization relies on gradient stop. Removing the asymmetric optimization eliminates the $\operatorname{StopGrad}$ from the equation. As both MSE and Similarity losses are symmetric, after removing this operation, Eq. \eqref{eq:L_cont} regresses to a symmetric optimization method.

5. \textbf{w/o Cos}: indicates the removal the cosine terms from Eq. \eqref{eq:L_rec}, \eqref{eq:L_denoise}, and \eqref{eq:L_cont} while keeping the MSE loss.

% 不同消融实验的结论如下：
The detailed ablation studies are in Table \ref{tab:abl_full}. 
The observation from ablation studies are summarized as below:
\begin{enumerate}
    \item After removing the \textbf{w/o Contrastive}, i.e., the removal of positive and negative sample contrastive learning, the F1 score of the model decreased by 7.88\%, indicating that ContrastFusion is important in learning distinct features to differentiate between normal and abnormal samples. 
    \item By disabling \textbf{w/o Aug}, i.e., the removal of data augmentation, the detection performance of SimAD shows a decrease of 43.69\% in UAff-F1, demonstrating its ability to improve the model's generalization.
    \item For \textbf{w/o Embeder}, the EmbedPatch in the attention mechanism degrades to a simple value matrix. The ablation of this module has the most significant impact on the model's F1 score, which indicates that incorporating EmbedPatch into the attention mechanism is crucial to learning appropriate representations of temporal data.
    \item \textbf{w/o Asymmetric} indicates the model without asymmetric optimization. The ablation of this module results in a decrease in all performance metrics, particularly a significant drop of 43.69\% in NAff-F1. This suggests that asymmetric optimization helps the model focus on abnormal samples. 
    \item The experimental results of \textbf{w/o Cos} demonstrate that upon removing the cosine similarity penalty, both the predictive continuity and detection performance of SimAD decline. 
\end{enumerate}
The comprehensive results of ablation studies, along with the setting of the experiment, are in Table \ref{tab:abl_full}.
% in Appendix \ref{sec:res}. 

\rv{To verify the effectiveness of the EmbedPatch encoder and ContrastFusion modules in our proposed SimAD framework for addressing the first Challenge outlined in the introduction, \textit{i.e.}, enhancing the dissimilarity between features of both normal and abnormal data, we extracted feature representations before and after the ContrastFusion projection head. Next, we conducted a multi-sampling to calculate the KL divergence between normal and abnormal features at both stages. The results are in Table \ref{tab:kl_div}. From this table, it can be observed that removing the ContrastFusion module (w/o Cont.) leads to an increase in KL divergence between the feature distributions of normal and abnormal data on the SWaT and WADI datasets, indicating that the representations become more distinct. Moreover, when both components are integrated into our model, the separation between normal and abnormal feature distributions is maximized. This demonstrates that SimAD effectively overcomes the limitations of previous approaches.}
% 在SimAD中，最重要的两个组件是EmbedPatch Encoder和ContrastFusion，为了探究它们能够真正解决引言中的挑战1并实现正常与异常的不相似性增加。我们深度的对这两个组件进行分析，我们首先得到计算ContrastFusion投影头之前的特征与之后的特征，随后通过多次采样的方式得到正常与异常特征在投影前后的KL散度，结果如图\ref{tab:kl_div}所示。如果移除ConstrastFusion（w/o Cont.），在SWaT和WADI两个数据集，正常与异常的分布空间都明显远离（KL散度变大），并且我们的模型在同时使用这两个组件时，正常与异常的特征分布空间距离最远，这说明SimAD能够真正解决过去的局限。遗憾的是，在PSM数据集上，这种效应几乎不存在，我们认为这是因为PSM的异常比率过高，达到27.8\%，数据集中主要是区间异常，此时基于高斯噪声的数据增强并不能噪声空间与异常空间进一步拉近。考虑到SimAD的设计初衷便是尽可能保持模型简洁，我们认为这是后续未来的方向，在附录中我们对此也进行了讨论。
\begin{table}[htbp]
  \centering
  \caption{The KL divergence between normal and anomalous features both before and after projection.}
  \resizebox{\linewidth}{!}{
    % Table generated by Excel2LaTeX from sheet '消融的KL散度'
\begin{tabular}{l|cc|cc}
\toprule
Datasets & \multicolumn{2}{c|}{SWaT} & \multicolumn{2}{c}{WADI} \\
\midrule
Variations & Before Proj. & After Proj. & Before Proj. & After Proj. \\
\midrule
w/o Cont. & 7.34E-01 & 1.66E-02 & 4.95E-05 & 2.85E-03 \\
w/o Embeder & 4.34E-01 & 2.04E-02 & 3.25E-03 & 1.59E-01 \\
Ours  & 4.25E-01 & 5.04E-01 & 3.00E-02 & 9.89E-01 \\
\bottomrule
\end{tabular}%
}
  \label{tab:kl_div}%
\end{table}%

\begin{figure*}
    \centering
    \subfloat[Loss variation.]{\includegraphics[width=.48\linewidth]{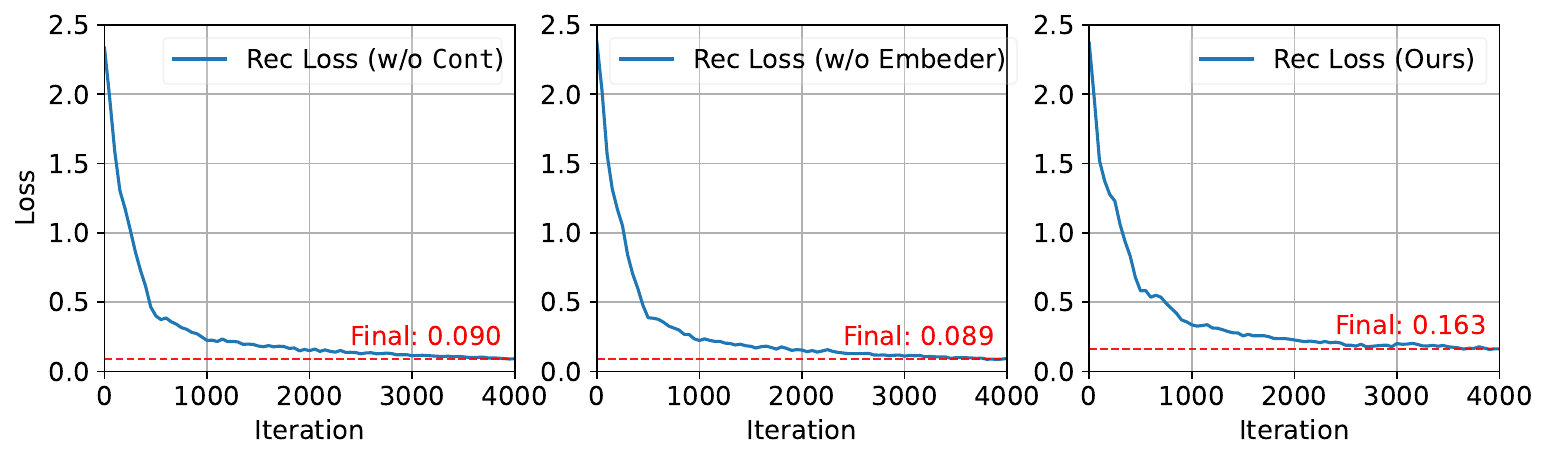}
    \label{sfig:abl_loss}
    }
    \subfloat[F1 score]{
    \includegraphics[width=.48\linewidth]{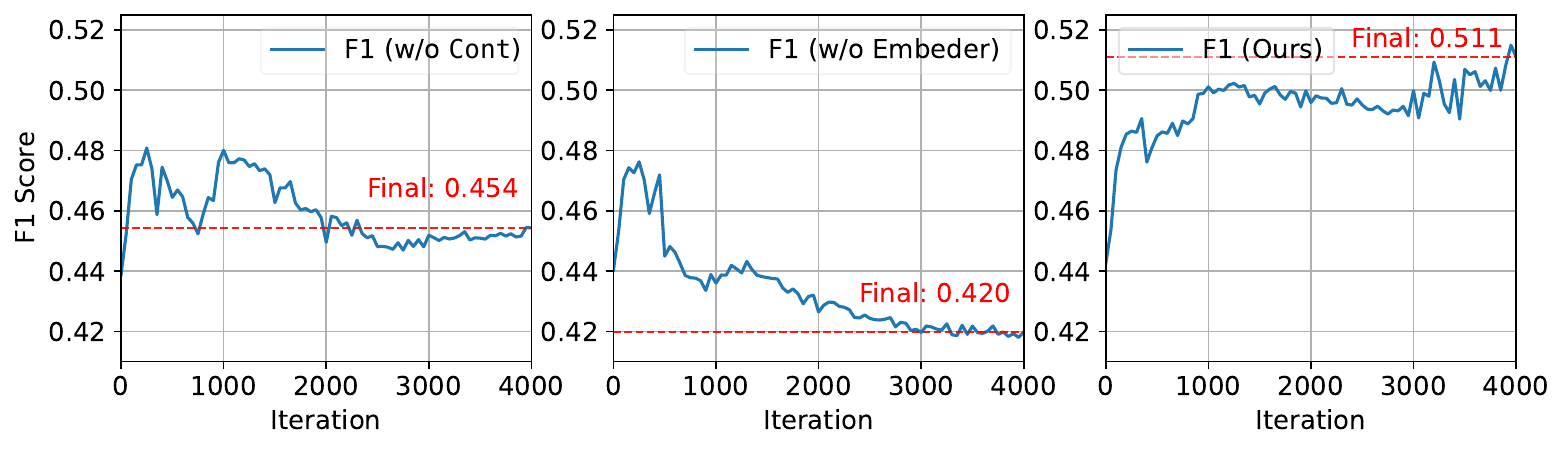}
    \label{sfig:abl_f1}
    }
    \caption{Loss and F1 score variations on dataset PSM.}
    \label{fig:abl_psm}
\end{figure*}

Moreover, Fig. \ref{fig:abl_psm} provides
an illustration of the training on dataset PSM. It is evident that the \textbf{w/o Embedder} module achieves the lowest loss but with the lowest F1 score. This can be attributed to that EmbedPatch contributes SimAD to memorizing normal samples and enhancing generalization. By capturing the characteristics of normal data, EmbedPatch improves the effectiveness of anomaly detection by leveraging the distinct differences between normal and abnormal samples, as demonstrated in Fig. \ref{fig:cases}(b) and Fig. \ref{fig:heat_map_all} in Appendix \ref{sec:res}.
The \textbf{w/o Contrastive} module also exhibits a low loss but suffers from negative optimization, resulting in a relatively low F1 score. In contrast, SimAD, despite having a higher loss, achieves the highest F1 score. This demonstrates that optimizing solely for reconstruction is inadequate, and SimAD effectively addresses the challenges outlined in the introduction.

\begin{figure*}[!t]
    \centering
    \subfloat[Detection performances in real-world.]{\includegraphics[width=.44\linewidth]{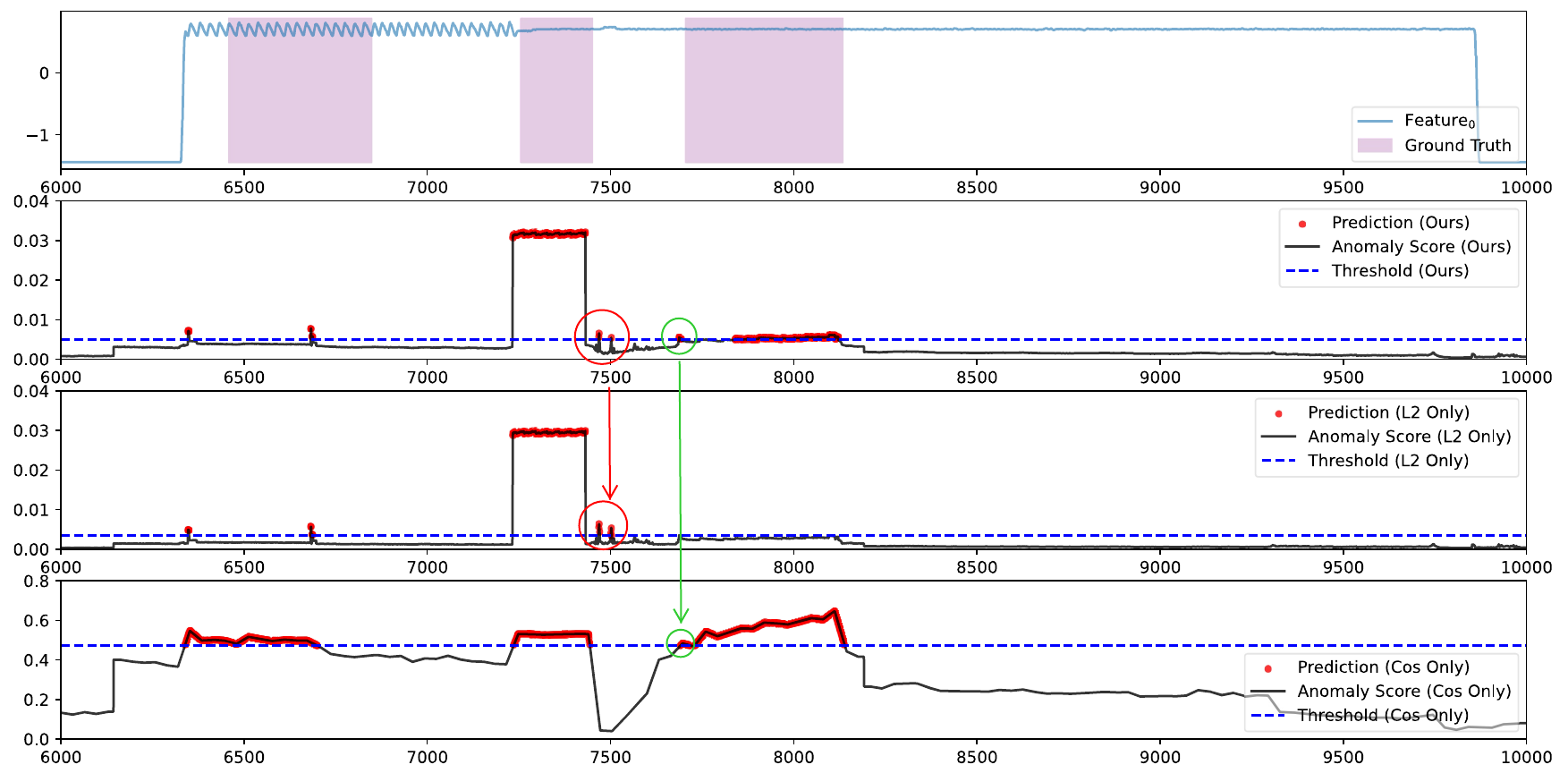}\label{sfig:case_study}}
    \subfloat[Similarity scores of query and embedding.]{\includegraphics[width=.52\linewidth]{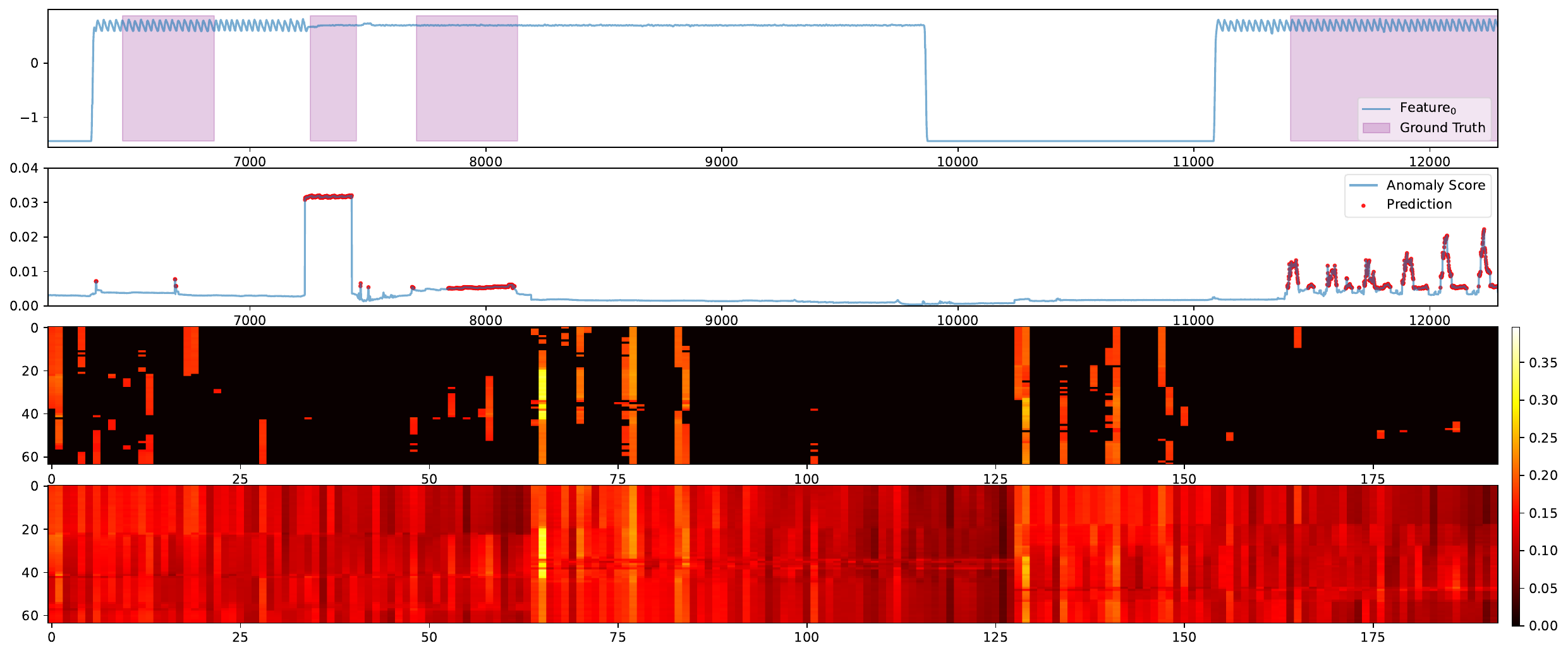}\label{sfig:map_heat}}
    \caption{Real world case study on dataset SWaT.}
    \label{fig:cases}
\end{figure*}

\subsection{Visualization Analysis}
\subsubsection{Real-world cases analysis}

% 图\ref{sfig:case_study}与图\ref{fig:case_study2}（附录中）展示了SimAD在SWaT数据集上的真实表现。第一排图片表示了原始数据与真实标签。第二排图片展示我们的模型的预测结果，我们的模型更够将大部分异常检测出来，并且异常的分数明显高于正常数据的。第三和第四排的图片表示分别采用L2损失和patch-based cosine similarity损失作为异常分数的结果。注意到，前者更关注短区间异常，后者能够更加关注长段异常，这说明了我们的异常分数设计的有效性与合理性。
The performance of SimAD on dataset SWaT is in Fig. \ref{sfig:case_study} and Fig. \ref{fig:case_study2} (in Appendix \ref{sec:vis}). The first row of images showcases the original data and the ground truth labels. 
\rv{Its y-axis represents the magnitude of the first channel in the time series.}
% 它的轴坐标表示时序第一个通道特征的值大小。
The second row of images denotes the predicted results, which effectively detects most anomalies by assigning significantly higher scores to abnormal data. 
\rv{The y-axis represents the magnitude of the anomaly score.}
The third and fourth rows of images illustrate the results obtained using L2 loss and patch-based cosine similarity loss as anomaly scores respectively. 
% 这两行的纵坐标含义一样，可视化使用的模型时间窗口为2048，patch大小为32，因此最终一个时间窗口内会产生64个patch。我们额外计算了patch与embedding之间的相关性，产生一个\(64\times 64\)的注意力特征图，即EmbedPatch Enocder中最后一层\(\mathbf{Q}^{(-1)}\)与\(\mathbf{V}^{(-1)}\)的相关性。注意，这里并不是Q与K的相关性，因为我们希望得到时序与embedding的相关性，说明正常数据区域与embedding的相关性更强。因此这里轴坐标表示与第几个embedding。第四个区域的图片，可以理解为每一个embedding都找到最相关的5个patch（时序区域）。热力图的颜色表示了这种相关性的强弱。因此，在第三区域中，如果某一位置的patch与任意embedding的相关性都不强，则表现为该位置的热力图为黑色。反之，则认为该位置的patch与某一些embedding的相关性强，且大于另外一些位置的patch与embedding的相关性。

It is noticed that the former focuses more on short-duration anomalies, while the latter can capture long-duration anomalies. In Appendix \ref{sec:vis}, Fig. \ref{fig:case_anom_trans} shows the detection results of AnomTrans and NPSR. Our method exhibits superior performance compared to AnomTrans, as the latter only detects a few anomalies and encounters difficulties in accurately detecting segment anomalies. The case study highlights the effectiveness and rationale behind our anomaly score design.

\subsubsection{Explanation of patch embedding}
Fig. \ref{sfig:map_heat} shows the detection performance of SimAD, where the fourth row denotes the similarity between the query and the embedding of the last layer, 
\rv{while the vertical axes of the third and fourth regions illustrates correlations between the query embedding and the learned embedding of value matrices. In the third and the fourth regions of the right subfigure, the horizonal axis denotes the ordinal number of the embedding. The colors in the heatmap indicate the strength of these correlations, with precise values corresponding to the color bar for reference. Specifically, given a time window of 2048, we divide it into 64 non-overlapping patches, each with a length of 32. There are 64 learned embedding in the last layer of our EmbedPatch Encoder. In the following, we consider the 64 non-overlapping patches from the input window as the query embedding. Correspondingly, the 64 learned embedding produced by the final layer of the EmbedPatch Encoder serves as the value matrices. Next, we compute the pairwise correlations between the 64 non-overlapping patches and 64 learned embedding representations, resulting into a  attention feature map. This feature map is depicted in the fourth region of the right subfigure. In contrast, the third region of the right subfigure highlights the top five most relevant query embeddings, selected from the 64 patches (i.e., segments of the 2048-length time window), for each learned embedding. In the third region, patches that exhibit weak relevance to all embeddings are shown in black. Conversely, patches with stronger relevance to certain embeddings are displayed in vibrant colors, highlighting their higher correlations relative to other patches.
It is important to note that the  attention feature map denotes the correlations between the query (\(\mathbf{Q}\)) vectors and the value (\(\mathbf{V}\)) vectors in the last layer of the EmbedPatch Encoder, i.e., the relationships between the input patches from the time series and the learned embedding, rather than the correlations between the query (\(\mathbf{Q}\)) vectors and the key (\(\mathbf{K}\)) vectors.}
% while the third row of images displays the similarity scores between the last layer's embedding and the top 5 most relevant queries. 
% \rv{The vertical axis in these two rows share the same meaning. The model used for visualization has a time window of 2048 and a patch size of 32, producing 64 patches per time window. We compute the correlation between patches and embeddings to generate a \(64\times 64\) attention feature map, specifically the correlation between \(\mathbf{Q}^{(-1)}\) and \(\mathbf{V}^{(-1)}\) in the last layer of the EmbedPatch Encoder. Note that this is not the correlation between Q and K, but rather between time series and embeddings, indicating stronger relevance in normal data regions. The axis coordinate here represents the ordinal number of the embedding. In the fourth area's image, each embedding identifies its top five relevant patches (time series regions). The heatmap color reflects the strength of these correlations. In the third area, if a patch has weak relevance to all embeddings, it appears black. Conversely, patches with stronger relevance to some embeddings appear in contrasting colors, indicating their higher correlation compared to other patches.}
Combined with Fig. \ref{fig:heat_map_all} in Appendix \ref{sec:vis}, it can be observed that in the lower layers, the embedding establishes relationships with a majority of queries. However, only the normal data can establish relationships with EmbedPatch in the higher layers. 
This means that in the shallow layers, SimAD learns general representations. While in the higher layers, SimAD learns features associated with normal data, which results in less similarity between abnormal data and EmbedPatch. Consequently, the patch embeddings enhance the detection performance of SimAD.

\section{Conclusion}
% 本文提出了a Simple Dissimilarity-based approach for 时序异常检测。该框架简单通用且支持更长的时间窗口。实验证明了我们的方法远超已有工作。同时我们提出改进的UAff与NAff评估指标推动算法的性能评估。在未来我们希望将该检测方案扩展到其他更多场景，并实现异常检测统一模型。

This paper introduces a \textbf{Sim}ple dissimilarity-based approach for time series \textbf{A}nomaly \textbf{D}etection (\textbf{SimAD}). The proposed framework is distinguished by its simplicity and versatility, allowing for the use of longer time windows. Experimental results demonstrate the superiority of our method compared to existing approaches. Additionally, we present two improved evaluation metrics, \textbf{UAff} and \textbf{NAff}, which effectively assess the algorithm's performance and overcome many shortcomings of previous metrics. In future work, we plan to extend this detection scheme to various other scenarios and aim to develop a unified model for anomaly detection.

\section*{Acknowledgment}
This work was supported in part by National Natural Science Foundation of China No. 62476101, 92467109, U21A20478, 62306052, National Key R\&D Program of China 2023YFA1011601, the Major Key Project of PCL, China under Grant PCL2023AS7-1, and Natural Science Foundation of Chongqing under Grant CSTB2023NSCQ-LZX0092.

\setcounter{section}{0}
\renewcommand{\thesection}{\Alph{section}}

% \newpage
% \appendix

\section*{Appendix}

% \section*{\huge{Appendix}}
This is the appendix of SimAD: A Simple Dissimilarity-based Approach for Time Series Anomaly Detection.

\section{Related work}\label{sec:app_releted}
Unsupervised methods have garnered considerable attention due to the scarcity of labeled data for anomaly detection in time series sequences. These methods are particularly advantageous in scenarios where obtaining labeled data is expensive, time-consuming, or impractical. The primary objective is to identify outliers or anomalies that deviate significantly from expected patterns without relying on prior annotations. Broadly, these methods can be categorized as follows:

(1) Algorithms based on classical machine learning, such as those proposed in \cite{deep_svdd, deep_if}, transform traditional machine learning techniques into deep neural network architectures. This transformation enhances their ability to handle complex and high-dimensional data, allowing for more nuanced feature extraction and representation learning. By integrating classical algorithms with deep learning, these methods can capture intricate patterns that signify anomalies more effectively than their purely classical counterparts.

(2) Reconstruction-based methods, exemplified by \cite{usad, tcn_ed, ae1}, primarily involve training models on normal data to learn how to reconstruct input time series. During testing, the reconstruction error is used as an anomaly score, with anomalous data typically yielding higher errors due to their deviation from learned normal patterns. This approach relies on the assumption that normal data can be accurately reconstructed, while anomalies will result in significant discrepancies.
For instance, USAD \cite{usad} employs a dual-decoder network structure and utilizes two-stage training to enhance performance. In this architecture, one decoder reconstructs the input data to capture normal patterns, while the second decoder focuses on learning the residuals—discrepancies that occur during reconstruction. This dual-decoder approach enables the model to effectively differentiate between normal and anomalous instances by analyzing both reconstruction fidelity and residual errors, thereby improving anomaly detection accuracy.

(3) Prediction-based techniques, as demonstrated by \cite{timesnet}, focus on learning from historical data to predict future observations. These models are trained to forecast the next time step or series of steps based on past values. The prediction errors—defined as the difference between predicted and actual values—serve as the basis for anomaly detection. This method effectively identifies anomalies that manifest as significant deviations from expected future behavior. For example, PatchTST \cite{patchtst} employs a single-scale patch-based MLP as its core architecture, whereas the work in \cite{yi2024frequency} applies an MLP to capture frequency-domain information in time series data for predictive modeling.

(4) Generative adversarial learning approaches, such as those discussed in \cite{gan1, gan2, gan3}, employ a dual-network framework consisting of a generator and a discriminator. The generator learns to model the distribution of normal data, while the discriminator is tasked with distinguishing between normal instances and anomalies. This adversarial training process enhances the system's ability to detect subtle anomalies by leveraging the power of generative models to simulate normal data distributions.

In recent years, some innovative algorithms have emerged in the field of anomaly detection, including:
(1) Transformer-based methods \cite{anom_trans, trans1, trans_ad} harness the architecture of Transformers, which have shown exceptional performance in natural language processing tasks. These models apply self-attention mechanisms that allow them to weigh the importance of different parts of the input data, thus enabling them to capture temporal dependencies and contextual information crucial for identifying anomalies in time series data.

(2) Contrastive learning-based methods \cite{couta, ncad, dcdetector, cont1, cont2} focus on learning robust representations by contrasting positive and negative pairs. In the context of anomaly detection, these methods can effectively differentiate between normal and anomalous instances by learning to identify subtle differences in representation, thereby enhancing the model's sensitivity to anomalies. For example, COUTA \cite{couta} emphasized calibrating irregular noise and enhancing the significance of normal samples through uncertainty modeling, introducing a one-class loss based on prior distribution to improve prediction reliability.

(3) Diffusion-based approaches \cite{d3r, diffusion1} utilize mathematical diffusion processes to model how anomalies propagate through complex networks and time series sequences. These techniques are particularly effective in scenarios where the relationships between data points are crucial, as they capture the dynamics of anomaly spread and enhance detection capabilities in interconnected systems.
For example, D3R \cite{d3r} addresses the instability of time series data in non-stationary environments, a common challenge that often leads to a high false positive rate. D3R employs a decomposition and reconstruction method to manage data drift, integrating a noise diffusion model that directly recovers contaminated data. This strategy mitigates the substantial retraining costs typically associated with changes in information bottlenecks, thereby improving overall detection reliability.

(4) Large Language Models (LLMs), such as those adapted for anomaly detection tasks \cite{gpt2adpt}, leverage advanced architectures like GPT-2. These models utilize their extensive capabilities in knowledge representation and context understanding to identify anomalies in time series data. By applying techniques from language modeling to time series analysis, LLMs can uncover complex patterns and detect outliers with high accuracy.
% 通常，这些模型的基本骨架与GPT-2保持一致，即都使用了自注意力机制、前馈神经网络、层规范化学习序列化的特征，基本模块是自注意力模块与残差前馈神经网络组成，多次叠加此组件构成模型的解码器。由于模型是完全的解码器模型，通过自回归的方式预测下一个token，因此无需编码器。GPT-Adapter \cite{gpt2adpt}只需要微调前馈神经网络与层规范化的参数，并且只使用了预训练GPT模型的一部分参数，即可将原本的文本符号预测转换成时序序列预测，并根据预测损失的大小评估异常。
\rv{Generally, the basic structure of these models is consistent with GPT-2. They use self-attention mechanism, feed-forward neural network, and layer normalization to learn sequential features. Their basic units are self-attention module and residual feed-forward neural network, which are stacked multiple times to form the model's decoder. As a pure decoder model, it predicts the next token in an autoregressive way, so an encoder isn't needed. GPT-Adapter \cite{gpt2adpt} just fine-tunes the parameters of the feed-forward neural network and layer normalization. It uses only some of the pre-trained GPT model's parameters to transform text token prediction into time series prediction and assess anomalies based on prediction loss.}

\section{Detailed experimental settings}\label{sec:set}
We utilize official or open-source baselines that have been published on GitHub. These baselines can be downloaded by following:\newline

\begin{enumerate}
    \item Deep SVDD:\url{https://github.com/xuhongzuo/DeepOD}
    \item USAD: \url{https://github.com/xuhongzuo/DeepOD}
    \item TCN-ED:\url{https://github.com/xuhongzuo/DeepOD}
    \item COUTA:\url{https://github.com/xuhongzuo/DeepOD}
    \item TranAD: \url{https://github.com/xuhongzuo/DeepOD}
    \item NCAD: \url{https://github.com/xuhongzuo/DeepOD}
    \item Deep IF: \url{https://github.com/xuhongzuo/DeepOD}
    \item AnomTrans:\url{https://github.com/thuml/Anomaly-Transformer}
    \item TimesNet: \url{https://github.com/xuhongzuo/DeepOD}
    \item DCdetector: \url{https://github.com/DAMO-DI-ML/KDD2023-DCdetector/}
    \item D3R: \url{https://github.com/ForestsKing/D3R/}
    \item GPT2-Adapter: \url{https://github.com/PSacfc/GPT4TS_Adapter}
    \item NPSR: \url{https://github.com/andrewlai61616/NPSR/}
\end{enumerate}

\subsection{Datasets details}\label{sec:data}

% Existing methods extensively utilize many publicly available real-world datasets. We evaluate model performance using six multivariate datasets: (1) MSL (Mars Science Laboratory dataset)，由 NASA 的火星科学实验室（MSL）任务收集，能够反映火星表面的各种物理现象和环境变化 \cite{MSL_dataset}, (2) SMAP (Soil Moisture Active Passive dataset)，MSL和SMAP数据集是NASA收集的公共数据集，包含来自航天器监测事件意外异常（ISA）报告的遥测异常数据系统 \cite{SMAP_dataset}。, (3) SWaT (Secure Water Treatment Testbed)，研究者基于该测试平台模拟了一个小型的水处理厂，包含多个处理单元和传感器，用于监测和控制水处理过程中的各种参数，如流量、压力、水质等，反映了水处理过程中的正常和异常运行状态 \cite{anom_trans}, (4) WADI (Water Distribution 该测试平台模拟了一个城市水分配系统，包括多个水泵、阀门、水箱和管道，以及各种传感器和控制器，常用于研究工业控制系统中的异常检测。 (5) PSM (Pooled Server Metrics)，PSM 数据集 \cite{PSM_dataset}是从 eBay 的多个应用程序服务器节点内部收集的。 有 13 周的训练数据和 8 周的测试数据。由多个服务器的性能指标数据汇总而成，收集了服务器在运行过程中的各种性能指标，如 CPU 使用率、内存使用率、磁盘 I/O、网络带宽，可用于分析服务器的性能瓶颈、预测服务器故障, and (6) Swan是“物联网”的饮用水质量数据集，极低的异常率，属于高难度异常检测数据集. 

\rv{
Existing methods extensively utilize many publicly available real-world datasets. We evaluate model performance on six multivariate datasets:
\begin{enumerate}
    \item MSL (Mars Science Laboratory dataset): Collected by NASA’s MSL mission, it reflects various physical phenomena and environmental changes on Mars \cite{MSL_dataset}.
    \item SMAP (Soil Moisture Active Passive dataset): Along with MSL, it’s a NASA-collected dataset with telemetry anomaly data from spacecraft monitoring events \cite{SMAP_dataset}.
    \item SWaT (Secure Water Treatment Testbed): This platform simulates a small water treatment plant with multiple processing units and sensors, reflecting normal and abnormal states in water treatment \cite{anom_trans}.
    \item WADI (Water Distribution Testbed): It simulates a city’s water distribution system with pumps, valves, tanks, pipes, and sensors, often used for anomaly detection in industrial control systems.
    \item PSM (Pooled Server Metrics): From eBay’s application server nodes, it has 13 weeks of training and 8 weeks of testing data, covering server performance metrics like CPU and memory usage \cite{PSM_dataset}.
    \item Swan: An IoT-based drinking water quality dataset with a very low anomaly rate, making it a challenging benchmark for anomaly detection.
\end{enumerate}
}

%The details are provided in Table \ref{tab:data_set}. 

They can be downloaded in:\newline
\begin{itemize}
    \item MSL \& SMAP: \url{https://github.com/ML4ITS/mtad-gat-pytorch}
    \item  SWaT: \url{https://itrust.sutd.edu.sg/itrust-labs_datasets}
    \item  WADI: \url{https://itrust.sutd.edu.sg/itrust-labs_datasets} or \url{https://github.com/Conviss/visual_dataset}
    \item UCR:\url{https://www.cs.ucr.edu/~eamonn/time_series_data_2018/UCR_TimeSeriesAnomalyDatasets2021.zip}
    \item Swan: \url{https://github.com/Conviss/visual_dataset}
    \item  Others: \url{https://github.com/DAMO-DI-ML/KDD2023-DCdetector}.
\end{itemize}

\subsection{Implementation details}
% 在我们的默认设置中，滑动窗口的长度是2048，模型的隐藏维度是512，注意力头为8，层数为8层，嵌入特征的数量为1000。我们对所有方法都采用了最佳F1搜索策略。训练使用AdamW优化器，batch size为256，学习率为\(10^{-3}\)，并使用余弦学习率调整策略。训练共计20个epoch，每个epoch随机采样500个样本。我们的算法使用Python和PyTorch实现。所有的实验在一块NVIDIA A800 （80GB）上完成。
In our default configuration, we set the length of the sliding window to 2048. The patch size is 32. The model is configured with a hidden dimension of 512, 8 attention heads, and 8 layers. We utilize 1000 embedding features. For all methods, we use the optimal F1 search strategy. The training process employs the AdamW optimizer with a batch size 256 and a learning rate of \(10^{-3}\). We also utilize cosine learning rate scheduling. The training is conducted over 20 epochs, with each epoch randomly sampling 500 samples. Our algorithm is implemented in Python using PyTorch. All experiments were performed on an NVIDIA A800 (80GB) GPU.

\rv{To emphasize the engineering implementation details of the algorithm, we explain the key components in model construction. This section can be directly cross-referenced with the documentation and code comments in our repository.
\textbf{Positional Embedding:}
We use cosine-based positional encoding for temporal positions but do not encode channel position information. The Instance Normalization layer follows the original author's implementation without modifications.
\textbf{EmbedPatch Encoder:}
Each layer contains two transformation matrices, Q and K, implemented using PyTorch's Linear layer with tensor reshaping for multi-head attention. The value matrix V for feature embedding uses PyTorch's native Embedding layer combined with Linear for feature selection.
\textbf{Reconstruction Layer:}
The final layer of SimAD uses a linear transformation to reconstruct features into the original time-series data, recovering all channels of every patch.
\textbf{ContrastFusion Projection Head:}
A two-layer feedforward network (FFN) consisting of Linear \(\rightarrow\) ReLU \(\rightarrow\) Linear is used. Note that low-dimensional features for each patch must be derived concurrently.
\textbf{Anomaly Score Construction:}
During inference, ensure alignment between the two loss terms. We use linear interpolation to match the length of the cosine similarity score to the MSE score.}

\subsection{Complexity Analysis}
Assume that batch size is denoted by $B$, time series length $T$, channels $C$, and patch length $P$, the number of patches is $M = \frac{T}{P}$, the shape of the input can be denoted as $(B, M, P \times C)$. With hidden dimension $D$, the intermediate feature shape is $(B, M, D)$. At this point, the computational bottleneck lies in the Transformer's self-attention and FFN. The complexity of self-attention is quadratic concerning the sequence length, with a computational complexity of $(B \cdot M^2 \cdot D)$, while the FFN complexity is $(B \cdot M \cdot D^2)$. The overall computational complexity of the model is $B \cdot (M^2 \cdot D + M \cdot D^2)$. When the number of patches is greater than the model dimension, this complexity can be approximated as $(B \cdot M^2 \cdot D)$. At this point, the length of the time series affects the computational efficiency, leading to slower model performance.
% 为了强调算法在工程上的实现细节，我们对模型构建中主要组件进行说明，该部分也可以直接参考我们的代码仓库的文档说明与代码注释。
% 对于改进后的Positional Embedding，我们在时序位置采用余弦位置编码，并且不对通道的位置信息编码。Instance Norm直接采用了原作者代码仓库的实现，没有做改动。对于EmbedPatch Encoder的每一层，都有Q和K两个参数变化矩阵，在实现上我们直接采用PyTorch的Linear，并通过张量形状变化实现多头注意力；同时对于每一层还有一个V矩阵，用于保存嵌入特征，在实现上，我们采用PyTorch官方的原始Embedding实现，并结合了Linear实现嵌入特征选择。在SimAD模型的最后一层，使用原始的Linear将所有特征重构为原始时序特征，并且复原了所有patch的所有通道。对于ConstrastFusion的投影头，采用两层的FFN网络，即Linear，ReLU，Linear的两层网络，需要注意的是，在获得低维特征时，需要同时得到每一个patch的低维特征。在模型推理时，对于异常分数的构建，需要注意两个损失项的长度对齐，我们使用线性插值将余弦相似度分数的长度对齐MSE分数的长度。

% \section{Analyses of UAff and NAff metrics}\label{app:ana_uaff_naff}

\section{Analyses of UAff and NAff metrics}\label{sec:ana_uaff_naff}
% 为了从理论与实践上证明所提出的UAff and NAff metrics的可靠性，我们对它们的数学性质与实际表现分析。
To demonstrate the reliability of the proposed UAff and NAff metrics both theoretically and practically, we analyze their mathematical properties and practical performance.
\subsection{Mathematical analysis of metrics}\label{sec:math_metric}
\newcommand{\tuap}{\textbf{uap}}
\newcommand{\tuaf}{\textbf{uaf}}
\newcommand{\tap}{\textbf{ap}}
\newcommand{\tar}{\textbf{ar}}
\newcommand{\tna}{\textbf{ua}}
\newcommand{\tapb}{\textbf{ap}_b}

\subsubsection{Range analysis}
% 由于NAff是UAff的一种特性情况，因此我们首先分析UAff。

% 为了简洁起见，这里我们将NAff-F1，NAff-Pre，Aff-Pre，Aff-Rec和\({\text{Aff-Pre}_{bias}}\)分别简称为\textbf{naf}, \textbf{uap}, \textbf{ap}，\tar 和\(\tapb\)。

Since NAff is a specific case of UAff, we begin by analyzing UAff. To keep it brevity, we use the following abbreviations: \textbf{uaf} for UAff-F1, \textbf{uap} for UAff-Pre, \textbf{ap} for Aff-Pre, \textbf{ar} for Aff-Rec, and \(\tapb\) for \({\text{Aff-Pre}_{bias}}\).

% 那么，\(\textbf{nap}=\frac{\textbf{ap} - \textbf{ap}_b}{ 1 - \textbf{ap}_b }\)。且理论满足上\(\tap,\tapb \in [0, 1]\)。
% 分别将\textbf{uap}对\textbf{ap}与$\textbf{ap}_b$求偏导，分别得到：

Thus, we have \(\textbf{uap}=\frac{\textbf{ap} - \textbf{ap}_b}{ 1 - \textbf{ap}_b }\).It is theoretically confirmed that both \(\tap\) and \(\tapb\) fall within the range of [0, 1]. The partial derivatives of \textbf{uap} with respect to \textbf{ap} and \(\textbf{ap}_b\) can be given by:
\begin{equation}\label{eq:part_ap}
    \frac{\partial \textbf{uap}}{ \partial \tap } = \frac{1}{1-\tapb} \geq 0
\end{equation}

\begin{equation}\label{eq:pp2}
    \frac{\partial \tuap}{ \partial \tapb } = \frac{\tap - 1}{(1-\tapb)^2} \leq 0
\end{equation}

From the above formulas, it can be seen that \(\tuap\) increases monotonically with \(\tap\) and decreases monotonically with \(\tapb\). When \(\tap = 1\), \(\tuap\) reaches its maximum value of 1. In the case where \(\tap = 0\) and \(\tapb = 1\), \(\tuap\) approaches a minimum value of \(-\infty\). As concluded in \cite{aff}, \(\tapb = 1\) is applicable only when all the data are anomalies. However, the proportion of anomalies in most datasets is typically below 0.5. Hence, it follows that \(\tapb < \frac{1}{2} + \frac{1}{2} \times 0.5^2 = 0.6125\). Furthermore, the minimum value of \(\tapb\) tends to approach 0.5. Given that \(\tapb\) is derived from simulated calculations, we propose a more lenient condition of \(\tapb > 0.4\). In practical scenarios, the range for \(\tapb\) becomes broader, specifically \(\tapb \in (0.4, 0.6125)\). From Eq.\eqref{eq:pp2}, in the case where \(\tap = 0\) and \(\tapb \to 0.6125\), \(\tuap\) tends towards \(\frac{0 - 0.6125}{1 - 0.6125} = -1.587\). Therefore, it can be concluded that \(\tuap\) falls within the range of \((-1.587, 1]\).
% 接下来我们分析\(\tuaf\)，因为\(\tar \in [0,1]\)，根据Eq.\eqref{eq:aff_f1}，容易知道，当\(\tuap \in [0, 1]\)时，有\(\tuaf_{max} = 1\)。
% 当\(\tapb \in (-1.587, 0)\)时，计算\tuaf 的最小值，这一问题可以转换为：当\(\tapb \in (0, 1.587)\)时，计算\tuaf 的最大值的相关数。

Next, let us proceed with the analysis of \(\tuaf\) next. Given that \(\tar\) falls within the range of \([0,1]\), as indicated by Eq. \eqref{eq:aff_f1}, it becomes evident that when \(\tuap\) is within the range of \([0, 1]\), the maximum value of \(\tuaf\) is 1. In the case where \(\tapb\) falls within the interval of \((-1.587, 0)\), it becomes necessary to determine the minimum value of \(\tuaf\). This problem can be reformulated as the search for the corresponding \(\tapb\) within the range of \((0, 1.587)\) that yield the maximum \(\tuaf\).
% 因此，\(\tuaf_{min} = -\frac{2 \times 1.587 \times 1}{1.587 + 1} = -1.227\)。
% 因此，\textbf{UAff-F1}的范围是\((-1.227,1]\)。
% 由于\textbf{NAff-F1}是\textbf{UAff-F1}的一种特殊情况，根据上述分析，易知其范围为\([-1,1]\)

The minimum value of \(\tuaf\) can be obtained by computing \(-\frac{2 \times 1.587 \times 1}{1.587 + 1} = -1.227\). Consequently, the range of \textbf{UAff-F1} is \((-1.227,1]\). Based on the above analysis, it can be concluded that the range of \textbf{NAff-F1} is \([-1,1]\) since it is a specific case of \textbf{UAff-F1}.

\subsubsection{Distinctiveness of metrics}
% 我们简要分析UAff与NAff指标相比原始的Aff指标区分度更高。我们的设计初衷是在同等Aff-Pre的情况下，UAff与Aff的比值始终大于1或小于1。此时只取决于Aff-Pre。因此问题转换为，证明：\( \frac{\tuap}{\tap} \geq 1 \ \text{or}\  \frac{\tuap}{\tap} \leq 1\)。

It can be briefly analyzed that UAff provides greater discrimination than the original Aff. The design intention is to ensure that the ratio of UAff to Aff, \(\frac{\tuap}{\tap}\), is always greater than 1 or less than 1 in the case of equal Aff-Pre. In this case, the ratio depends solely on Aff-Pre. Therefore, the problem can be reformulated as proving: \( \frac{\tuap}{\tap} \geq 1 \ \text{or}\  \frac{\tuap}{\tap} \leq 1\).

By calculating the partial derivative of \(\mathcal{D}=\frac{\tuap}{\tap} = \frac{\tap - \tapb}{\tap (1- \tapb)}\) with respect to \(\tap\), we obtain:

\begin{equation}
    \frac{\partial \mathcal{D}}{\partial \tap} = \frac{\tapb (1-\tap)}{1-\tapb} \geq 0.
\end{equation}

Hence, when \(\tap = 1\), \(\mathcal{D}_{max} = 1\), indicating that \(\mathcal{D} \leq 1\), which aligns with our objective. Based on results in Table \ref{tab:comparison_avg}, it is evident that the random algorithm attains the 9th position in Aff-F1, whereas it only achieves the 19th and 15th positions in UAff-F1 and NAff-F1, respectively. This observation highlights that the enhanced metrics can provide a more precise evaluation of the algorithm's actual performance. The rankings strongly suggest that the UAff-F1 and NAff-F1 metrics offer a superior assessment, compared to the original Aff-F1.

\subsection{Bias data of Affiliation precision}\label{sec:bias_data_ap}

% 表\ref{tab:bias}展示了不同数据集的实际bias与ideal bias，其中理论ideal bias是假设测试中只有一段异常，根据异常率与\cite{aff}中的理论分析得到的。注意到，两者bias的差距在可接受的范围，因此如果假如实际情况难以得到真实bias，也可以通过估计异常率计算理论偏差得到UAff指标。

Table \ref{tab:data_set} presents various datasets' actual and ideal biases. The theoretical ideal bias is determined by assuming the presence of only one anomaly segment in the test data, considering the anomaly rate and theoretical analysis from \cite{aff}. It is noted that the disparity between two biases falls within an acceptable range. In practical scenarios where obtaining the true bias is challenging, the UAff can still be computed by estimating the anomaly ratio and calculating the theoretical bias.
\begin{equation}
    \textbf{Bias}_{ideal} = \frac{1}{2} + \frac{1}{2} \cdot {(\text{AnomalyRatio})}^2
\end{equation}
% 因为在实际情况下，异常的发生比例往往是很低的，在实际情况下，大部分的异常比例只有不足10\%，因此理论上，\(\textbf{Bias}_{ideal}<0.505\)，因此我们在实践上设置\(\beta = \textbf{Aff-Pre}_{Bias}= 0.5\)作为NAff的标准参数。尽管这一常数设置与公开数据集的异常比例略有不同，但是我们不难发现，我们使用的真实数据集，如Swan，的最高异常率是32.6，但是此时\(\textbf{Bias}_{ideal}=0.5531\)，这种情况是是非少见的，同时，即使是这种情况下，将\(\beta\)设置为0.5并不会严重影响我们对不同方法的判断。因此我们在实践中设置\(\beta = 0.5\)是合理的；更进一步的，设置\(\beta = 0.5\)而不是0.55也更加符合实际场景与人类认知，并且更加符合我们的指标设计初衷。
As the occurrence ratio of anomalies is typically low in real-world scenarios, with most anomalies accounting for less than 10\% of the data, theoretically, \(\textbf{Bias}_{ideal}<0.505\). Therefore, in practice, we set \(\beta = \textbf{Aff-Pre}_{Bias}= 0.5\) as the standard parameter for NAff. Despite the slight variation from the anomaly rates in publicly available datasets, it is important to note that in the case of real datasets such as Swan, the highest anomaly ratio is 32.6\%, resulting in \(\textbf{Bias}_{ideal}=0.5531\). This scenario is not uncommon, and even in such cases, setting \(\beta\) to 0.5 does not significantly impact the evaluation of different methods. Therefore, setting \(\beta = 0.5\) in practice is deemed reasonable. Furthermore, setting \(\beta = 0.5\) instead of 0.55 aligns better with real-world scenarios and human cognition, and is more consistent with the original design intent of our metrics.

\subsection{Experimental analysis of UAff and NAff metrics}\label{sec:exp_ana_metric}
Our previous proof compared UAff and NAff with the original Aff, providing empirical and theoretical analyses in Appendix \ref{sec:ana_uaff_naff}.
We drew inspiration from the ranking-based approach \cite{VUS}. To highlight the strengths and weaknesses of different metrics, we designed a new experiment.

\textbf{Experiment Design}: We assume a time series of length 1000 with AnomSeq anomaly segments, with a minimum length of MinLen and a maximum length of MaxLen. Using a Random algorithm, we generate uniformly distributed anomaly scores, and then create models with varying performance levels, such as M60, which correctly predicts 60\% of normal and anomalous points. Correct anomaly predictions receive a higher anomaly score using a random value $U$, \textit{i.e.}, $U*0.1+0.9$. Incorrect normal predictions as anomalies also receive higher scores. In other cases, the model outputs an anomaly score of 0. This experiment is repeated 20 times, and the average result is taken.

\begin{table*}[!htbp]
  \centering
  \caption{Comparison of different metrics on three different synthetic datasets. F1PA is a point-adjusted F1 score.}
  % \vspace{-10pt}
    \setlength{\tabcolsep}{2pt}
  \resizebox{0.7\linewidth}{!}{
\begin{tabular}{|c|c|cccccccccccccc|}
\toprule
\textbf{Demo} & \multicolumn{1}{c}{\textbf{Method}} & \textbf{F1} & \textbf{Acc} & \textbf{Pre} & \textbf{Rec} & \textbf{F1PA} & \textbf{AUC} & \textbf{Aff-Pre} & \textbf{Aff-Rec} & \textbf{Aff-F1} & \textbf{NAff-F1} & \textbf{R\_A\_R} & \textbf{R\_A\_P} & \textbf{V\_ROC} & \textbf{V\_PR} \\
\midrule
      & Random & 5.48  & 90.54  & 4.60  & 4.72  & 26.31  & 50.25  & 49.91  & 87.98  & 63.50  & 0.06  & 77.35  & 37.85  & 75.62  & 36.02  \\
      & M10   & 2.30   & 90.05 & 0.70   & 0.68  & 3.45  & 9.78  & 46.37 & 82.99 & 59.38 & -12.08 & 74.52 & 36.27 & 69.24 & 34.37 \\
\textbf{Demo1} & M60   & 7.33  & 90.74  & 7.30  & 7.37  & 39.57  & 59.75  & 54.60  & 88.18  & 67.18  & 15.67  & 74.21  & 38.71  & 73.24  & 36.57  \\
AnomSeq=5 & M70   & 10.63  & 91.00  & 10.70  & 10.57  & 48.15  & 69.51  & 52.71  & 91.76  & 66.88  & 9.47  & 74.08  & 43.20  & 74.42  & 41.50  \\
MinLen=10 & M80   & 17.90  & 91.77  & 18.00  & 17.83  & 65.04  & 80.32  & 60.68  & 97.02  & 74.55  & 34.03  & 72.01  & 43.93  & 73.45  & 42.34  \\
MaxLen=12 & M90   & 30.83  & 93.11  & 30.70  & 31.00  & 72.16  & 90.04  & 66.80  & 98.70  & 79.56  & 49.28  & 70.70  & 49.95  & 73.46  & 49.18  \\
      & M95   & 50.95  & 95.12  & 50.70  & 51.27  & 78.37  & 95.08  & 75.74  & 99.50  & 85.96  & 67.53  & 70.10  & 57.06  & 73.45  & 57.52  \\
      & M100  & 99.18  & 99.92  & 100.00  & 98.40  & 100.00  & 100.00  & 100.00  & 99.99  & 99.99  & 99.99  & 69.89  & 76.48  & 73.55  & 79.23  \\
\midrule
      & Random & 4.98  & 90.42  & 4.50  & 4.47  & 57.40  & 49.63  & 50.39  & 97.82  & 66.45  & 1.18  & 67.02  & 14.05  & 65.29  & 13.84  \\
      & M10   & 2.29  & 90.06 & 0.80   & 0.80   & 23.36 & 10.65 & 48.99 & 95.60  & 64.71 & -3.73 & 52.15 & 10.23 & 47.37 & 10.26 \\
\textbf{Demo2} & M60   & 8.47  & 90.73  & 8.10  & 8.01  & 65.13  & 60.45  & 51.17  & 98.54  & 67.28  & 4.26  & 66.11  & 15.77  & 65.45  & 15.57  \\
AnomSeq=1 & M70   & 11.28  & 91.11  & 11.30  & 11.26  & 68.92  & 70.41  & 54.26  & 99.25  & 70.07  & 14.81  & 69.77  & 18.01  & 69.83  & 17.79  \\
MinLen=50 & M80   & 18.66  & 91.81  & 18.80  & 18.53  & 70.99  & 79.97  & 57.17  & 99.54  & 72.54  & 24.20  & 72.56  & 23.88  & 73.50  & 24.00  \\
MaxLen=60 & M90   & 32.38  & 93.20  & 32.60  & 32.17  & 74.68  & 89.97  & 64.07  & 99.80  & 77.91  & 42.83  & 74.93  & 33.12  & 76.66  & 33.63  \\
      & M95   & 47.65  & 94.72  & 48.00  & 47.34  & 79.29  & 94.99  & 72.65  & 99.90  & 84.06  & 61.87  & 77.05  & 45.98  & 79.01  & 46.70  \\
      & M100  & 98.83  & 99.88  & 100.00  & 97.81  & 100.00  & 100.00  & 100.00  & 100.00  & 100.00  & 100.00  & 78.15  & 80.18  & 80.47  & 82.45  \\
\midrule
      & Random & 8.99  & 68.11  & 31.50  & 5.24  & 94.59  & 50.05  & 55.52  & 98.24  & 70.88  & 19.01  & 58.19  & 38.59  & 57.72  & 38.42  \\
    
      & M10   & 1.39  & 65.19 & 3.90   & 0.65  & 74.13 & 10.17 & 37.36 & 90.23 & 52.78 & -38.78 & 25.81 & 23.96 & 25.12 & 24.04 \\
\textbf{Demo3} & M60   & 11.01  & 68.16  & 39.50  & 6.40  & 95.28  & 59.59  & 61.33  & 98.67  & 75.47  & 35.14  & 63.53  & 45.90  & 63.47  & 45.89  \\
AnomSeq=1 & M70   & 14.77  & 69.19  & 53.30  & 8.58  & 96.36  & 70.33  & 69.39  & 99.01  & 81.50  & 54.99  & 71.59  & 55.12  & 71.62  & 55.02  \\
MinLen=300 & M80   & 18.32  & 71.28  & 64.40  & 10.68  & 97.13  & 79.68  & 76.43  & 99.15  & 86.23  & 68.40  & 78.64  & 64.09  & 78.77  & 64.13  \\
MinLen=350 & M90   & 22.22  & 72.69  & 78.00  & 12.96  & 98.20  & 89.84  & 85.52  & 99.38  & 91.88  & 82.62  & 86.43  & 77.39  & 86.69  & 77.47  \\
      & M95   & 25.05  & 73.51  & 88.50  & 14.59  & 99.06  & 94.72  & 92.28  & 99.43  & 95.68  & 91.20  & 90.13  & 85.88  & 90.42  & 86.00  \\
      & M100  & 28.36  & 74.72  & 100.00  & 16.53  & 100.00  & 100.00  & 100.00  & 99.54  & 99.77  & 99.77  & 93.93  & 96.02  & 94.30  & 96.33  \\
\bottomrule
\end{tabular}%
}
\label{tab:exp_metric}
\end{table*}

To investigate different scenarios, we designed three demos in Table \ref{tab:exp_metric}. Demo1 denotes many anomalies with short intervals. Demo2 denotes a few anomalies with long intervals. Demo3 denotes a few anomalies with extremely long intervals. The score threshold is set to 95\%. From the results in this table, we have the following observations: 
\begin{enumerate}
    \item \textbf{NAff-F1 is better than Aff-F1}: In Demo1 and Demo2, the Aff-F1 for the Random model are 63.50 and 66.45, which are misleading. Our NAff-F1 scores are 0.06 and 1.18, indicating that NAff-F1 is more robust and better at distinguishing random models. Besides, NAff-F1 has a wider value interval, enhancing its discriminative power.
    \item \textbf{Weak model evaluation}: For the M10 model, which performs worse than a random model, an effective metric should be able to distinguish it from random model. Metrics that achieve this include AUC, NAff-F1, R\_A\_R, and V\_ROC. However, the original Aff falls short, assigning a score of 59.38 to M10 in Demo1.
    \item \textbf{Scoring misleadingness}: F1PA is a point-adjusted F1 score that can be misleading when the anomaly interval is long, as seen in Demo3. In such cases, even a random model can achieve a high score of 94.59. Our metrics provide meaningful differentiation, with scores of 19.01, -38.78, and 35.14 for Random, M10, and M60, respectively, distinguishing between these models clearly.
    \item \textbf{Parameter selection}: R\_A\_R requires selecting an interval length, which can lead to inaccurate evaluations when anomalies are short, as demonstrated in Demos 1 and 2. In Demo 1, the R\_A\_R scores for Random, M10, and M60 are all above 74, showing that the metric fails to distinguish between these models. This issue is mitigated with longer anomaly intervals, as seen in Demo 3, where M10 has a R\_A\_R score of 25.81, significantly lower than 50. In contrast, our metric does not require parameter selection and provides accurate evaluations across various scenarios.
    \item \textbf{Weakness of point-based evaluation}: F1, Acc, Pre, and Rec are point-based metrics that are easily influenced by threshold selection and strict label matching, resulting in suboptimal results, even for perfect models like M100. In contrast, event-based metrics are not affected.
\end{enumerate}
Based on these studies, we demonstrate the advantages of our metric over previous ones. We recommend using NAff-F1 and VUS\_PR in the future, and other VUS series metrics can also be used when the anomaly interval is long.

\section{Investigation of existing evaluation metrics}\label{sec:Investigation_metrics}

\subsection{Flaws in previous metrics}\label{sec:flaw}

% 图\ref{fig:metric_demo}展示了一个人工的时序异常检测案例。第一行序列是时序数据的真实标签。Random1,Random2,Pred1,Pred2分别是两种随机算法的异常分数和其他算法的异常分数。以PA结尾表示点调整之后的标签，如果一段异常中有一部分被模型预测出来，则视作整段异常被识别。

% Random1将所有时刻都预测为异常，Random2均匀地随机预测异常。然而二者的点调整F1分数（F1PA）都很高，这表明F1PA的鲁棒性并不好。
% F1分数同样会存在这种问题，他只适合评估较短的区间异常（如最后一个异常）。但是面对非随机预测时，他的结果是可参考的。
% Pred1是这里面最佳的模型，但是他的Aff-F1分数低于Random1，这是因为原始的Aff-F1指标并不会给错误预测惩罚。而改进后的UAff-F1能更好的比较Pred1与Random1的表现。同时，Pred1比Pred2效果更好，但是两者的Aff-F1差距很小，甚至都不如Random1。而此时UAff-F1能够更好的比较两者，分别是0.1818和0.0057，并且给Random1和Random2更低的分数。

% 综上，我们认为F1是非随机算法的可参考指标，F1PA容易产生虚假的进步，Aff-F1主要问题是区分度不足，以及对随机算法评估不准。最后，改进后的UAff-F1更推荐使用。

\begin{figure}[ht]
    \centering
    \includegraphics[width=1.\linewidth]{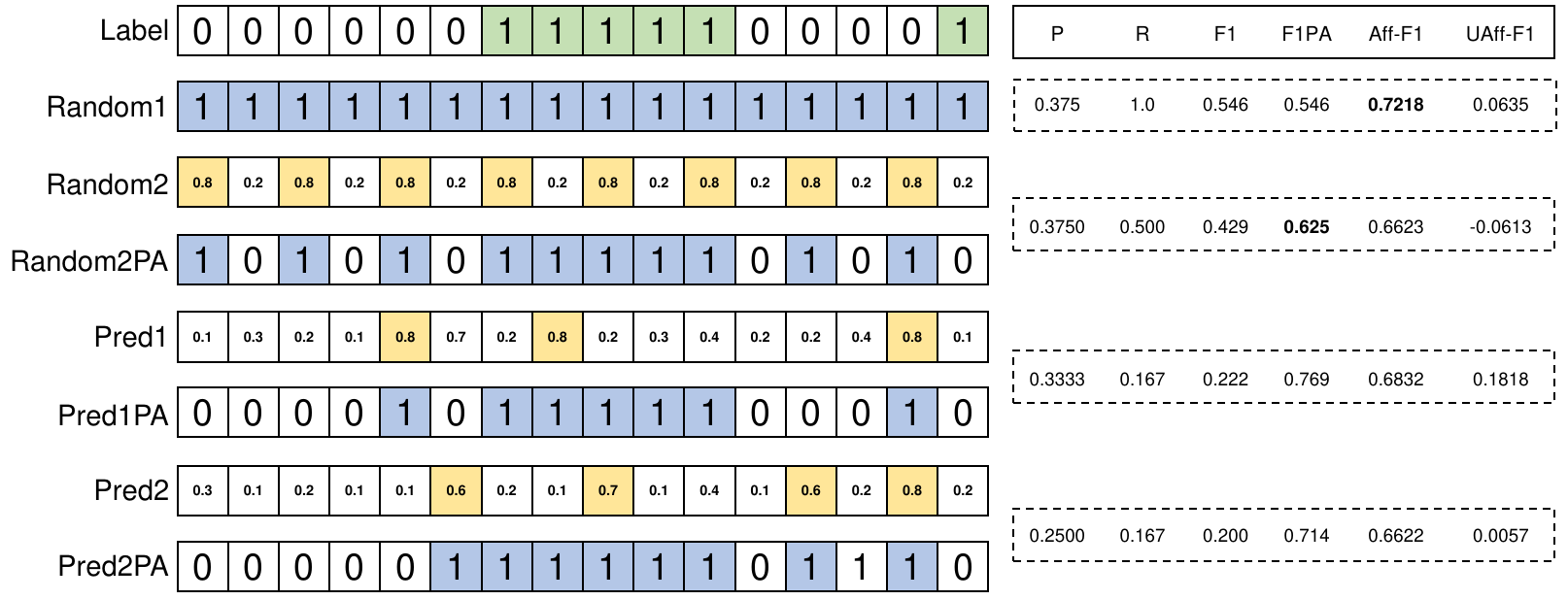}
    \caption{A case of artificial data illustrating the shortcomings of different metrics.}
    \label{fig:metric_demo}
\end{figure}

Fig. \ref{fig:metric_demo} illustrates an artificial example of time series anomaly detection. The top row represents the true labels of the time series data. Random1, Random2, Pred1, and Pred2 denote the anomaly scores generated by two random and other algorithms. The labels ending with ``PA" indicate point adjustment, where if the model predicts any part of an anomaly segment, the entire segment is considered an identified anomaly.

Random1 predicts all timestamps as anomalies, while Random2 uniformly and randomly predicts anomalies. However, both algorithms achieve high point-adjustment F1 scores (F1PA), indicating that F1PA lacks robustness in this case.

The F1 score can exhibit this issue, as it is more suitable for evaluating shorter interval anomalies (such as the last). However, its results can still be informative and provide some reference when facing non-random predictions.

Pred1 is the top-performing model among the mentioned ones. However, its Aff-F1 score is lower than that of Random1. This is because the original Aff-F1 metric does not account for false predictions. However, the enhanced UAff-F1 metric provides a better basis for comparing the performance of Pred1 and Random1. Additionally, Pred1 outperforms Pred2, although the difference in their Aff-F1 scores is small and even lower than that of Random1. In this case, the UAff-F1 metric can provide a more effective comparison between the two models, yielding scores of 0.1818 and 0.0057 for Pred1 and Pred2, respectively, while assigning lower scores to Random1 and Random2.

Based on the above analysis, we can conclude that F1 can serve as a valuable reference metric for non-random algorithms. F1PA may result in false improvements, whereas Aff-F1 lacks sufficient discriminative power and provides inaccurate evaluations for random algorithms. Ultimately, we recommend using the enhanced UAff-F1 metric for a more accurate performance assessment.

\subsection{Limitations \& Advantages of metrics}

We referenced the study VUS \cite{VUS} and analyzed the attributes of various metrics, focusing on discrimination and semantics as key evaluation criteria. It is important to note the following properties of existing evalutation metrics:

\begin{enumerate}
    \item Non-Threshold: This indicates that the metric does not require the setting of a threshold.
    \item Sequence: This signifies that the metric can effectively evaluate sequential anomalies.
    \item Parameter-Free: This means the metric does not require additional parameters for its evaluation.
    \item Discrimination: This attribute reflects the metric's ability to differentiate between random, weak, and strong models. For example, our research found that the F1PA metric exhibits low discrimination, as even a random model could achieve a score of 94.
    \item Semantics: This denotes that the metric is associated with human-understandable meaning. For instance, the AUC metric provides clear semantics for point anomalies or short anomalies. Similarly, the Aff metric offers semantic relatedness by indicating how closely the detected anomalies align with the actual anomaly range—a higher score suggests a closer match.
\end{enumerate}

\begin{table}[]
    \centering
    % Table generated by Excel2LaTeX from sheet 'Sheet2'
    \caption{The properties of different metrics.}
    \resizebox{\linewidth}{!}{
\begin{tabular}{l|l|l|l|l|l}
\toprule
Metric/Property & Score Threshold & Sequence-adapted & Parameter-free & Discrimination & Semantics \\
\midrule
ACC   & \ding{55} & \ding{55} & \ding{55} & \ding{55} & \ding{55} \\
F1    & \ding{55} & \ding{55} & \ding{55} & \ding{55} & \ding{55} \\
F1PA  & \ding{55} & \ding{51} & \ding{55} & \ding{55} & \ding{55} \\
AUC   & \ding{51} & \ding{55} & \ding{55} & \ding{51} & \ding{51} \\
Aff-F1 & \ding{51} & \ding{51} & \ding{51} & \ding{55} & \ding{51} \\
UAff-F1 & \ding{51} & \ding{51} & \ding{51} & \ding{51} & \ding{51} \\
R\_A\_R & \ding{51} & \ding{51} & \ding{55} & \ding{51} & \ding{51} \\
R\_A\_P & \ding{51} & \ding{51} & \ding{55} & \ding{51} & \ding{51} \\
V\_ROC  & \ding{51} & \ding{51} & \ding{51} & \ding{51} & \ding{51} \\
V\_PR  & \ding{51} & \ding{51} & \ding{51} & \ding{51} & \ding{51} \\
\bottomrule
\end{tabular}%
}
    
    \label{tab:metric_props}
\end{table}

These attributes are critical for evaluating the effectiveness of different metrics in anomaly detection. 
Table \ref{tab:metric_props} summarizes the properties of the different metrics.

1. Affs (including the metrics we proposed) are based on local affiliation, which enhances their ability to evaluate a model's early warning and post-alert performance.
2. The VUS series of metrics are designed based on AUC, with a focus on evaluating a model's tendency towards anomalies and normal events. In contrast, our metrics and VUS represent two distinct approaches to evaluation.
3. For a comprehensive comparison of other metrics, please see Appendix~\ref{sec:exp_ana_metric}. Below, we summarize the key conclusions:
\begin{enumerate}
    \item NAff-F1 has a better discriminative ability than Aff-F1, which was our initial design goal.
    \item ACC, F1PA, and Aff exhibit insufficient weak model evaluation capabilities and cannot effectively evaluate random models and weak models.
    \item Scoring misleadingness: F1PA has the strongest misleading score.
    \item Refer to the table above; some metrics require additional parameters. As \citet{VUS} suggests: ``We observe that this change implies a larger variation for several threshold-based measures. Thus, the latter confirms the limitations and the non-robustness of threshold-based measures to the anomaly cardinality ratio.'' Threshold-based metrics have certain limitations.
    \item Weakness of point-based evaluation: Previous research \cite{sv_wuCurrentTimeSeries2022} has shown that current datasets may have unreasonable labels. When using point-based metrics, the weaknesses of the evaluation metrics are magnified. Strict label matching can lead to inaccurate evaluations.
\end{enumerate}

These observations highlight the complexity of real-world scenarios and the varied emphasis of different metrics. In the TSAD community, interval evaluation is currently deemed more crucial, which is why we prioritize Aff and VUS metrics. As the field of TSAD is diverse, no single evaluation metric is universally appropriate, and the choice of metric should be made carefully for each case \cite{Navigating_the_metric_maze}.

In summary, it is currently recommended that ``when publishing results in TSAD research, multiple metrics should be included, and both the code and the anomaly scores should be made available to facilitate easy comparison with any evaluation metric." \cite{Navigating_the_metric_maze} A more comprehensive and systematic investigation would require a broader research scope, which is beyond the focus of this paper. To contribute to the advancement of TSAD, we have made our model code, weights, and training and testing scripts available in order to reproduce results reported in our work.

\section{Detailed results of experiments}\label{sec:res}

% 表\ref{tab:ucr_cnt}中展示了SimAD,AnomTrans,DCdetector三个模型在UCR数据集上的表现。综合考虑模型在所有子数据集的表现，可以发现在大部分评价指标下SimAD模型的F1分数均高于基线模型。此外，通过SimAD模型进行异常检测得到评价指标分数大于0的数据集数量最多，且去除评价分数小于0的数据集后SimAD的各评价指标综合分数远高于其他基线模型。这说明SimAD能够检测出更多不同的异常。消融实验具体结果如表\ref{tab:abl_full}所示。

Table \ref{tab:ucr_cnt} presents the performance of three models, namely SimAD, AnomTrans, and DCdetector, on the UCR dataset. Considering the overall performance across all sub-datasets, it is evident that the SimAD model consistently achieves higher F1 scores compared to the baseline models. Additionally, the SimAD model for anomaly detection demonstrates the highest number of datasets with positive evaluation scores. By excluding datasets with negative evaluation scores, SimAD exhibits significantly higher composite scores for all evaluation metrics than the other baseline models. These findings indicate that SimAD can detect a wider range of anomalies. The specific results of the ablation experiments are presented in Table \ref{tab:abl_full}.

\begin{table}[htbp]
  \centering
  \caption{Comparison results on UCR datasets.}
  \resizebox{1.\linewidth}{!}{
    \begin{tabular}{c|c|cccccc}
    \toprule
    \multicolumn{2}{c|}{\textbf{Datasets}} & \multicolumn{6}{c}{\textbf{UCR}} \\
    \midrule
    \textbf{Metrics} & \textbf{Methods} & \textbf{F1} & \textbf{Aff-Pre} & \textbf{Aff-Rec} & \textbf{Aff-F1} & \textbf{UAff-F1} & \textbf{NAff-F1} \\
    \midrule
    \multirow{3}[2]{*}{\textbf{Count}} & AnomTrans & 147 & 244 & 244 & 244 & 141 & 148 \\
       & DCdetector & 149 & 206 & 206 & 206 & 127 & 144 \\
       & Ours & 240 & 240 & 240 & 240 & 173 & 177 \\
    \midrule
    \multirow{3}[2]{*}{\textbf{Avg.+}} & AnomTrans & 1.17  & 50.66  & 98.94  & 66.88  & 9.41  & 10.88  \\
       & DCdetector & 1.66  & 50.90  & 99.96  & \uline{67.41 } & 6.71  & 5.52  \\
       & Ours & 14.99  & 57.83  & 99.91  & \textbf{72.52 } & 35.16  & 34.38  \\
    \midrule
    \multirow{3}[2]{*}{\textbf{Avg.}} & AnomTrans & 1.15  & 50.39  & 97.73  & 66.55  & 1.30  & 1.23  \\
       & DCdetector & 1.61  & 50.63  & 83.30  & \uline{67.06 } & 2.01  & 2.64  \\
       & Ours & 14.99  & 57.83  & 97.47  & \textbf{72.52 } & 19.53  & 1.00  \\
    \bottomrule
    \end{tabular}%
    }
  \label{tab:ucr_cnt}%
\end{table}%

\subsection{Additional results of comparison}\label{sec:add_compa}

Since the VUS series evaluates anomaly detection models differently, we included four VUS metrics in the comparative experiments: Range-AUC-ROC (R\_A\_R), Range-AUC-PR (R\_A\_P), V\_ROC, and V\_PR. The metrics V\_ROC and V\_PR assess performance based on the surfaces created by ROC and PR curves. As shown in Table \ref{tab:compa_vus} our method achieves the best performance on MSL, SMAP, SWaT, and WADI, and demonstrates strong competitiveness on PSM and Swan. This can be attributed to the anomaly ratios of dataset and evaluation metrics. Specifically, PSM and Swan have a higher ratio of anomalies, which may impact performance. Overall, SimAD is the most stable, performing consistently well across multiple metrics and datasets, compared with other methods.

\begin{table*}[htbp]
  \centering
  % \caption{不同方法在VUS指标上的表现}
    \caption{Performance of different methods on VUS metrics.}
    \resizebox{\linewidth}{!}{
    % Table generated by Excel2LaTeX from sheet 'WWW'
\begin{tabular}{c|cccc|cccc|cccc|cccc|cccc|cccc}
\toprule
Datasets & \multicolumn{4}{c|}{MSL}      & \multicolumn{4}{c|}{SMAP}     & \multicolumn{4}{c|}{SWaT}     & \multicolumn{4}{c|}{WADI}     & \multicolumn{4}{c|}{PSM}      & \multicolumn{4}{c}{NIPS-TS-Swan} \\
\midrule
Methods & R\_A\_R & R\_A\_P & V\_ROC & V\_PR & R\_A\_R & R\_A\_P & V\_ROC & V\_PR & R\_A\_R & R\_A\_P & V\_ROC & V\_PR & R\_A\_R & R\_A\_P & V\_ROC & V\_PR & R\_A\_R & R\_A\_P & V\_ROC & V\_PR & R\_A\_R & R\_A\_P & V\_ROC & V\_PR \\
\midrule
Random & 57.27  & 14.15  & 57.56  & 14.66  & 57.59  & 17.20  & 52.71  & 14.14  & 57.29  & 16.26  & 51.21  & 12.74  & 56.81  & 7.70  & 58.35  & 8.58  & 55.72  & 33.01  & 55.09  & 32.59  & 38.17  & 35.43  & 86.19  & 83.58  \\
LOF   & 62.15  & 17.29  & 60.85  & 17.19  & 58.00  & 18.63  & 57.56  & 18.59  & 66.94  & 42.33  & 66.67  & 41.88  & 46.17  & 7.88  & 45.23  & 7.73  & 81.36  & 65.24  & 80.42  & 64.55  & 90.34  & 89.83  & 89.42  & 88.56  \\
IForest & 65.38  & 18.37  & 64.74  & 18.29  & 59.30  & 16.46  & 59.31  & 16.45  & 73.25  & 53.11  & 72.29  & 52.07  & 78.99  & 28.47  & 77.89  & 27.02  & 75.81  & 55.92  & 75.19  & 55.50  & 91.88  & 92.00  & 91.27  & 90.98  \\
PCA   & 59.81  & 18.20  & 59.42  & 18.08  & 44.03  & 11.69  & 43.90  & 11.69  & 61.61  & 44.38  & 61.87  & 44.75  & 48.85  & 7.83  & 48.25  & 7.72  & 74.52  & 54.93  & 73.97  & 54.55  & 91.50  & 91.64  & 90.42  & 90.23  \\
Deep SVDD & 60.15  & 19.44  & 59.54  & 19.36  & 42.00  & 11.55  & 41.95  & 11.56  & 50.96  & 56.48  & 50.96  & 56.48  & 57.21  & 54.82  & 57.16  & 54.73  & 54.00  & 66.98  & 53.97  & 66.92  & 82.27  & 92.73  & 80.89  & 91.74  \\
USAD  & 59.59  & 20.00  & 58.86  & 19.77  & 33.76  & 11.27  & 33.78  & 11.28  & 84.71  & 63.41  & 84.38  & 62.74  & 50.82  & 54.31  & 50.78  & 54.24  & 64.47  & 69.55  & 64.02  & 69.18  & 93.70  & 96.14  & 91.40  & 94.50  \\
TCN-ED & 58.98  & 18.20  & 58.62  & 18.13  & 43.93  & 11.73  & 43.79  & 11.73  & 61.53  & 57.95  & 61.53  & 57.94  & 50.86  & 54.31  & 50.83  & 54.24  & 58.20  & 43.02  & 57.48  & 42.64  & 82.27  & 92.72  & 80.88  & 91.73  \\
COUTA & 61.97  & 18.91  & 61.49  & 18.68  & 44.82  & 11.91  & 44.72  & 11.91  & 72.68  & 25.09  & 72.57  & 25.06  & 46.47  & 30.27  & 45.65  & 29.43  & 64.73  & 43.29  & 64.50  & 43.15  & 83.78  & 93.36  & 82.15  & 92.26  \\
TranAD & 57.77  & 17.67  & 57.25  & 17.55  & 45.27  & 11.98  & 45.15  & 11.98  & 86.59  & 64.29  & 86.25  & 64.19  & 60.27  & 44.03  & 59.94  & 43.65  & 72.85  & 56.28  & 72.42  & 56.02  & 89.09  & 93.61  & 88.01  & 92.59  \\
NCAD  & 67.98  & 21.47  & 67.29  & 21.32  & 48.39  & 13.21  & 48.37  & 13.22  & 76.86  & 44.96  & 76.76  & 44.91  & 45.80  & 8.51  & 44.52  & 8.13  & 67.60  & 48.55  & 67.09  & 48.26  & 87.40  & 94.73  & 85.31  & 93.44  \\
Deep IF & 53.19  & 17.24  & 52.33  & 17.04  & 56.93  & 14.77  & 56.89  & 14.77  & 50.19  & 56.40  & 50.19  & 56.39  & 47.97  & 13.58  & 47.07  & 12.23  & 68.37  & 46.48  & 67.71  & 45.98  & 82.27  & 92.73  & 80.89  & 91.74  \\
AnomTrans & 53.25  & 13.17  & 53.03  & 13.21  & 52.75  & 16.56  & 52.61  & 16.55  & 21.52  & 8.29  & 21.49  & 8.29  & 50.49  & 7.83  & 50.29  & 7.71  & 49.84  & 31.66  & 49.12  & 31.56  & 85.46  & 83.57  & 83.56  & 81.64  \\
TimesNet & 65.41  & 20.95  & 64.63  & 20.67  & 49.01  & 12.93  & 48.83  & 12.91  & 31.27  & 10.31  & 31.10  & 10.29  & 81.75  & 34.76  & 79.80  & 32.62  & 67.22  & 49.17  & 66.37  & 48.05  & 92.61  & 95.33  & 91.58  & 94.30  \\
DCDetector & 52.17  & 14.69  & 52.00  & 14.68  & 60.65  & 16.88  & 60.38  & 16.87  & 50.85  & 14.70  & 50.84  & 14.68  & 51.62  & 8.23  & 51.60  & 8.21  & 52.13  & 32.56  & 51.73  & 32.44  & 87.59  & 85.09  & 85.84  & 83.11  \\
D3R   & 69.74  & 21.04  & 69.02  & 20.93  & 54.98  & 16.56  & 48.67  & 13.23  & 71.03  & 39.91  & 84.38  & 57.27  & 49.88  & 7.49  & 49.25  & 8.42  & 66.47  & 49.44  & 66.92  & 49.38  & 52.36  & 44.87  & 89.27  & 88.29  \\
GPT2-Adapter & 52.97  & 16.79  & 53.57  & 17.38  & 59.90  & 19.03  & 57.31  & 15.78  & 51.11  & 14.10  & 50.01  & 10.29  & 50.98  & 8.67  & 51.32  & 9.52  & 48.24  & 36.56  & 52.45  & 32.79  & 47.37  & 41.59  & 84.31  & 84.23  \\
NPSR  & 67.58  & 19.18  & 66.44  & 19.72  & 47.60  & 14.22  & 41.91  & 11.42  & 87.52  & 67.26  & 84.93  & 65.42  & 87.94  & 54.45  & 88.23  & 56.72  & 72.57  & 58.65  & 70.55  & 52.12  & 53.92  & 47.61  & 90.90  & 90.35  \\
\rv{M2N2}  & 65.82  & 21.69  & 65.40  & 21.55  & 48.80  & 14.36  & 48.74  & 14.36  & 61.41  & 26.33  & 61.14  & 26.16  & 45.07  & 8.15  & 43.99  & 7.98  & 39.19  & 30.22  & 39.20  & 30.10  & 63.95  & 29.51  & 63.69  & 29.15  \\
\rv{AdaMemBLS} & 58.29  & 16.13  & 57.46  & 16.06  & 55.73  & 14.26  & 55.57  & 14.26  & 84.27  & 68.64  & 82.00  & 66.12  & 66.50  & 23.11  & 63.47  & 21.40  & 66.16  & 49.81  & 65.07  & 49.24  & 88.85  & 87.85  & 87.12  & 85.92  \\
\midrule
Ours  & 69.45  & 24.86  & 68.45  & 24.15  & 61.90  & 20.77  & 58.64  & 17.02  & 86.19  & 69.46  & 83.89  & 66.70  & 91.26  & 61.99  & 91.95  & 64.32  & 71.25  & 54.79  & 68.06  & 53.81  & 87.59  & 85.09  & 90.95  & 91.74  \\
\bottomrule
\end{tabular}%
    }
  \label{tab:compa_vus}%
\end{table*}%

\subsection{Additional visualization analysis}\label{sec:vis}
% 图\ref{fig:case_study2}清楚展示了模型使用不同损失函数对预测结果,阈值和异常分数值的影响。从中我们容易推断出我们模型综合使用L2和余弦损失函数比仅使用L2损失函数和仅使用余弦损失函数的性能都要好。图\ref{fig:heat_map_all}展示了模型不同层的相似度分数。在模型的第一层相似度换较为分散。随着层数增加，相似度分数逐渐增大，且注意力逐渐集中。

\begin{figure*}
    \centering
    \subfloat[(Ours) Case 1]{
    \includegraphics[width=0.49\linewidth]{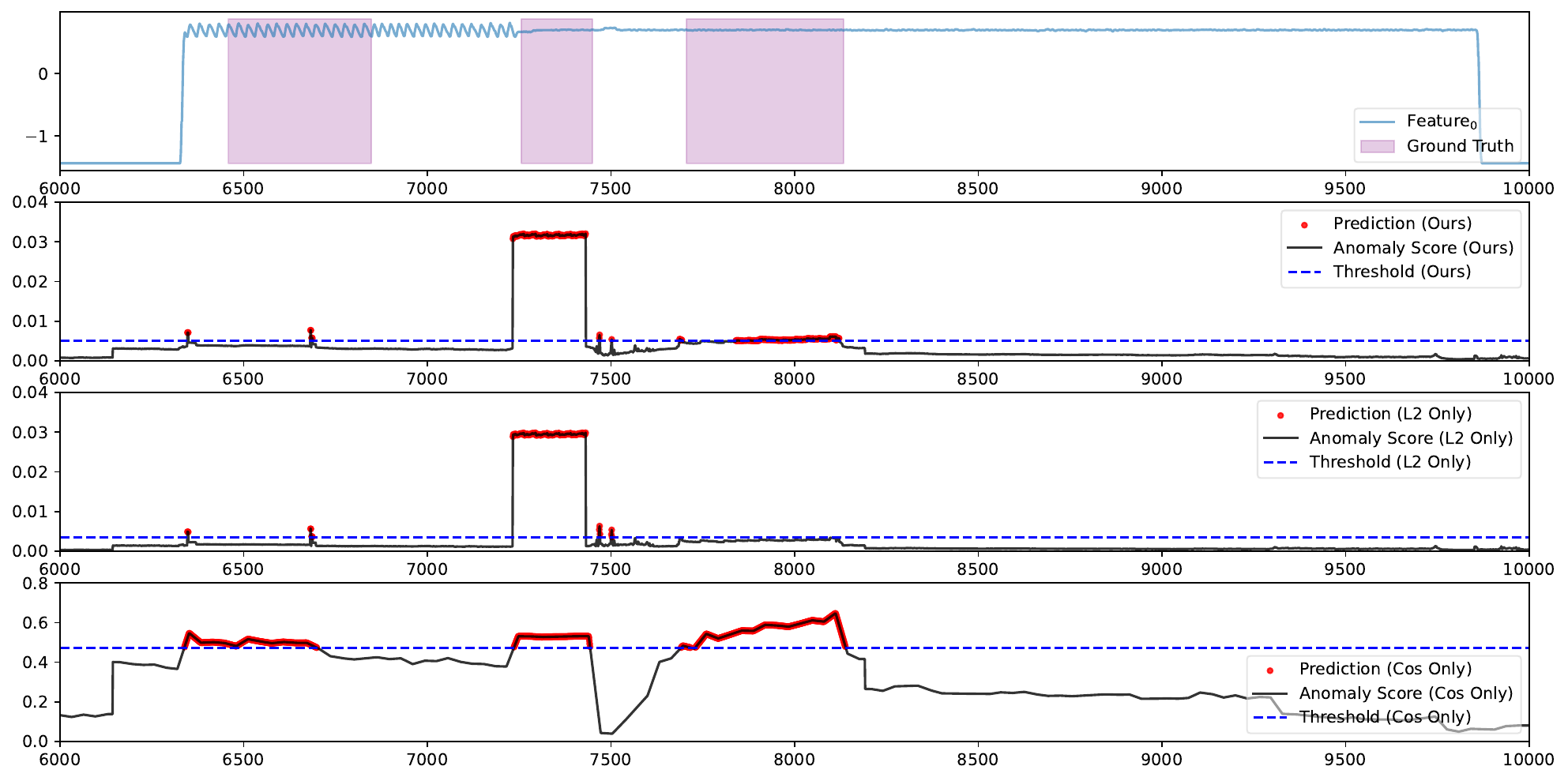}}
    \subfloat[(Ours) Case 2]{
    \includegraphics[width=0.49\linewidth]{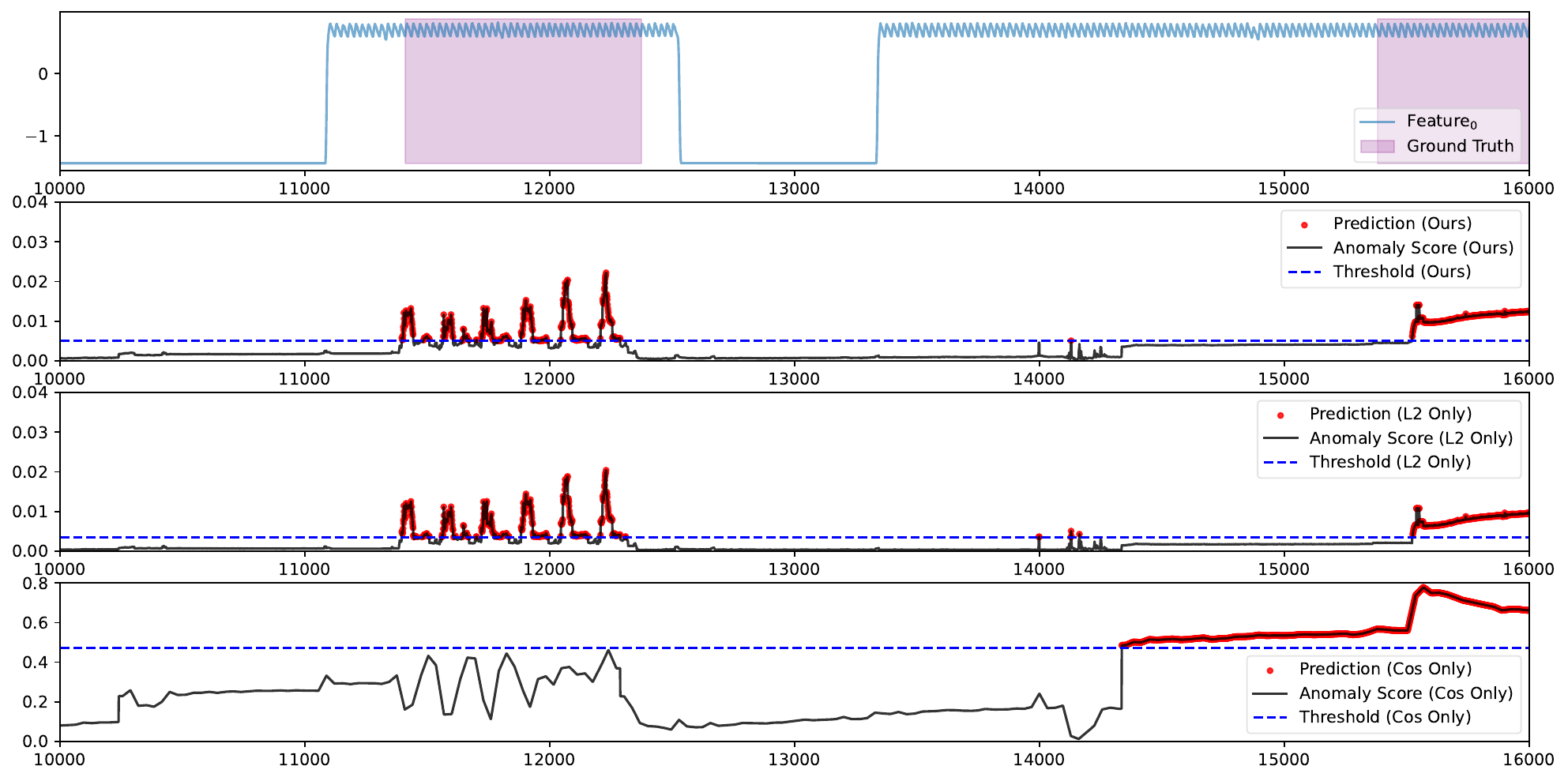}}
    \caption{SimAD's detection performances in real-world.}
    \label{fig:case_study2}
\end{figure*}

Fig. \ref{fig:case_study2} demonstrates the impact of different loss functions on the predicted results, threshold values, and anomaly scores. The figure shows that our model's performance is enhanced when both the L2 and cosine loss functions are employed, as compared to using either the L2 loss function alone or the cosine loss function alone. 

\begin{figure*}
    \centering
    \subfloat[(AnomTrans) Case 1]{\includegraphics[width=0.49\linewidth]{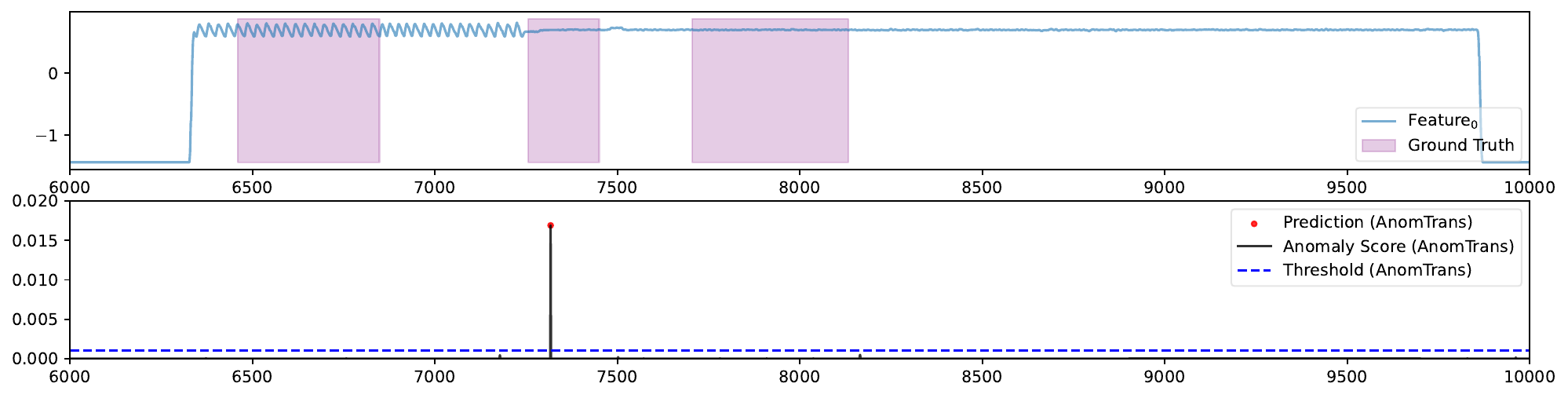}}
    \subfloat[(AnomTrans) Case 2]{\includegraphics[width=0.49\linewidth]{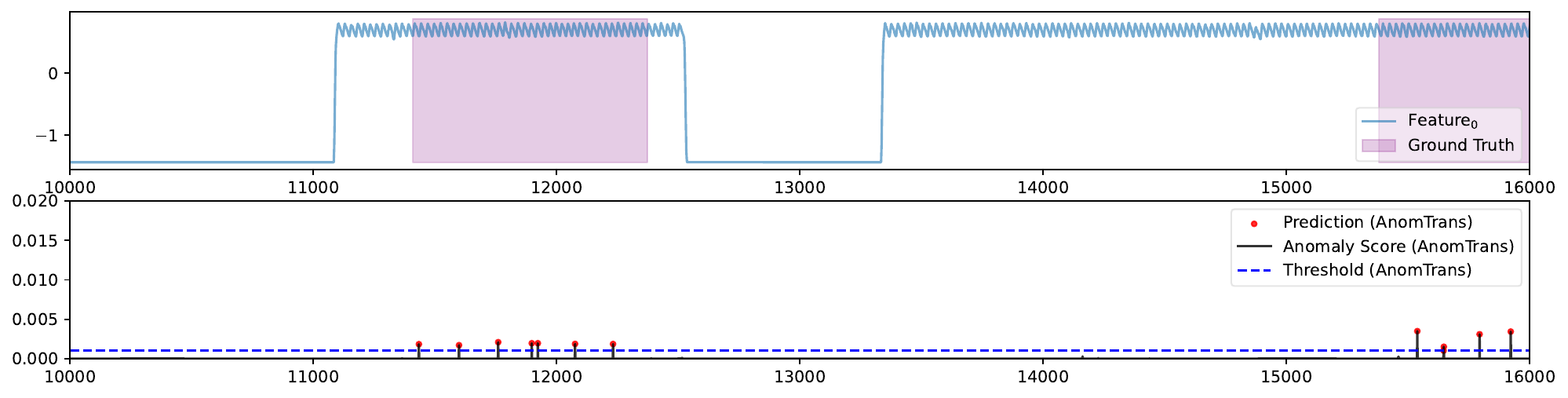}}\newline
    \subfloat[(NPSR) Case 3]{\includegraphics[width=0.49\linewidth]{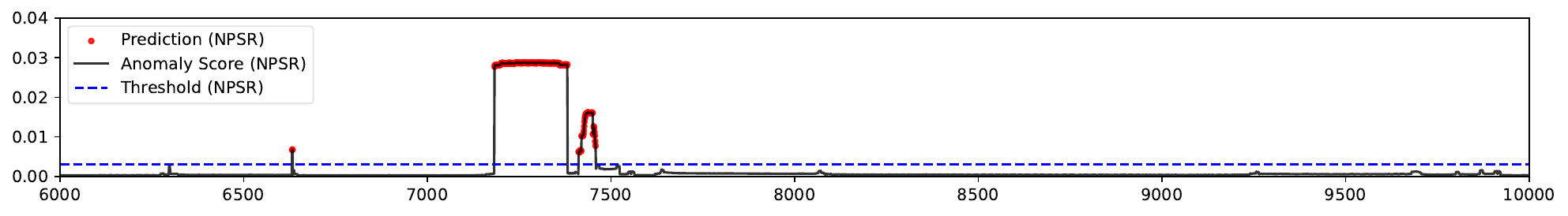}}
    \subfloat[(NPSR) Case 4]{\includegraphics[width=0.49\linewidth]{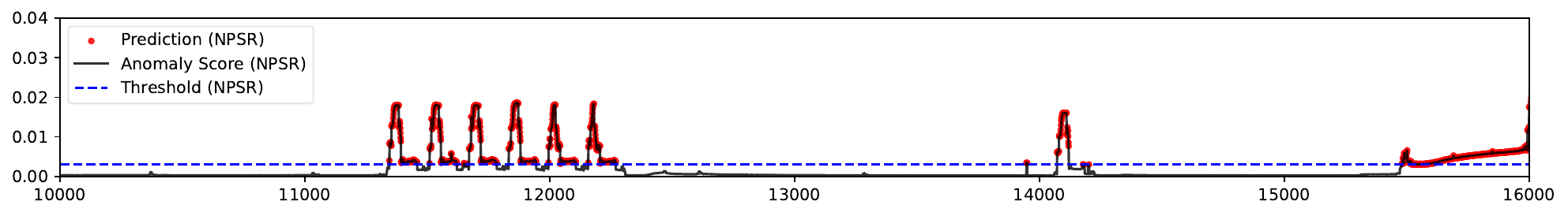}}
    \caption{Other algorithms' detection performances in real-world.}
    \label{fig:case_anom_trans}
\end{figure*}

\begin{figure*}
    \centering
    \subfloat[Similarity score in first layer]{\includegraphics[width=0.49\linewidth]{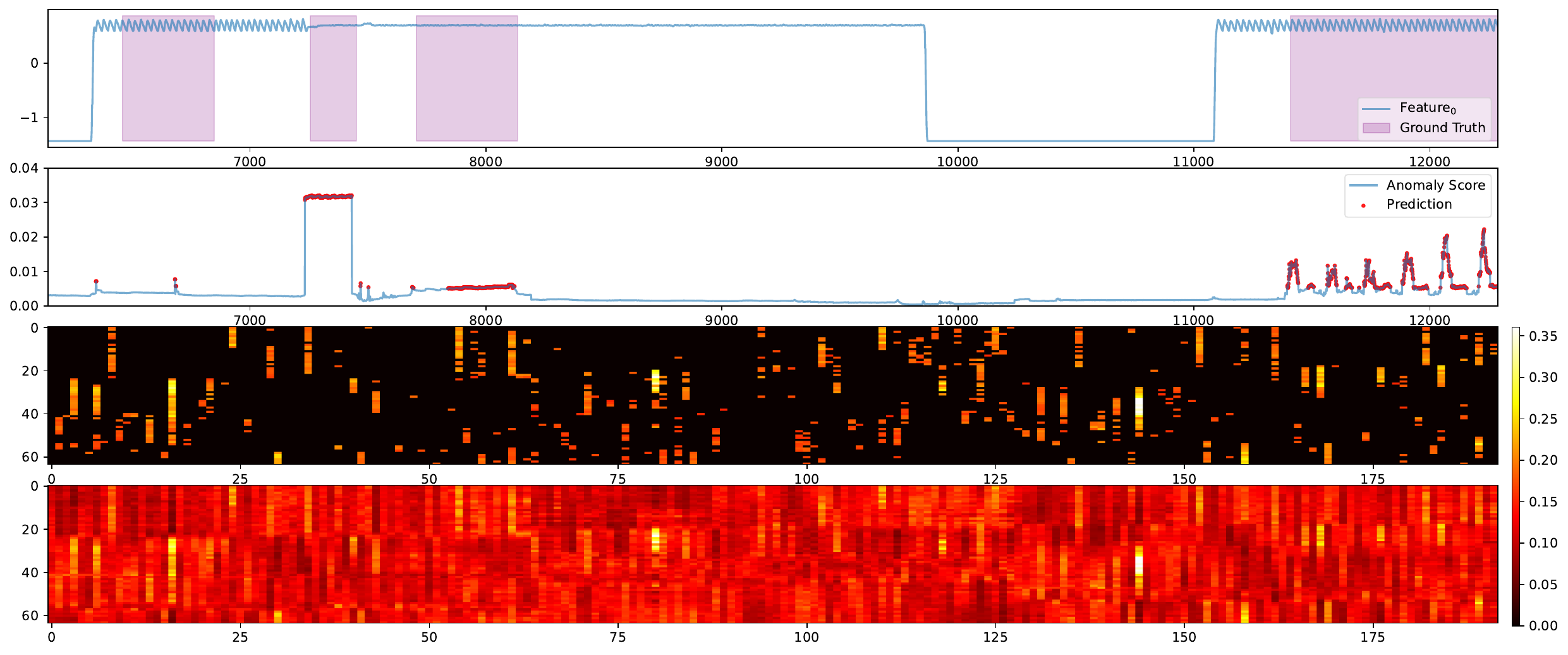}}
    % \subfloat[Attention map in second layer]{\includegraphics[width=.49\linewidth]{heat_plot/6144_12288heat_layer1.pdf}}\\
    \subfloat[Similarity score in third layer]{\includegraphics[width=0.49\linewidth]{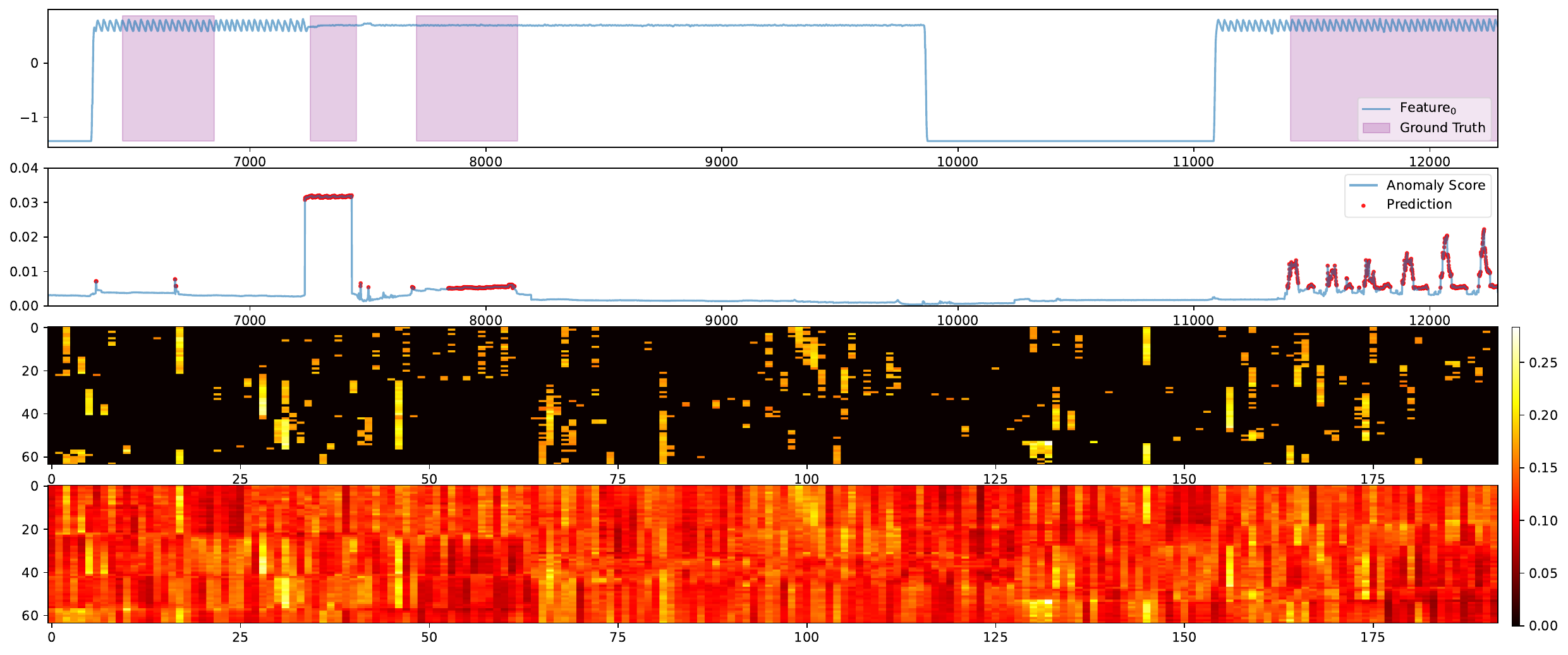}}\\
    % \subfloat[Attention map in fourth layer]{\includegraphics[width=\linewidth]{heat_plot/6144_12288heat_layer3.pdf}}\\
    \subfloat[Similarity score in fifth layer]{\includegraphics[width=0.49\linewidth]{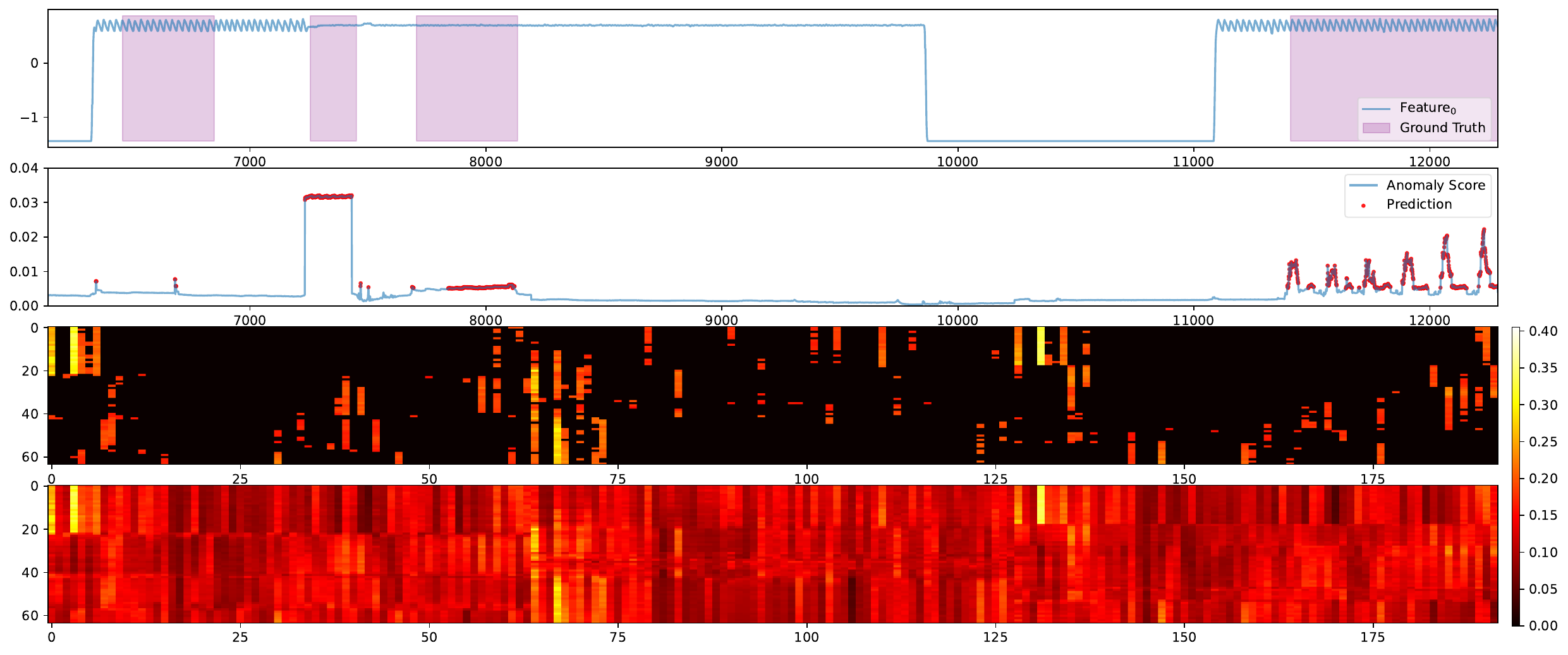}}
    % \subfloat[Attention map in sixth layer]{\includegraphics[width=\linewidth]{heat_plot/6144_12288heat_layer5.pdf}}\\
    \subfloat[Similarity score in seventh layer]{\includegraphics[width=0.49\linewidth]{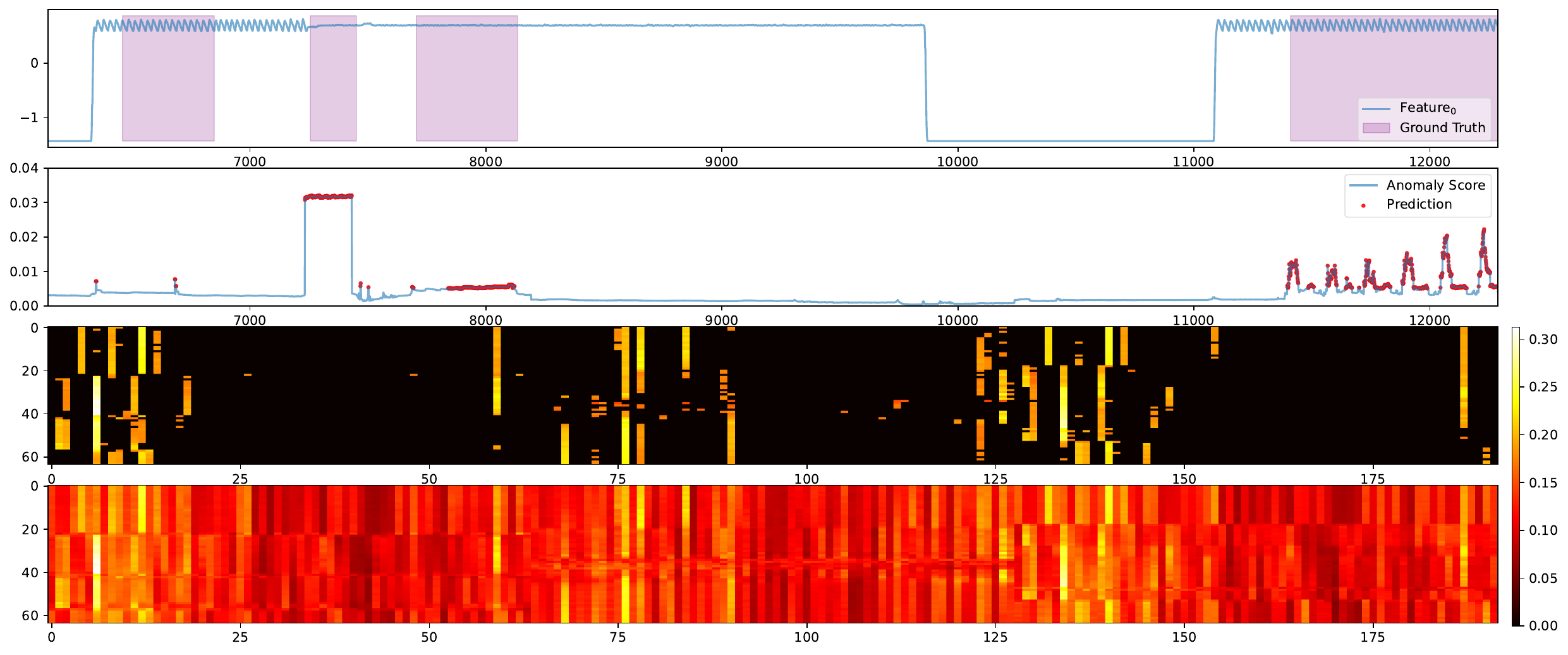}}
    \caption{Similarity scores in different layers.}
    \label{fig:heat_map_all}
\end{figure*}

% 图\ref{fig:case_anom_trans}展示了Anomaly Transformer的真实检测效果，可以发现，它只能检测出少数异常，在Case 1中有两个异常没有检测出来，并且无法准确检测段异常。并且异常分数区分度过小，在Case 2中只能勉强检测出异常。
Fig. \ref{fig:case_anom_trans} illustrates the actual detection performance of the AnomTrans and NPSR on the SWaT dataset. AnomTrans can only detect a few anomalies. In Case 1, two anomalies were not detected and AnomTrans struggled with accurately detecting segment anomalies. Moreover, the anomaly score differentiation is relatively low, as in Case 2, where it has limited success detecting anomalies.
% 在案例3和4显示，NPSR可以检测出比AnomTrans更多的异常（尤其是段异常），但是他依然有一部分异常无法检测出来（如6500和8000时间点附近的这两段异常），并且他的误报更多一些，如时间点140000附近存在一些负正类。
In cases 3 and 4, NPSR is more effective in detecting anomalies than AnomTrans, especially segment anomalies. However, NPSR fails to detect certain anomalies, such as near time points 6500 and 8000. Additionally, NPSR tends to produce more false positives, as observed around time point 140000, where a few instances are falsely classified as positive.
Furthermore, Fig. \ref{fig:heat_map_all} illustrates the similarity scores at various model layers. In the initial layer, the similarity scores exhibit a relatively dispersed pattern. However, as the number of layers increases, the similarity scores gradually rise, leading to more focused attention.

\subsection{Analysis of cosine similarity loss
}\label{sec:cos_sim_loss}

1. \textbf{Direct Explanation}: MSE enforces point-wise reconstruction by requiring the model to return each time point in the sequence to its original value, acting as a point-wise loss. Cosine similarity, on the other hand, enhances the alignment between patches, promoting smoother transitions and reducing noise interference during reconstruction.
2. \textbf{Semantic Explanation}: Patch-based time series studies suggest that patches are more effective at capturing the semantic information of sequences. While interpolative learning can reconstruct individual time points, patches benefit from more than just simple interpolation. Traditional patch learning often relies solely on MSE loss, ignoring this benefit.
3. \textbf{Visual Explanation}: As shown in Fig. \ref{fig:case_study2}, incorporating cosine loss enables the model to better detect longer anomalies. For additional details, see Section \ref{sec:vis}.
4. \textbf{Numerical Explanation}: Ablation experiments were conducted by removing cosine terms from Equations 4, 5, and 7 while keeping the MSE loss. The results, presented in Table \ref{tab:abl_full}, indicate a significant decrease in model performance after removing these terms.

\subsection{Detailed quantified comparison}\label{sec:quan_ana}

For a detailed comparison, we group Deep SVDD, Deep IF, USAD, TCN-ED, NPSR, and TimesNet, as they can be considered traditional deep learning methods. The remaining methods are grouped as in Section \ref{sec:baselines}. Below is the analysis and discussion.

1. In the \textbf{traditional deep learning group}, NPSR has achieved the highest average ranking. Our proposed method significantly outperforms NPSR on datasets SWaT, WADI, and Swan. All of these dataset are more complex and higher-dimensional. The superiority of our proposed method over NPSR can be attributed to that the former can handle high-dimensional data better and have a larger parameter capacity.

2. Among \textbf{Transformer-based models}, TranAD has obtained the highest average ranking. With more parameters, TranAD performs better on datasets SWaT, PSM, and Swan. However, since these models rely on point-wise modeling of time series data, their complexity is high, limiting their performance. In contrast, SimAD employs a patch mechanism, allowing the model to support window sizes of 1024 or higher, resulting in a larger receptive field and the ability to capture more information.
\textbf{SimAD's patch mechanism enables larger window sizes and better information capture compared to point-wise Transformer models like TranAD.}

3. \textbf{Contrastive learning-based} methods perform similarly on average. Taking COUTA as an example, it utilizes multiple methods to generate negative samples and leverages contrastive learning to enhance model robustness. However, its reliance on a TCN network structure and complex sample generation strategy necessitates high-quality datasets. This results in superior performance on MSL and SMAP but causes sharp declines on other datasets. In contrast, our method employs a simpler negative sample generation strategy and a more complex network structure, making it less dependent on dataset quality.
\textbf{Our method's simpler negative sample generation and more complex network structure make it more adaptable to various dataset qualities compared to contrastive learning methods like COUTA.}

4. D3R is the only \textbf{diffusion model}. It preserves stable components in time series data, such as trends and periodic information, which enhances anomaly detection, particularly for unstable time series data with sudden increases. However, its performance is limited by stronger prior assumptions and data quality requirements, making it less effective across all datasets. In contrast, SimAD operates on more general assumptions, recognizing that normal and anomalous data have different representations. This universal modeling approach allows SimAD to adapt more effectively to a wide range of datasets.
\textbf{SimAD's general assumptions about data representation enable it to adapt more effectively across various datasets compared to the diffusion model D3R.}

5. GPT2-Adapter is an \textbf{LLM-enhanced} time series anomaly detection model that integrates large language models with time series data. Although it represents a pioneering effort in applying LLMs to this field, its performance in anomaly detection falls short. This is mainly due to challenges with aligning natural language models with time series data, difficulties in fine-tuning LLM parameters, and the lack of specialized optimization for time series anomaly detection. Consequently, GPT2-Adapter often underperforms in practical scenarios.
\textbf{Despite being a pioneer in LLM-based approaches, GPT2-Adapter falls short due to modal alignment issues and insufficient optimization for time series anomaly detection.}

\section{Time costs}\label{sec:time}
% 我们将SimAD与其他模型在模型大小，训练与测试时间进行比较对比，均使用SWaT数据集。为了统一设置，训练时间估计方法为设置所有方法学习256\times 500个数据样本，即128000个样本。其结果如表\ref{tab:time_cost}所示。
% 注意，由于D3R的原始代码可能存在一些问题，这导致它的测试时间(2794.25s)过长，我们估计其实际测试时间为15.98s。

Table \ref{tab:time_cost} presents a comparative analysis of SimAD and other models regarding model size, training time, and testing time using the SWaT dataset. The estimated training time for all methods involved training on 128,000 samples to ensure a fair comparison.
However, an issue with the original code of the D3R model caused its reported testing time to be excessively long (2794.25s). To provide a more accurate estimation, we estimated the actual testing time of D3R to be 15.98s.

In terms of model size, SimAD demonstrates a significant advantage over D3R and GP2-Adapter (LLM model). Its smaller model size indicates efficient memory usage.
SimAD also performs well regarding inference time, showing shorter inference times than other models. This implies that SimAD can make predictions faster during deployment, which can be crucial in real-time applications.

However, it's important to note that SimAD has a longer training time and a larger model size than the other models in the analysis. Despite this, considering that SimAD achieves performance improvements of over 60\%, the model size and inference speed of SimAD are still within an acceptable range.
To further improve SimAD, future work will explore model optimization techniques to reduce the model size while maintaining or enhancing its performance. This would address the longer training time and larger model size, making SimAD even more efficient and practical for deployment.

% 我们的模型大小远小于D3R与GP2-Adpater(LLM模型)，且推理时间更短。但是与其他模型相比，训练时间较长，模型偏大。考虑到SimAD的性能提升在60\%以上，因此模型大小与推理速度仍在接受的范围内。未来我们希望探索模型优化，进一步减小模型大小。

\begin{table}[htbp]
  \centering
  \caption{The time costs of different algorithms. The lower is better.}
    \begin{tabular}{c|c|c|c}
    \toprule
    Methods & Model Size (MB) & Train (sec) & Test (sec) \\
    \midrule
    USAD & 332.07  & 72.14  & 1.51  \\
    TCN-ED & 0.04  & 41.34  & 0.86  \\
    NCAD & 0.19  & 141.73  & 1.77  \\
    AnomTrans & 28.29  & 33.25  & 3.91  \\
    TimesNet & 19.15  & 24.29  & 3.62  \\
    D3R & 225.10  & 1900.37  & 15.98 \\
    GPT2-Adapter & 241.58  & 241.58  & 71.24  \\
    Ours & 114.75  & 245.88  & 1.27  \\
    \bottomrule
    \end{tabular}%
  \label{tab:time_cost}%
\end{table}%

\section{Boader impacts}\label{sec:imp}
% 时序异常检测旨在检测不正常行为与事件。它在风险防范，健康监测，工业检测等领域均有应用。因此其覆盖产业较广，需求较多。实际场景下往往缺少异常事件的标签，而我们的方法可以适应这种情况，并且提供高精准度的检测结果。但是部分场景需要能够在计算资源紧张的条件也良好工作，我们的模型还需要采用其他技术减小模型大小适应这种情况。在未来，我们希望考虑模型轻量化的同时保证检测精度以解决这一缺点，创造更大的社会价值。
Time series anomaly detection is crucial in identifying abnormal behaviours and events. Its applications span across various domains, including risk prevention, health monitoring, and industrial inspection. The demand for effective anomaly detection techniques is high due to its wide industrial coverage. In practical scenarios, labeled data for anomalous events is often scarce. However, our method is designed to adapt to this situation and provide highly accurate detection results. There are scenarios where the model needs to perform well under resource-constrained conditions. Therefore, it becomes necessary for our model to incorporate other techniques that can reduce its size and enable it to adapt to such situations. In the future, we aspire to address this limitation by considering the model lightweight while ensuring detection accuracy, thereby creating greater societal value.

\section{Limitations}\label{sec:lim}
% 由于我们的模型采用了类似Transformer的架构作为主干。这导致模型的计算消耗较大。在未来，我们会考虑寻找其他更加轻量化且高效的网络框架替换，例如MAMBA，卷积网络。同时，考虑到目前时序异常检测的算法仍然存在缺点，如评估不准，计算耗时等，未来我们会进一步提出新的评估准确且高效的异常检测指标。
Significant computational costs are incurred due to the adoption of a Transformer-like architecture as the backbone of the model. In the future, we will explore alternative network frameworks, such as MAMBA or convolutional networks, that are more lightweight and efficient. Furthermore, considering the existing limitations of current time series anomaly detection algorithms, such as inaccurate evaluation and lengthy computation, we aim to propose novel evaluation metrics that are both accurate and efficient for anomaly detection.
% 以下是我们对模型更深层缺陷的分析：

Below is the in-depth analysis of the intrinsic shortcomings of the model.

\textbf{TSAD in Practical Scenarios}:
TSAD often requires fast and effective models. Our Transformer-based model's computational cost is a limitation. Assuming batch size $B$, time series length $T$, channels $C$, and patch length $P$, the number of patches is $M = \frac{T}{P}$, and the final input is $(B, M, P \times C)$. With hidden dimension $D$, the intermediate feature shape is $(B, M, D)$. At this point, the computational bottleneck lies in the Transformer's self-attention and \rv{feed-forward network (FFN)} \cite{attention_is}. The complexity of self-attention is quadratic concerning the sequence length, with a computational complexity of $(B \cdot M^2 \cdot D)$, while the FFN complexity is $(B \cdot M \cdot D^2)$. The overall computational complexity of the model is $B \cdot (M^2 \cdot D + M \cdot D^2)$. When the number of patches is greater than the model dimension, this complexity can be approximated as $(B \cdot M^2 \cdot D)$. 
\rv{Note that SimAD's ContrastFusion is a two-layer FFN projection head, whose complexity equals that of the FFN. During reconstruction, an extra linear layer with lower complexity than the FFN is used, and their complexities differ by a constant factor. Thus, SimAD's overall complexity is dominated by the EmbedPatch Encoder's complexity, with other components contributing less.}
% 注意，SimAD的ContrastFusion是一个两层FFN的投影头，它的复杂度即为FFN的复杂度，在重构时，我们额外使用一个线性层，它的复杂度小于FFN的复杂度，且存在一个常数量级关系。因此，最终SimAD的复杂度由EmbedPatch Encoder的复杂度主导，其他组件的复杂度低于主复杂度.
At this point, the length of the time series affects the computational efficiency, leading to slower model performance.

To address the computational cost, we can replace the attention backbone in SimAD with frameworks like SSM or MAMBA, whose complexity is linear concerning the sequence length, resulting in $(B \cdot M \cdot D^2)$ overall, with FFN as the bottleneck. However, when $N$ is less than the model dimension, linear complexity models do not resolve the bottleneck. In this case, we can explore other FFN structures, such as using grouped Linear layers, dividing the input into $G$ groups of size $\frac{D}{G}$. This reduces the FFN complexity to $(B \cdot M \cdot \frac{D^2}{G})$. 

\textbf{Comparison with Non-Patch-Based Transformer Models}: 
For non-patch-based Transformer models like AnomTrans, the computational bottleneck lies in the self-attention calculation, leading to a complexity of $(B \cdot T^2 \cdot D)$. Since we slice the sequential data, the actual sequence length in our network is shorter. Therefore, our computational complexity is more efficient.

\textbf{Comparison with Channel-Independent and Patch-Based Models:} 
For channel-independent, patch-based models like DCdetector, the complexity is $(B \cdot C \cdot M^2 \cdot D)$, where $B$ is the batch size, $C$ is the number of channels, $M$ is the length of the patch sequence, and $D$ is the dimensionality of features. These models must process each channel independently, resulting in higher computational complexity. Consequently, our approach offers a more efficient alternative in this respect.



\section{Insights behind SimAD}\label{sec:insights}
1. \textbf{Our model emphasizes simplicity \& effectiveness, avoiding excessive complexity.} Our goal is to establish a new paradigm for time series anomaly detection (TSAD), akin to how SimCLR sets the foundation for image representation. We consider that a simpler, more effective model allows for further exploration and development, leaving ample room for future improvements to SimAD. For example, SimAD employs the simplest noise augmentation for generating negative samples and utilizes a straightforward MSE loss for constructing an asymmetric optimization strategy, yet it still delivers excellent results. We expect that SimAD's focus on simplicity and core principles will make it a cornerstone of future TSAD research, offering new perspectives to the community.

2. \textbf{Our proposed UAff and NAff are simple \& effective metrics. Many related works have highlighted the shortcomings of existing TSAD metrics.} While the Affiliation metric provides a solid foundation, it has notable deficiencies: it fails to accurately evaluate random algorithms and weaker methods, and its discrimination is limited, overlooking performance differences between models. We believe that simplicity and effectiveness are central to our approach, as reflected in our improvements to the metrics.

3. \textbf{We fully leverage the concept of viewing similarity from different perspectives, operating under the belief that normal and abnormal time series representations are inherently dissimilar.}
\rv{Although SimAD may superficially resemble BYOL \cite{byol} in terms of its loss function design, the underlying motivations differ significantly. BYOL is based on the premise that representations of different views of the same data should be similar in the shared latent space. This is achieved in BYOL by employing contrastive learning to bring representations of different views closer together. Besides, BYOL updates its parameters via exponential moving averages. In contrast, SimAD employs a strategy that emphasizes maximizing the dissimilarity between representations of normal data and those of abnormal data. Furthermore, our model updates its parameters through an asymmetric loss optimization objective function, guided by a stop-gradient mechanism. 
SimSiam \cite{SimSiam} is a self-supervised learning framework that enhances image generalization by learning view-invariant representations, using noise as a form of real-world augmentation. It utilizes two networks without parameter sharing and employs a cosine similarity loss to align projected and pre-projected features, effectively preventing collapse to trivial solutions. In contrast, our proposed method, SimAD, is designed to distinguish between normal and anomalous data by encouraging the projected features of two views to diverge in a low-dimensional space. This key difference in both implementation and objective distinguishes SimAD from SimSiam.}
% 并且在实现上，我们与BYOL也完全不同，BYOL采用指数滑动平均的在线更新的方式更新参数。在过去同样类似的图像自监督经典工作是SimSiam，我们同样受到了这一篇论文的启发，但是与前面所述一致。SimAD与它的实现目标或关键思想是不同的。SimSiam同样是通过学习视图的相似性以提升对图像的泛化识别能力。因为在自监督过程中引入的噪声可以视作真实世界的噪声，因此这种噪声学习可以增强模型的泛化能力。在实现上，两者同样存在一些区别，SimSiam采用两个阮生的不共享参数的网络学习知识，并且采用的基本损失是余弦距离；同时，更大的区别是，SimSiam的最终目标是让投影后的特征与投影之前的特征靠近，以此获得非平凡解。而我们是让两种视图的投影后特征彼此远离，引导模型在低维空间中区分正常与异常。

4. \textbf{The motivation of negative samples}: Previous research (Sec. 4.5.1 in \cite{amsl}) has shown that adding noise and scaling to time series data can alter the underlying information within the series. Study \cite{couta} employed a single time point alteration strategy to generate synthetic anomalies (contaminated data), while \cite{COCA} used contrastive learning-based single-class algorithms. Both approaches indicate that adding noise can create varying perspectives, thereby helping to clarify the model's decision boundaries.

The distinction between normal and anomalous instances is influenced by previous research, which indicates that adding noise can distort the original information in time series, potentially generating synthetic anomalies. This is intuitive, as increased noise leads to greater deviations from the original patterns. While TSAD currently does not use labels during training, research has shown that contrastive learning with negative samples can enhance a model's decision boundaries. As the gap between original time series data and augmented data widens, the model becomes better at capturing consistent information, such as trends and periodic patterns.

Our experiments and visualizations suggest that EmbedPatch primarily learns from normal data, resulting in less resemblance to anomalous time series. It is acknowledged that simple noise may not capture all the complexities of real-world anomalies. Future research aims to develop strategies that better address various real-world anomalies, as exemplified in \cite{COCA}. Nonetheless, our paper presents a more foundational approach using simpler strategies, offering a fresh perspective for the TSAD community.

In time series anomaly detection, \cite{cont1} introduces noise into the embedding and tasks the model with denoising to reconstruct the original time series. Similarly, in image anomaly detection, \cite{uni_ad} demonstrates the effectiveness of ``adding perturbations to feature tokens, guiding the model to learn knowledge of normal samples through the denoising task." Building on this prior work, we assert that the primary function of denoising is to guide the model in learning the patterns of normal data, implicitly directing its focus towards normal behavior. Our ablation experiments further validate this point, showing that the model's performance deteriorates when denoising is not applied.

5. \textbf{The concept of ``Dissimilarity" refers to the dissimilarity between normal and abnormal samples or representations}. Specifically, this concept manifests in the following aspects:

\begin{itemize}
    \item \textbf{Dissimilarity between positive and negative samples}: We expand the representation space between positive (original time series) and negative (constructed using noise) samples, thereby reducing their similarity and increasing their dissimilarity.
    \item \textbf{Dissimilarity between normal and abnormal samples}: By diminishing the similarity between positive and negative samples, our objective is to heighten the differentiation between normal and abnormal samples, thus enabling the model to enhance anomaly detection efficacy. The distinction between normal and abnormal sample representations forms the core motivation behind SimAD. 
    \rv{As no labels are accessible during training, it is hard to determine whether a sample is anomalous.}
    Nonetheless, prior research as well as our own work indicate that even in the absence of labels, SimAD can effectively model the dissimilarity between normal and abnormal instances through learning the dissmilarity of positive and negative samples.
    \item \textbf{Dissimilarity embedded in EmbedPatch}: EmbedPatch, a crucial component in our model, introduces learnable parameters independent from input data. These parameters enable the model to capture the most prevalent features in data and increase the difficulty of reconstruction, compelling the model to learn the universal characteristics of time series. A higher-layer EmbedPatch acquires more abstract normal semantic information, while a lower-layer EmbedPatch extracts simpler patterns from the time series. 
    \item \textbf{Contrastive learning design utilizing dissimilarity}: We consider normal and abnormal time series as having dissimilar view representations. Equation \ref{eq:L_cont} uses MSE and cosine similarity to separate the representation spaces of normal and abnormal data, thereby enlarging their dissimilarity.
    % 尽管我们基于不相似的方法在一定程度是与对比学习的思想类似，但是从原理与实现目标上均存在不同。从原理上，对比学习往往认为增强后的样本与原样本相似，而其他不同样本作为负样本；然而根据过往研究\cite{COCA}的结论，在时序中将增强样本视作负样本，即与原样本是不相似的，这样更符合时序数据的特征。因此从原理是，我们基于不相似的方法与对比学习的方法存在差异。从实现上，我们基于不相似的方法更加强调patch之间的不相似性以及正常与异常之间的不相似性，并利用余弦相似性刻画这种关系。这在实现上也与对比学习方法存在不同，因此我们的方法被称作“基于不相似”路线更加合理。
    While our approach based on dissimilarity shares some similarities with the concept of contrastive learning to a certain extent, there are fundamental differences in principles and implementation goals. From a theoretical perspective, contrastive learning often assumes that the augmented samples are similar to the original samples, with other dissimilar samples serving as negative samples. However, based on the findings of previous studies \cite{COCA}, considering augmented samples as negative samples in time series, i.e., dissimilar to the original samples, aligns better with the characteristics of time series data. Therefore, in principle, our dissimilarity-based approach differs from contrastive learning methods. In terms of implementation, our dissimilarity-based approach places greater emphasis on the dissimilarity between patches and the dissimilarity between normal and abnormal instances, utilizing cosine similarity to characterize this relationship. This implementation aspect also sets our approach apart from contrastive learning methods, hence the designation of our method as the ``dissimilarity-based" approach seems more appropriate.
\end{itemize}

\section{Discussion}
\rv{\textbf{How does patch-based feature extraction work?} 
Patch-based feature extraction methods are not new in the context of time series, and they have been effectively verified in time series prediction and anomaly detection in the past. For example, in PatchTST \cite{patchtst}, the authors use the patch approach to segment the time series, and then use an attention model to learn the time series features and predict future time series data. In DCdetector \cite{dcdetector}, the authors also adopt the patch-based method to model the local and global time series information. In this way, the model is guided to capture the differences between normal and abnormal time series in both global and local aspects, and finally, these features are utilized to complete the anomaly detection. However, in previous papers, there is a lack of in-depth exploration of the patch-based approach. Mainly, these methods are limited by the size of the time window. Usually, the time window size they set by default is 105, and the size of the patch is often 3, with a maximum of 7. Under such settings, the intuitive understanding is that they use three time points to represent the features of a very small interval of the time series. In fact, this method does not have a significant difference from the past modeling methods that are completely based on the time window. Because ultimately, the model is still limited by the time window of 105, which makes the model only learn the time series features in a local area. But in the real world, time series data often contains at least tens of thousands or even hundreds of thousands of timestamps. Therefore, a time window of 105 makes it difficult for the model to learn longer context relationships. To address this issue, for the first time, we scale the time window to 2048, which is 20 times larger compared to the previous models. However, this expansion comes at a cost. Since a longer time window requires the model to process more data at once, it will lead to extremely slow model inference speed, which is at least one-twentieth of the previous speed. And in the actual inference scenario, this is completely unusable. To solve this problem, we make improvements in the way of time series feature extraction, that is, we change the fixed mindset of using the original patch. We also expand the size of the patch to at least 32 lengths. That is to say, we use the data of 32 time points for representation. Finally, the model only needs to process 2048/32 basic units, so the features actually processed by the model are reduced. At the same time, due to the existence of a lot of redundant data in the time series itself, this processing method does not limit the model's learning of time series data. For example, in some time series, the channel values of the time series may remain around a certain level for a long time, with only slight fluctuations. In this case, the truly effective information is to mine the stable values of the time series during this period. Or, in the time series, periodic information is often very common, and there may be slight changes within each period. Similarly, most of the time series data is redundant at this time. Therefore, the patch-based processing and expanding the size of the patch to 32 will not affect the model's learning of time series data. Meanwhile, our experiments also show that when the time window is larger and the patch is larger, the loss of the model is actually smaller, which means that the model fits the time series more perfectly.}

\bibliographystyle{IEEEtranN}
\bibliography{main.bib}

\begin{IEEEbiography}[{\includegraphics[width=1in,height=1.25in,clip,keepaspectratio]{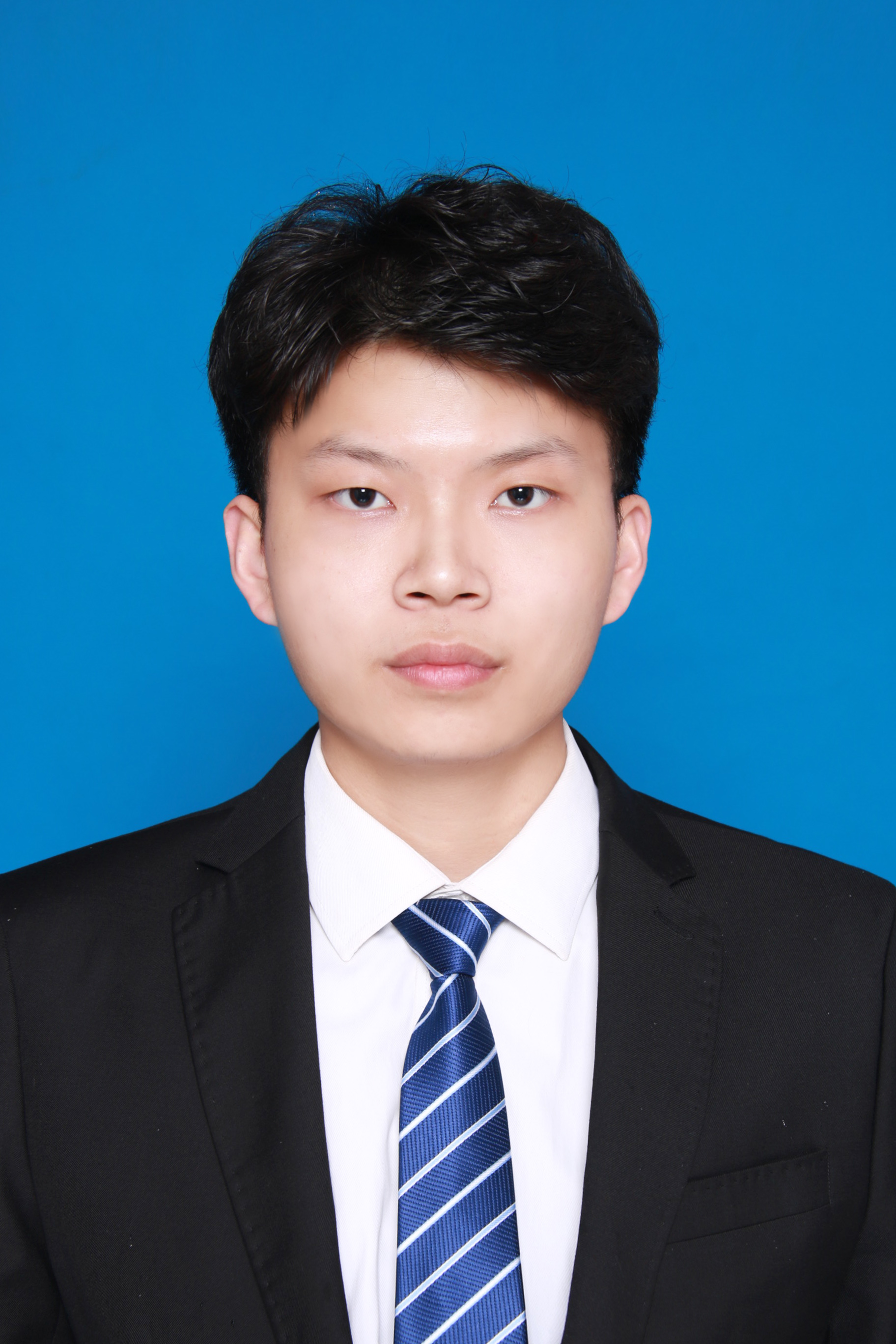}}]{Zhijie Zhong} received the B.S. degree in 2022 from the Harbin Engineering University, Harbin, China and he is currently working toward the Ph.D. degree with the South China University of Technology, Guangzhou, China. His research interests include data mining, machine learning, time series analysis, anomaly detection, and large language model (LLM).
\end{IEEEbiography}

\begin{IEEEbiography} [{\includegraphics[width=1in,height=1.25in,clip,keepaspectratio]{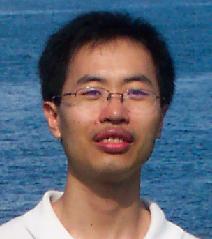}}]{Zhiwen Yu (S'06-M'08-SM'14)} is a Professor in School of Computer Science and Engineering, South China University of Technology, China. He received the Ph.D. degree from the City University of Hong Kong, Hong Kong, in 2008. Dr. Yu has authored or coauthored more than 200 refereed journal articles and international conference papers, including more than 70 articles in the journals of IEEE Transactions. His google citation is more than 10000, and h-index is 44. He is an Associate Editor of the IEEE Transactions on systems, man, and cybernetics: systems. He is a senior member of IEEE and ACM, a Member of the Council of China Computer Federation (CCF).
\end{IEEEbiography}

\begin{IEEEbiography} [{\includegraphics[width=1in,height=1.25in,clip,keepaspectratio]{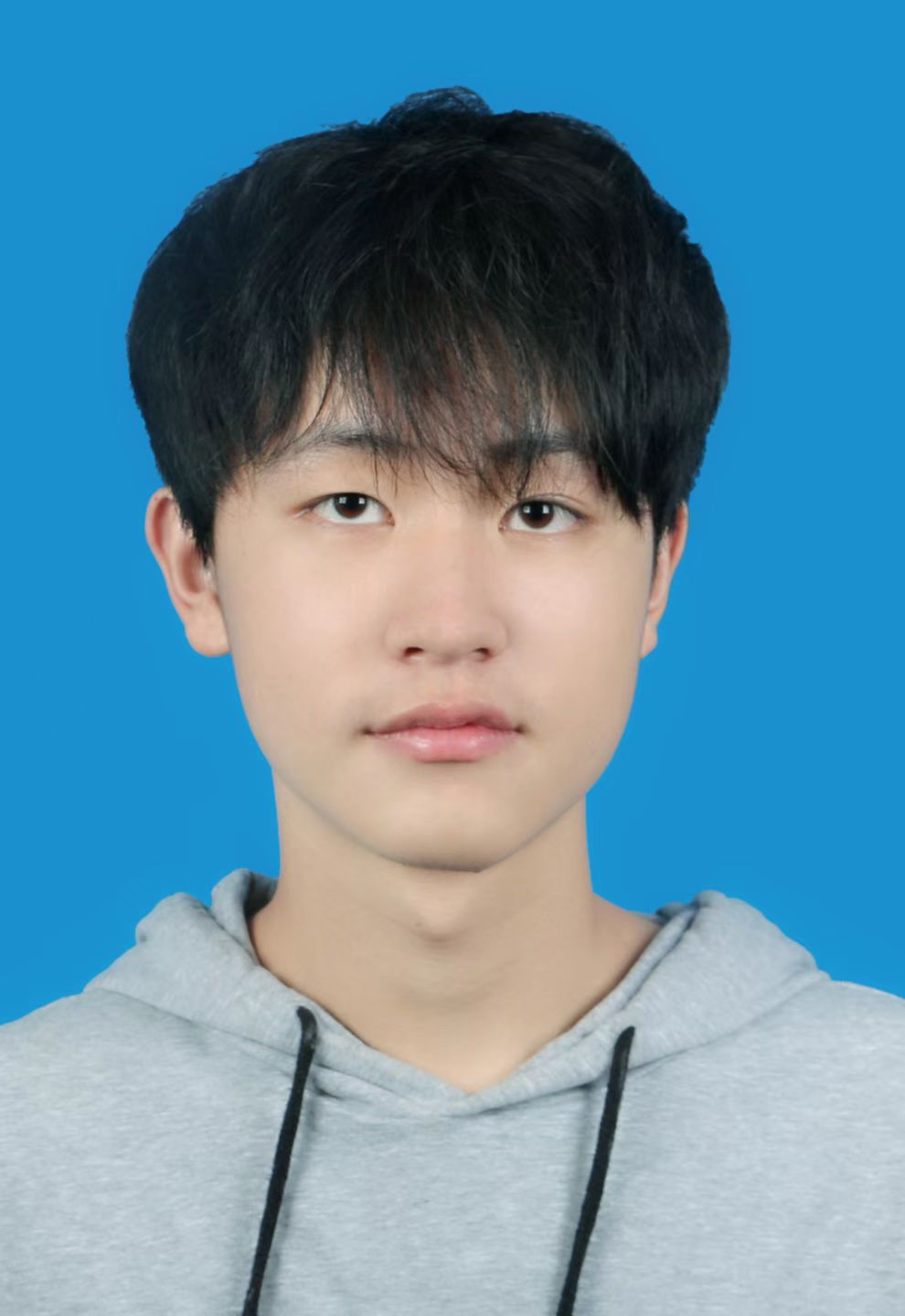}}]
{Xing Xi} received the B.S. degree in 2020 from the Guangdong Baiyun University and the M.A. degree from the Guangdong University of Technology, Guangzhou, China. Current, he is currently working toward the D.E degree with the South China University of Technology. His research interests include object detection, open world and vocabulary object detection.
\end{IEEEbiography}

\begin{IEEEbiography} [{\includegraphics[width=1in,height=1.25in,clip,keepaspectratio]{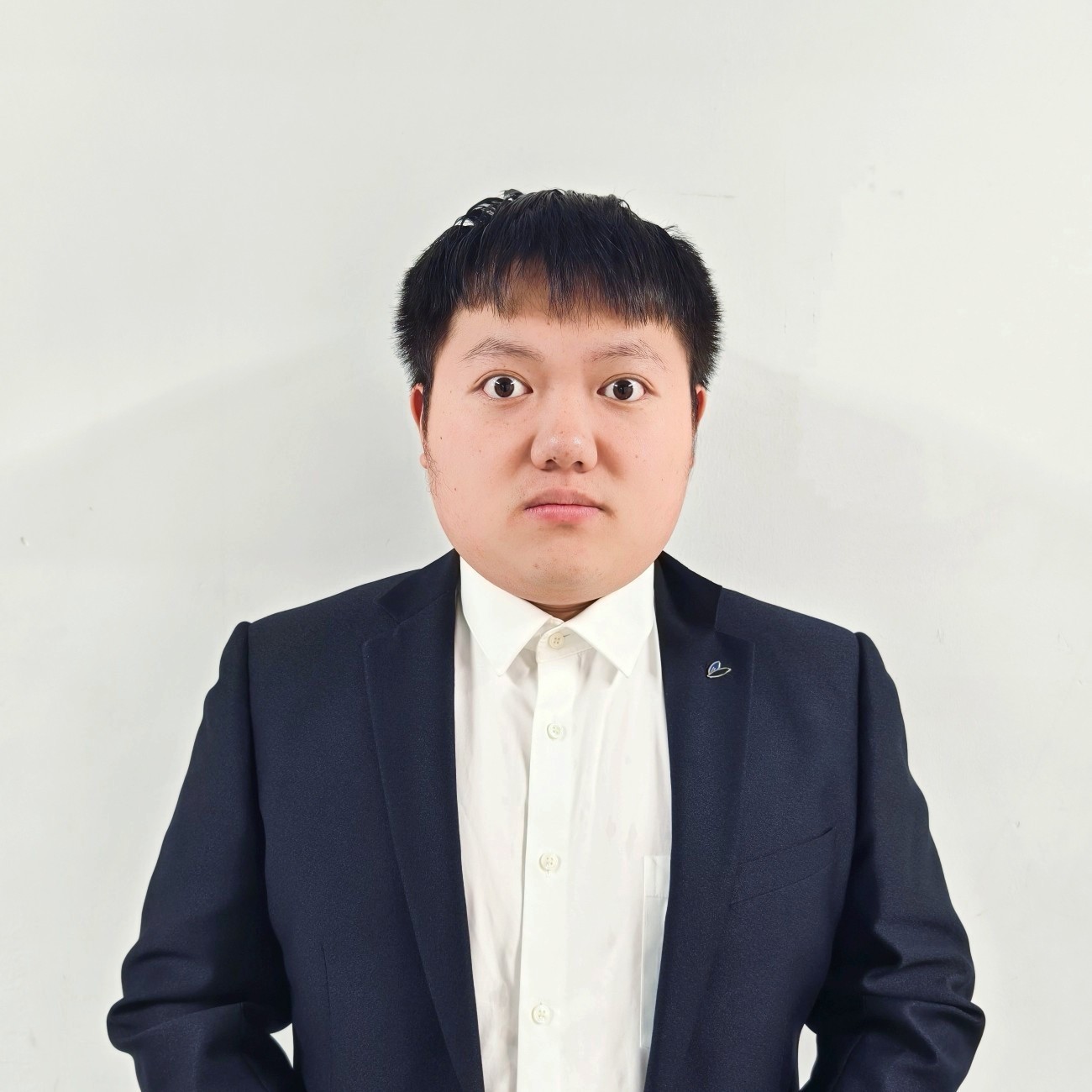}}]
{Yue Xu} received the B.S. degree in 2025 from the South China University of Technology, Guangzhou,China and he is currently working toward the M.A. degree with the South China University of Technology, Guangzhou, China. His research interests include adversarial attack, continual learning and anomaly detection.
\end{IEEEbiography}

\begin{IEEEbiography}[{\includegraphics[width=1in,height=1.25in,clip,keepaspectratio]{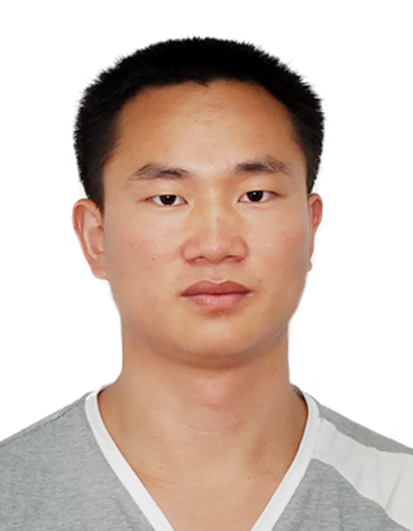}}]{Wenming Cao (S'16)}
received M.Sc. degree from the School of Automation, Huazhong University of Science and Technology (HUST), Wuhan, China, 2015, and Ph.D. degree at the Department of Computer Science, City University of Hong Kong, 2019. He has been a Postdoctoral Fellow at the University of Hong Kong from 2019 to 2021. He is an associate professor with the Department of Information and Computing Science, Chongqing Jiaotong University. 
His research interests include data mining and machine learning.
\end{IEEEbiography}

\begin{IEEEbiography}[{\includegraphics[width=1in,height=1.25in,clip,keepaspectratio]{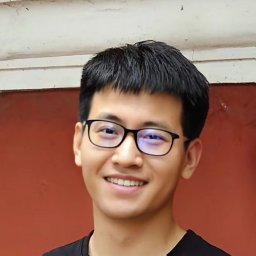}}]{Yiyuan Yang} is a DPhil student at the Department of Computer Science, University of Oxford, United Kingdom. He focuses on the field of intelligent sensing systems, time series, spatiotemporal data mining, anomaly detection, and generative models. He previously studied at the Department of Automation at Tsinghua University and interned at the Alibaba DAMO Academy and Huawei Noah's Ark Lab.
\end{IEEEbiography}

\begin{IEEEbiography} [{\includegraphics[width=1in,height=1.25in,clip,keepaspectratio]{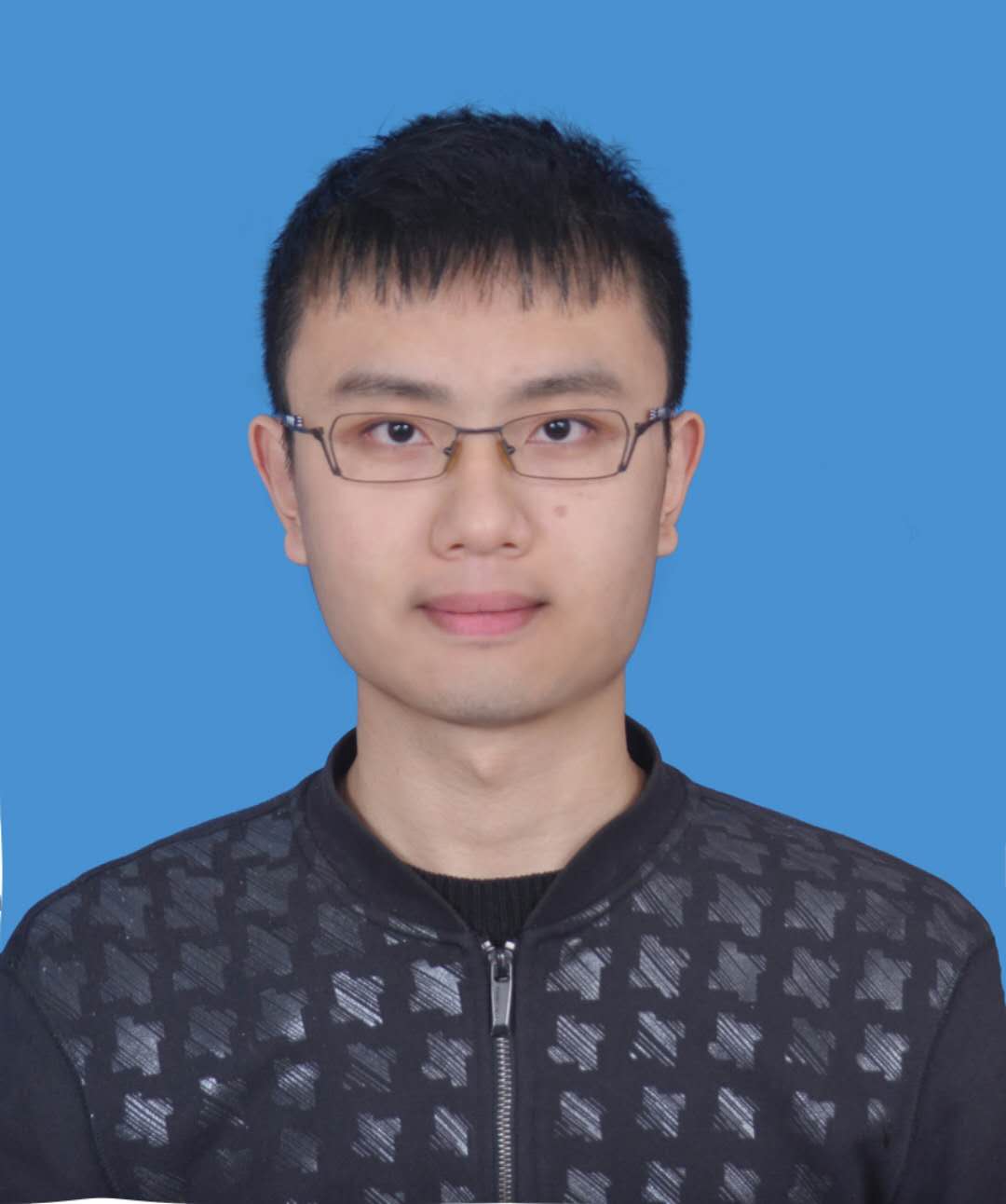}}] {Kaixiang Yang (M'21)} received the B.S. degree and M.S. degree from the University of Electronic Science and Technology of China and Harbin Institute of Technology, China, in 2012 and 2015, respectively, and the Ph.D. degree from the School of Computer Science and Engineering, South China University of Technology, China, in 2020.

He has been a Research Engineer with the 7th Research Institute, China Electronics Technology Group Corporation, Guangzhou, China, from 2015 to 2017, and has been a Postdoctoral Researcher with Zhejiang University from 2020 to 2021. He is now with the School of Computer Science and Engineering, South China University of Technology. His research interests include pattern recognition, machine learning, and industrial data intelligence.
\end{IEEEbiography}

\begin{IEEEbiography} [{\includegraphics[width=1in,height=1.25in,clip,keepaspectratio]{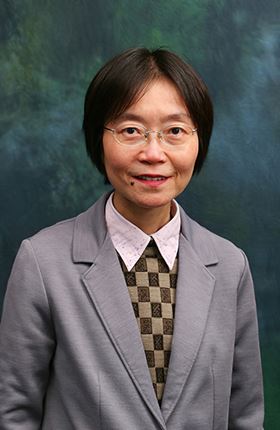}}] {Jane You} received the BEng degree in electronics engineering from Xi’an Jiaotong University, Xi’an, China, in 1986, and the PhD degree in computer science from La Trobe University, Melbourne, VIC, Australia, in 1992. She was a Lecturer with the University of South Australia, Adelaide SA, Australia, and a Senior Lecturer
with Griffith University, Nathan, QLD, Australia,
from 1993 to 2002. She is currently a Full Professor with The Hong Kong Polytechnic University, Hong Kong. Her current research interests
include image processing, pattern recognition, medical imaging, biometrics computing, multimedia systems, and data mining.
\end{IEEEbiography}

\medskip

\end{document}